\documentclass[10pt,twocolumn,letterpaper]{article}

\usepackage{iccv}
\usepackage{times}
\usepackage{epsfig}
\usepackage{graphicx}
\usepackage{amsmath}
\usepackage{amssymb}
\usepackage{color}
\usepackage{subfigure}
\usepackage{booktabs}
\usepackage{tabularx}
\usepackage{multirow}
\usepackage{pifont}
\usepackage{colortbl}
\usepackage{arydshln}
\usepackage{bbm}

\newcommand{\figref}[1]{Fig. \ref{#1}}
\newcommand{\tabref}[1]{Table \ref{#1}}

\newcommand{\secref}[1]{Sec. \ref{#1}}

\makeatletter
\newcommand{\ssymbol}[1]{^{\@fnsymbol{#1}}}
\def\@fnsymbol#1{\ensuremath{\ifcase#1\or *\or \dagger\or \ddagger\or
   \mathsection\or \mathparagraph\or \|\or **\or \dagger\dagger
   \or \ddagger\ddagger \else\@ctrerr\fi}}
\def\hlinewd#1{%
	\noalign{\ifnum0=`}\fi\hrule \@height #1 \futurelet
	\reserved@a\@xhline}
\newcommand{\cmark}{\ding{51}}%
\newcommand{\xmark}{\ding{55}}%

\usepackage[pagebackref=true,breaklinks=true,letterpaper=true,colorlinks,bookmarks=false]{hyperref}

\iccvfinalcopy 

\def\iccvPaperID{7528} 

\ificcvfinal\fi

\begin{document}

\title{Deep Matching Prior: Test-Time Optimization for Dense Correspondence}
\author{Sunghwan Hong \\ 
Korea University \\
{\tt\small sung\_hwan@korea.ac.kr}
\and
Seungryong Kim\thanks{Corresponding author} \\
Korea University \\
{\tt\small seungryong\_kim@korea.ac.kr}
}

\maketitle
\ificcvfinal\thispagestyle{empty}\fi

\begin{abstract}
Conventional techniques to establish dense correspondences across visually or semantically similar images focused on designing a task-specific matching prior, which is difficult to model in general. To overcome this, recent learning-based methods have attempted to learn a good matching prior within a model itself on large training data. The performance improvement was apparent, but the need for sufficient training data and intensive learning hinders their applicability. Moreover, using the fixed model at test time does not account for the fact that a pair of images may require their own prior, thus providing limited performance and poor generalization to unseen images. 

In this paper, we show that an image pair-specific prior can be captured by solely optimizing the untrained matching networks on an input pair of images. 
Tailored for such test-time optimization for dense correspondence, we present a residual matching network and a confidence-aware contrastive loss to guarantee a meaningful convergence. Experiments demonstrate that our framework, dubbed Deep Matching Prior (DMP), is competitive, or even outperforms, against the latest learning-based methods on several benchmarks for geometric matching and semantic matching, even though it requires neither large training data nor intensive learning. With the networks pre-trained, DMP attains state-of-the-art performance on all benchmarks. 
\end{abstract}

\section{Introduction}
Establishing dense correspondences across visually or semantically similar images facilitates a variety of computer vision applications~\cite{liu2010sift,liao2017visual,jeon2019joint,kim2019semantic}. Unlike sparse correspondence~\cite{lowe2004distinctive,calonder2010brief,yi2016lift} that detects sparse points and finds matches across them, dense correspondence~\cite{liu2010sift,chang2018pyramid,kim2018recurrent,kim2019semantic} aims at finding matches at each pixel and thus can benefit from prior\footnote{It can also be called a smoothness or a regularizer in literature.} knowledge about matches among nearby pixels. 

Typically, stereo matching~\cite{kendall2017end,chang2018pyramid} and optical flow~\cite{ilg2017flownet,sun2018pwc} modeled the prior term that makes the correspondence smooth while aligning discontinuities to image boundaries. Some methods~\cite{liu2010sift,taniai2016joint,ham2016proposal,kim2017dctm} for semantic matching also exploited this to regularize the correspondence within a local neighborhood. Although they can be formulated in various ways, these optimization-based methods~\cite{liu2010sift,taniai2016joint,ham2016proposal,kim2017dctm} formulate an objective function with explicitly defined matching data and prior terms and minimize the objective on a single image pair. These methods are capable of making corrections to the estimated correspondence during optimization, but they require a task-specific prior, which is complex to formulate.   

\begin{figure}
    \centering
    \renewcommand{\thesubfigure}{}
    \subfigure[(a)]
    {\includegraphics[width=0.245\linewidth]{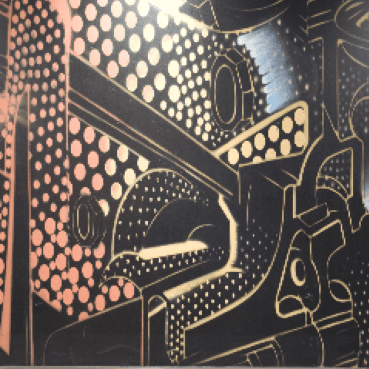}}\hfill
    \subfigure[(b)]
    {\includegraphics[width=0.245\linewidth]{figure/fig9/10_tgt.png}}\hfill
    \subfigure[(c)]
    {\includegraphics[width=0.245\linewidth]{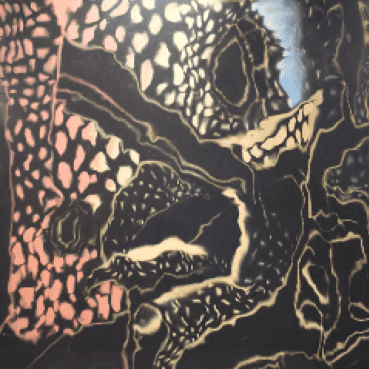}}\hfill
    \subfigure[(d)]
    {\includegraphics[width=0.245\linewidth]{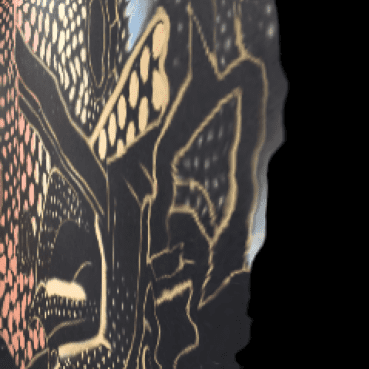}}\hfill\\
    \vspace{-5pt}
    \subfigure[(e)]
    {\includegraphics[width=0.245\linewidth]{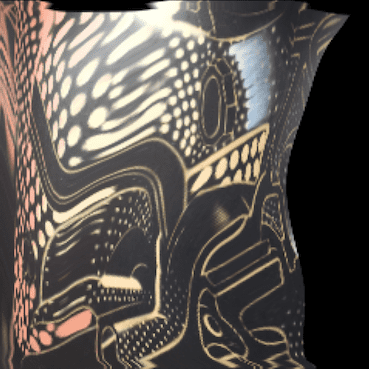}}\hfill
    \subfigure[(f)]
    {\includegraphics[width=0.245\linewidth]{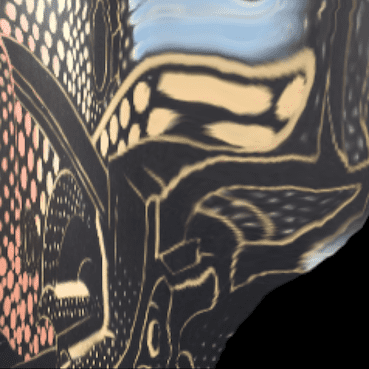}}\hfill
    \subfigure[(g)]
    {\includegraphics[width=0.245\linewidth]{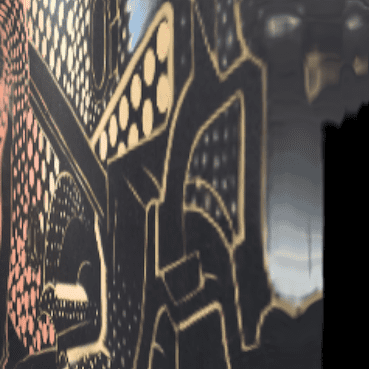}}\hfill
    \subfigure[(h)]
    {\includegraphics[width=0.245\linewidth]{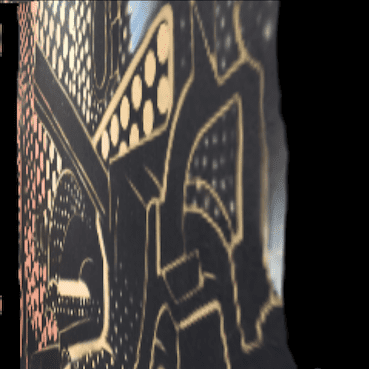}}\hfill
    \caption{\textbf{Visualization of DMP results:} (a) source image, (b) target image, results of 
    (c) winner-takes-all (WTA) and (d) learning-based method (e.g., GLU-Net~\cite{truong2020glu}). As the iteration evolves (e), (f), (g), and (h), DMP with \emph{untrained} networks estimates more optimal correspondence fields by optimizing the networks on a \emph{single pair of images}, while learning-based methods utilize \emph{pre-trained} and \emph{fixed} networks at test time, thus providing limitation.}\label{img:1}\vspace{-10pt}
\end{figure}

Unlike these learning-free methods~\cite{liu2010sift,taniai2016joint,ham2016proposal,kim2017dctm}, recent methods~\cite{rocco2017convolutional,Sun_2018_CVPR,shen2019self,melekhov2019dgc,truong2020glu} cast this task as a learning problem, seeking to learn a matching model to directly regress the correspondence. The model, often implemented based on convolutional neural networks (CNNs)~\cite{simonyan2014very,he2016deep}, is generally trained on large datasets of image pairs, based on the belief that an optimal matching prior can be learned from such observations. As proven in literature~\cite{melekhov2019dgc,truong2020glu}, these \emph{learning}-based methods outperform traditional optimization-based methods~\cite{liu2010sift,ham2016proposal,kim2017dctm}, which can be attributed to their high capacity to learn a good matching prior. They, however, often require large training data with ground-truth correspondences, which are notoriously hard to collect, or intensive learning~\cite{melekhov2019dgc,truong2020glu}, which hinders their applicability. In addition, a pair of images may require their own prior, and thus using \emph{pre-trained} and \emph{fixed} parameters at test time may provide limited and poor generalization performance to unseen image pairs.

In this paper, we show that, for the first time, the matching prior must not necessarily be learned from large training data; instead, it can be captured by optimizing the \emph{untrained} matching networks on a \emph{single pair of images}, proving that the structure of the networks is sufficient to capture a great deal of matching statistics. 
Our framework, dubbed Deep Matching Prior (DMP), requires neither large training sets nor intensive training, but it is competitive against learning-based methods~\cite{rocco2017convolutional,Sun_2018_CVPR,melekhov2019dgc,shen2019self,truong2020glu,shen2020ransac} or even outperforms, and does not suffer from the generalization issue. 

Such a test-time optimization for dense correspondence, however, is extremely hard to converge due to the limited samples, high-dimensional search space, and non-convexity of the objective. To elevate the stability and boost the convergence, we propose a residual matching network, which exploits a distilled information from matching correlation that plays as a guidance for providing a good starting match and suppressing the possibility of getting stuck in local minima. We also propose a confidence-aware contrastive loss that only takes the matches with high confidence to eliminate the ambiguous matches. \figref{img:1} visualizes the results of DMP in comparison with recent learning-based method~\cite{truong2020glu}. 

The presented approach is evaluated on several benchmarks for dense correspondence and examined in an ablation study. The extensive experiments show that our model produces competitive results and even outperforms other learning-based methods~\cite{rocco2017convolutional,Sun_2018_CVPR,shen2019self,melekhov2019dgc,truong2020glu}, and once the networks are pre-trained, state-of-the-art performance is attained on all the benchmarks in experiments.

\begin{figure*}[t]
	\centering
	\renewcommand{\thesubfigure}{}
	\subfigure[(a)]
	{\includegraphics[width=0.325\linewidth]{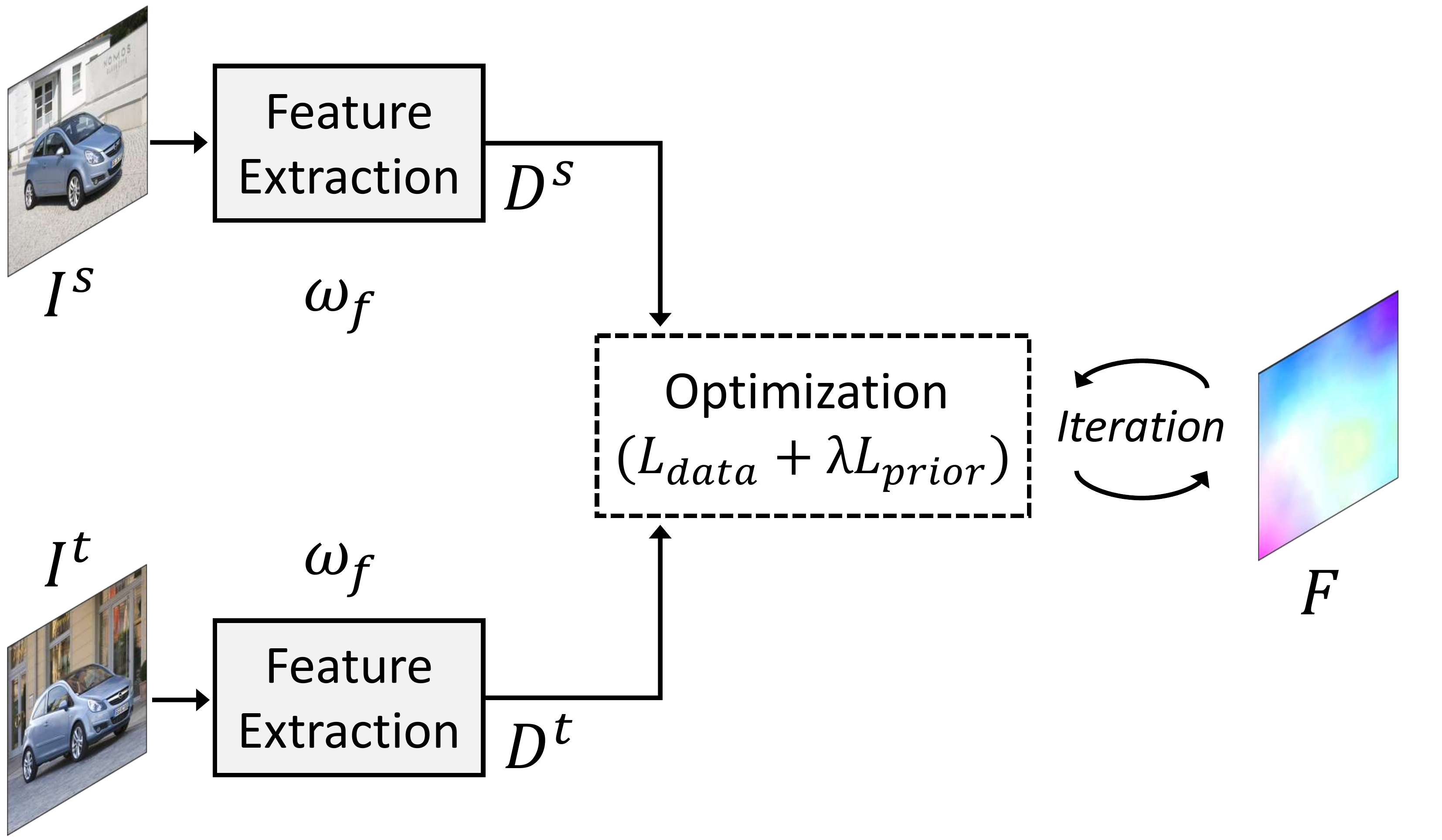}}\hfill
	\subfigure[(b)]
	{\includegraphics[width=0.325\linewidth]{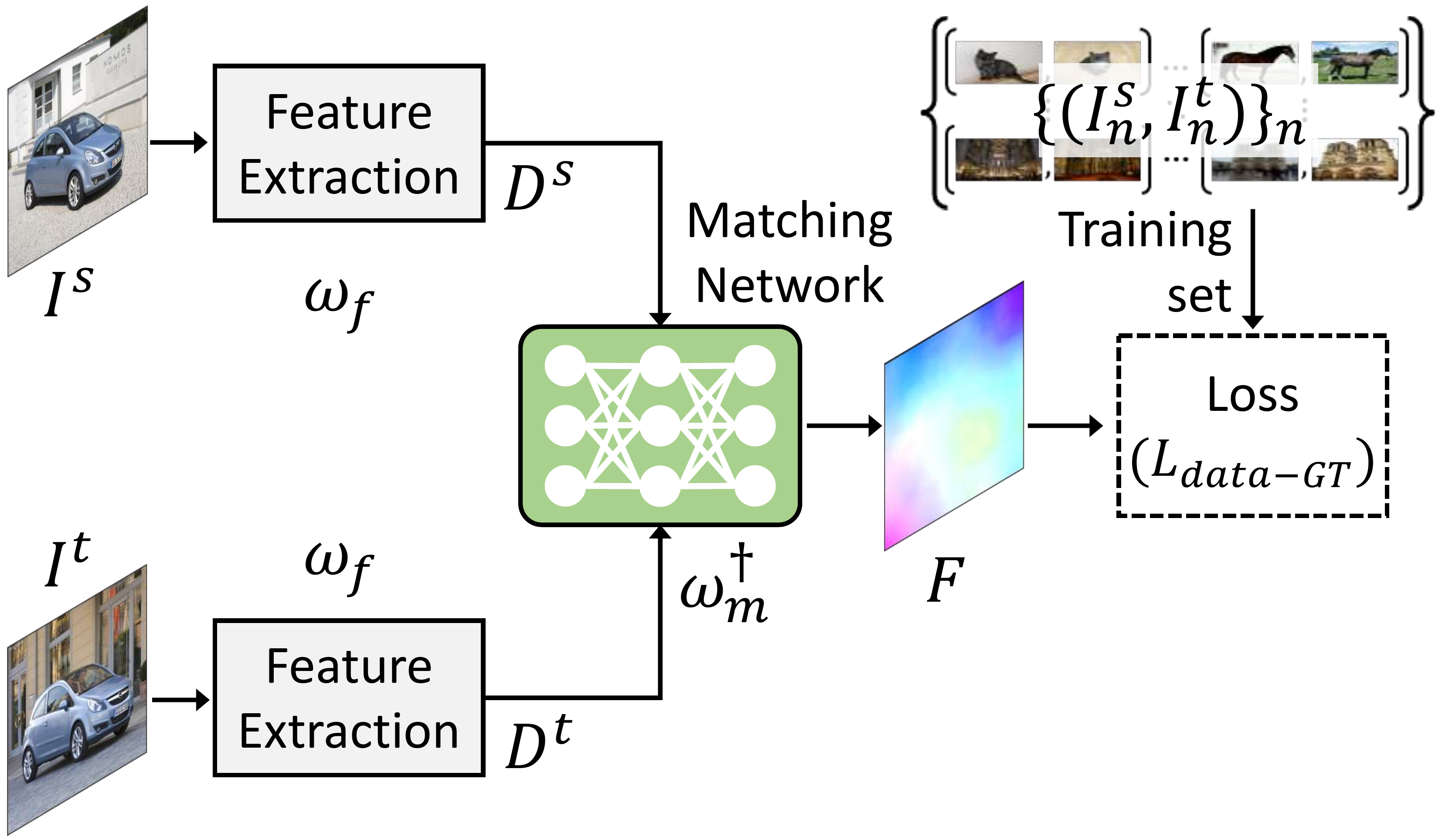}}\hfill
	\subfigure[(c)]
	{\includegraphics[width=0.325\linewidth]{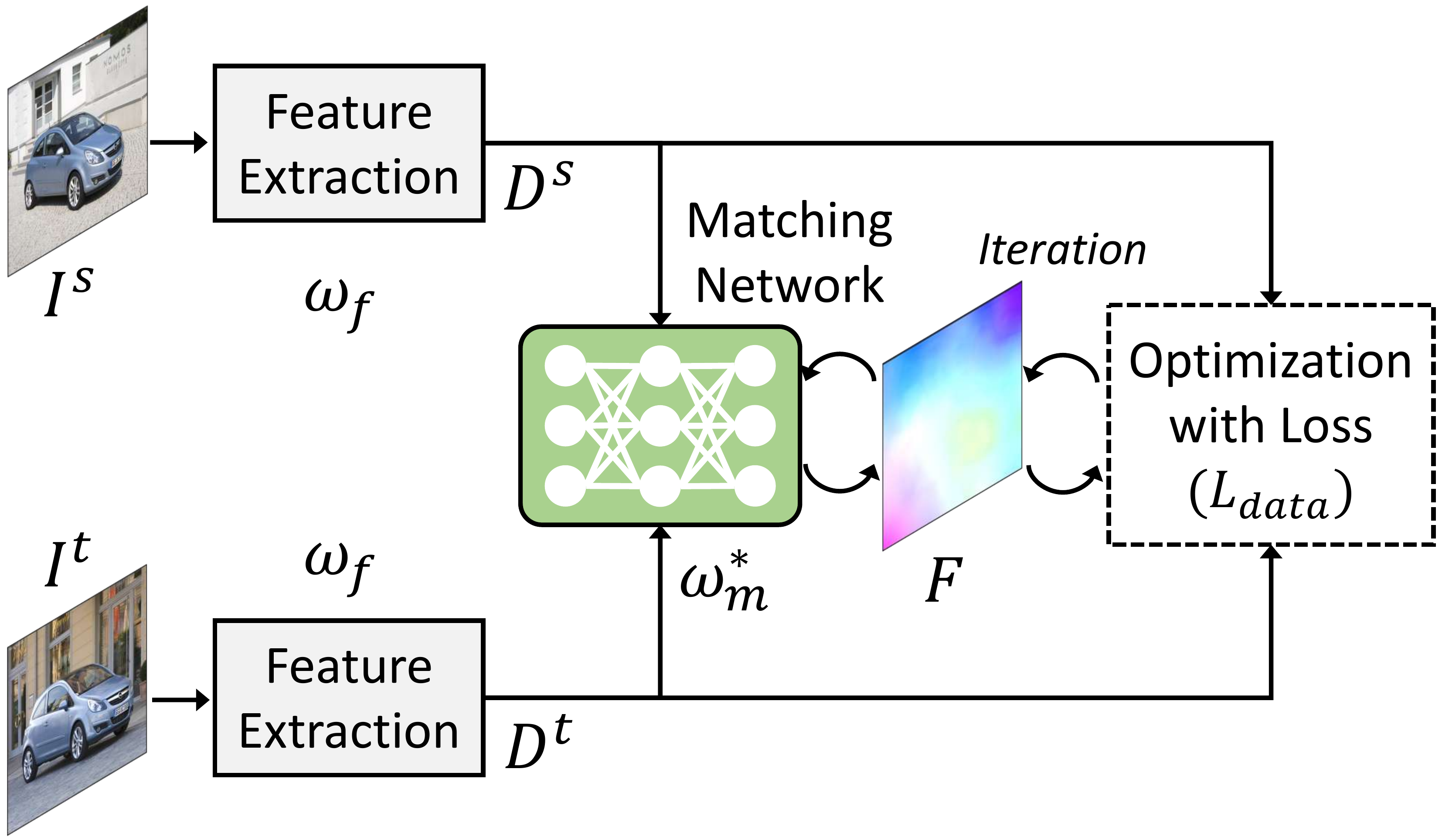}}\hfill\\
	\caption{\textbf{Intuition of DMP:} (a) optimization-based methods~\cite{liu2010sift,taniai2016joint,kim2017dctm} that formulate their objective function with data and prior terms, and minimize the energy function on a single image pair, (b) learning-based methods~\cite{melekhov2019dgc,truong2020glu,shen2020ransac} that learn a matching prior from large training set of image pairs, and (c) our DMP, which takes the best of both approaches to estimate an image pair-specific matching prior.}\label{img:2}\vspace{-10pt}
\end{figure*} 

\section{Related Works}
\paragraph{Dense correspondence.}
Most early efforts for dense correspondence~\cite{liu2010sift,kim2013deformable,taniai2016joint} have focused on designing a task-specific prior term in an objective function and improving optimization while employing hand-crafted features. Kim et al.~\cite{kim2017dctm} presented DCTM optimizer with a discontinuity-aware prior term to elevate geometric invariance. 

Similarly to other tasks, most recent works~\cite{kim2019semantic,teed2020raft,wiles2020d2d,jabri2020space,rocco2020efficient,li2020correspondence,li2020dual} leverage deep learning to achieve better performance for this task, at first replacing the hand-crafted features with the deep features~\cite{choy2016universal,kim2017fcss}, and rapidly converging towards end-to-end networks embodying the entire pipeline~\cite{rocco2018neighbourhood,kim2019semantic,li2020correspondence}. 
In this context, most recent efforts have been focusing on how to better design the matching networks and learn the networks without ground-truth correspondences. 
Rocco et al.~\cite{rocco2017convolutional,rocco2018end} proposed a geometric matching network, and their success inspired many variants that use local neighborhood consensus~\cite{rocco2018neighbourhood,li2020correspondence}, attention mechanism~\cite{huang2019dynamic}, or attribute transfer~\cite{kim2019semantic}. 
Similar to PWC-Net~\cite{sun2018pwc} designed specifically for optical flow estimation, PARN~\cite{jeon2018parn}, DGC-Net~\cite{melekhov2019dgc}, GLU-Net~\cite{truong2020glu} and GOCor~\cite{truong2020gocor} formulate their networks in a pyramidal fashion. RANSAC-Flow~\cite{shen2020ransac} leverages RANSAC~\cite{fischler1981random} within the network to improve the performance. On the other hand, to overcome the lack of ground-truth correspondences for training, synthetic supervision generated by applying random transformations~\cite{rocco2017convolutional,truong2020glu,min2020learning,truong2020gocor} and weak supervision in the form of image pairs based on the feature reconstruction~\cite{jeon2018parn,rocco2018end,kim2019semantic} have been popularly used. However, these aforementioned works require either large labeled dataset or intensive training, which hinders their applicability. 
\vspace{-10pt}

\paragraph{Self-supervised representation learning.}
Popularized by~\cite{chen2020simple,he2020momentum,chen2020improved,grill2020bootstrap}, contrastive learning seeks to learn feature representation in a self-supervised manner by minimizing the distance between two views augmented from an image and maximizing the distance to other images~\cite{hadsell2006dimensionality,dai2017contrastive,gupta2020contrastive}. Several methods~\cite{schmidt2016self,choy2016universal,kim2017dctm,kim2017fcss,wang2020dense,pinheiro2020unsupervised} bring contrastive learning to the task of dense correspondence. A key benefit of self-supervised learning is that, because there is no reliance on labeled data, training needs not be limited to the training phase~\cite{detone2016deep,oord2018representation,shocher2018zero,jabri2020space,lai2020mast,shen2020ransac}. CRW~\cite{jabri2020space} and RANSAN-Flow~\cite{shen2020ransac} attempted to apply such a self-supervised adaptation at test time, but the performance was limited. The use of self-supervised adaptation for dense correspondence has never been thoroughly investigated and this paper is the first step in this direction.
\vspace{-10pt}

\paragraph{Test-time optimization.} 
Deep Image Prior (DIP)~\cite{ulyanov2018deep} initiated the trend that the low-level statistics in a single image can be captured by the structure of randomly-initialized CNNs, of which many variants were proposed, tailored to solve an inverse problem~\cite{dong2014learning,xu2014deep,shocher2018zero,kanazawa2018end}. 
GAN-inversion aims to invert an image back to latent code, and then reconstruct the image from the latent code to utilize the pre-trained generative prior~\cite{gansteerability,abdal2019image2stylegan,gu2020image,karras2019style,karras2020analyzing,zhu2020domain}. Deep Generative Prior (DGP)~\cite{pan2020dgp} extends this trend by, for the pre-trained generator, optimizing both latent codes and parameters. These aforementioned methods were tailored to capture an image prior, but there was no attempt to capture a matching prior, which is the topic of this paper. 

\section{Deep Matching Prior}\label{sec:3}
\subsection{Motivation}\label{sec:3_1}
Let us denote a pair of images, i.e., source and target, as $I^s$ and $I^t$, which may represent visually or semantically similar images. The objective of dense correspondence is to establish a dense correspondence field $F(i)$ between two images that is defined for each pixel $i$, which warps $I^s$ towards $I^t$ so as to satisfy $I^t(i) \approx I^s(i+F(i))$.

To achieve this, the easiest solution may be, for a point in the source, to find the most similar one among all the points in the target with respect to a \emph{data} term that accounts for matching evidence between features, defined such that
\begin{equation}
F^{*} = \underset{F}{\mathrm{argmin}}
\ \mathcal{L}_{\it{data}}(D^{s}(F),D^{t}),
\end{equation}
where a feature $D$ is extracted with parameters $\omega_f$ such that $D = \mathcal{F}(I;\omega_f)$ and $D^{s}(F)$ is an warped source feature with $F$. This solution $F^{*}$, so-called winner-takes-all (WTA) solution, is highly vulnerable to local minima or outliers.
 
To alleviate these, \emph{optimization}-based methods~\cite{anguelov2005scape,liu2010sift,bleyer2011patchmatch,taniai2016joint,liao2017visual,kim2017dctm} typically formulate an objective function involving the data term and a \emph{prior} term that favors similar correspondence fields among adjacent pixels with a balancing parameter $\lambda$, as shown in~\figref{img:2} (a), such that
\begin{equation}
F^{*} = \underset{F}{\mathrm{argmin}}  
\left\{ \mathcal{L}_{\it{data}}(D^{s}(F),D^{t})
+\lambda \mathcal{L}_{\it{prior}}(F) \right\}.
\end{equation}
Traditional optimization-based methods~\cite{liu2010sift,taniai2016joint,ham2016proposal,liu2010sift,ham2016proposal,kim2017dctm} model an explicit prior term such as total variation (TV) or discontinuity-aware smoothness and optimize the energy function on a given pair of images. Since this energy function is often non-convex, they use an iterative solver, e.g., gradient descent~\cite{kingma2014adam}, to minimize this, thus they can benefit from an error feedback to find more optimal solution by making corrections to the estimated correspondence as the iteration evolves. However, they require a hand-designed task-specific prior, which is hard to design.

Unlike these optimization-based methods, recent \emph{learning}-based methods~\cite{rocco2017convolutional,Sun_2018_CVPR,shen2019self,melekhov2019dgc,teed2020raft,truong2020glu} learn a matching model to directly regress the correspondence such that $F={\mathcal{F}}(I^s,I^t;\omega_m)$ with parameters $\omega_m$, based on an assumption that an optimal prior could be learned within the model itself from massive training samples, $\{(I^s_n,I^t_n)\}_{n\in\{1,...,N\}}$, as shown in~\figref{img:2} (b). The model is often implemented in the form of convolutional neural networks (CNNs). During the training phase, the parameters $\omega^\dagger_m$ are first optimized, and the learned parameters are then used to regress a correspondence $F^{*}$ through the networks with the learned prior at test time as follows:   
\begin{equation}
\begin{split}
&\omega^\dagger_m =\underset{\omega}{\mathrm{argmin}}
\sum_{n}\mathcal{L}_{data-GT}({\mathcal{F}}(I^s_{n},I^t_{n};\omega),\hat{F}_{n}), \\
&F^{*}=\mathcal{F}(I^s,I^t;\omega^\dagger_m),
\end{split}
\end{equation}
where $\hat{F}_{n}$ denotes a ground-truth correspondence for $n$-th sample in the training set. As proven in literature~\cite{melekhov2019dgc,truong2020glu}, these learning-based methods have shown better performance than the traditional optimization-based methods~\cite{liu2010sift,ham2016proposal,kim2017dctm} thanks to their high capacity to learn the good matching prior. To optimize the parameters by minimizing the data term, e.g., $\mathcal{L}_{data-GT}(F,\hat{F})=\|F-\hat{F}\|_1$, they, however, require massive training image pairs with ground-truth correspondences, which are notoriously hard to collect. Some recent methods~\cite{rocco2017convolutional,shen2020ransac} overcome this limitation by presenting an unsupervised loss defined only with source and target features, e.g., $\mathcal{L}_{\it{data}}(D^{s}(F),D^{t})$, or synthetic ground-truths, but these methods inherit the limitation of requiring an intensive training procedure and frequently fail to learn an image pair-specific prior, thus providing limited performance and generalization power.
\begin{figure*}[t]
		\centering
		\renewcommand{\thesubfigure}{}
		\subfigure[]
		{\includegraphics[width=0.122\linewidth]{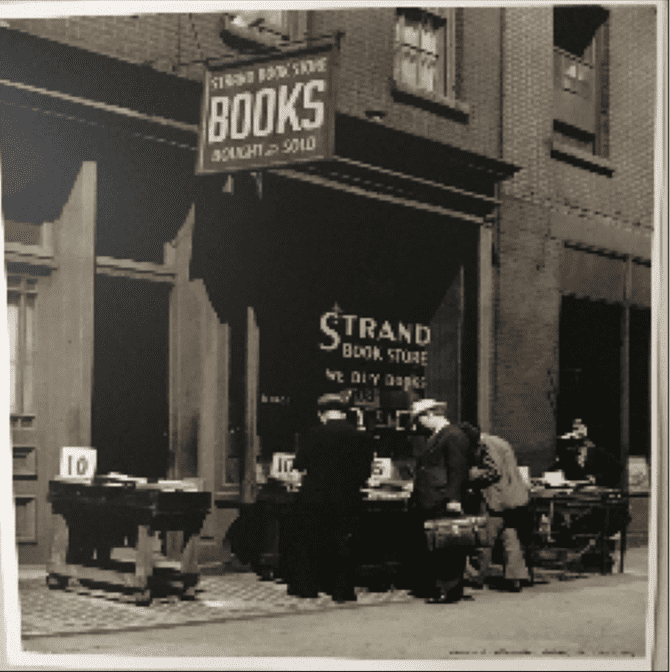}}\hfill
		\subfigure[]
		{\includegraphics[width=0.122\linewidth]{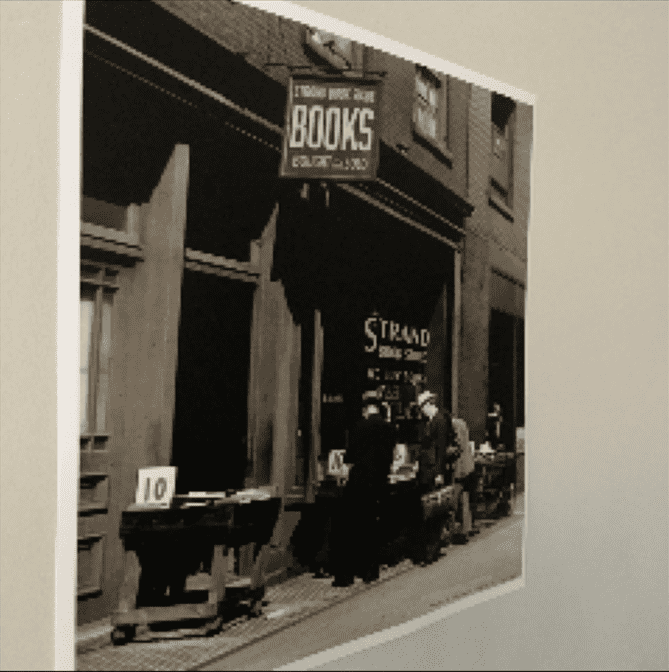}}\hfill
		\subfigure[]
		{\includegraphics[width=0.122\linewidth]{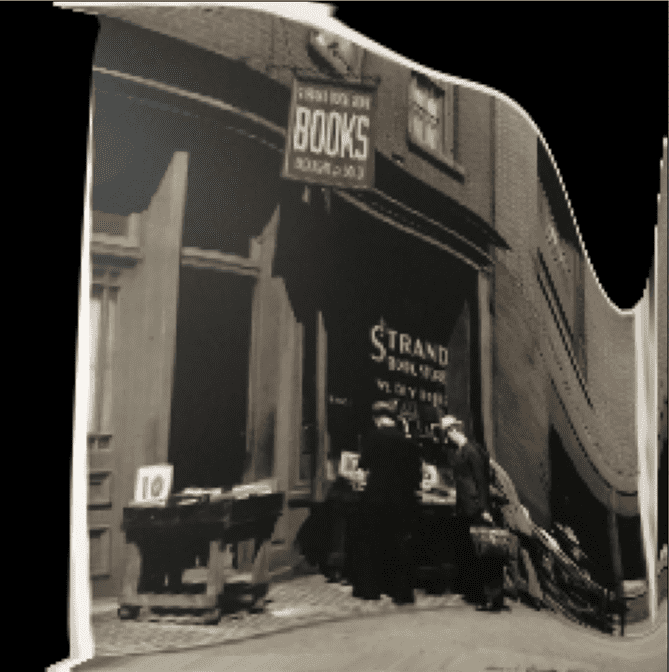}}\hfill
		\subfigure[]
		{\includegraphics[width=0.122\linewidth]{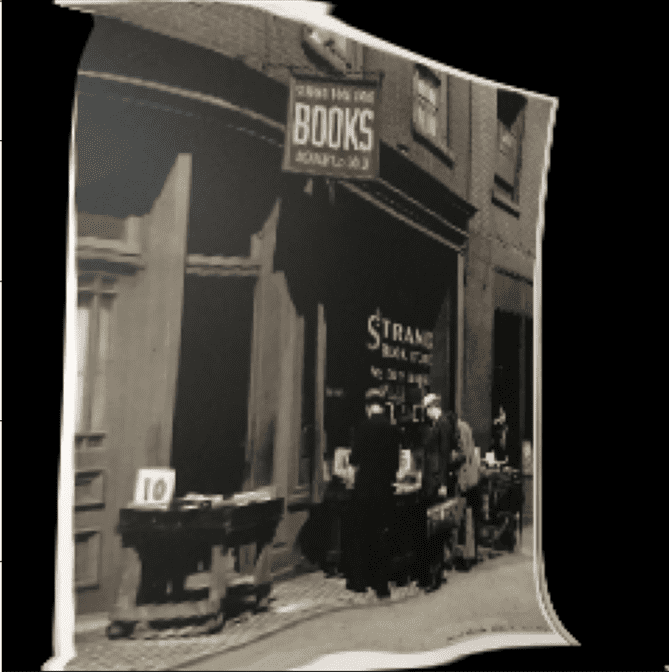}}\hfill
		\subfigure[]
		{\includegraphics[width=0.122\linewidth]{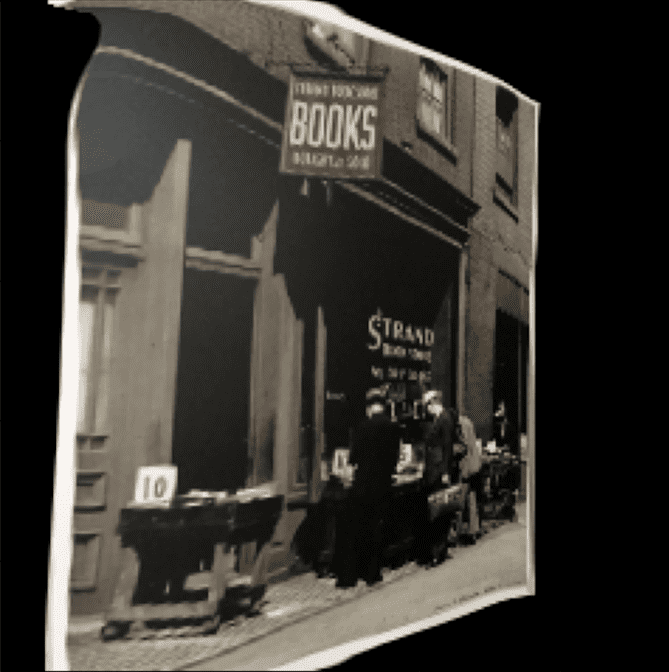}}\hfill
		\subfigure[]
		{\includegraphics[width=0.122\linewidth]{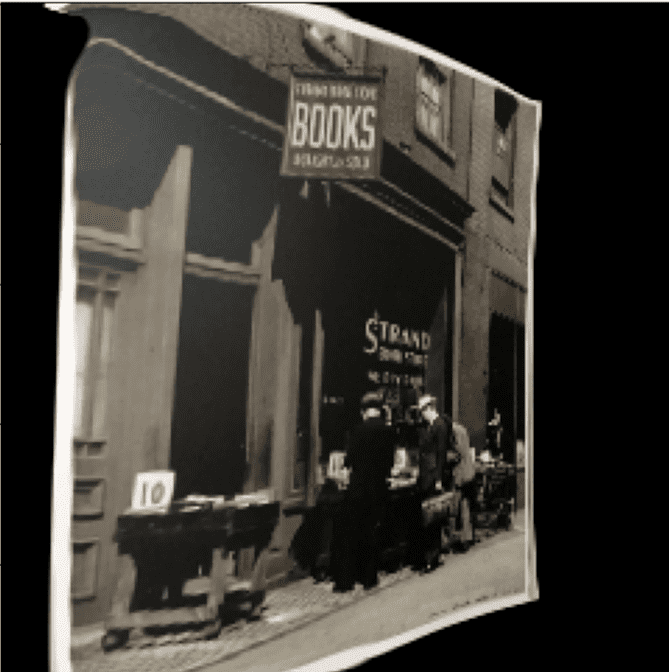}}\hfill
		\subfigure[]
		{\includegraphics[width=0.122\linewidth]{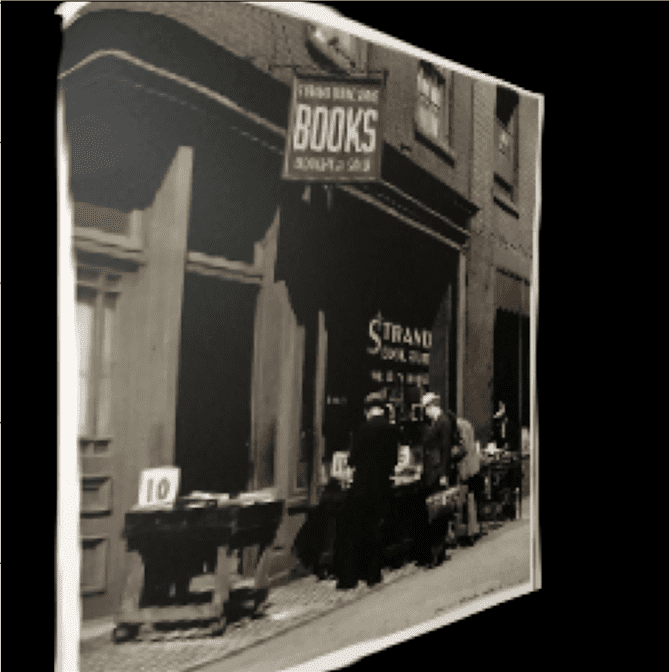}}\hfill
		\subfigure[]
		{\includegraphics[width=0.122\linewidth]{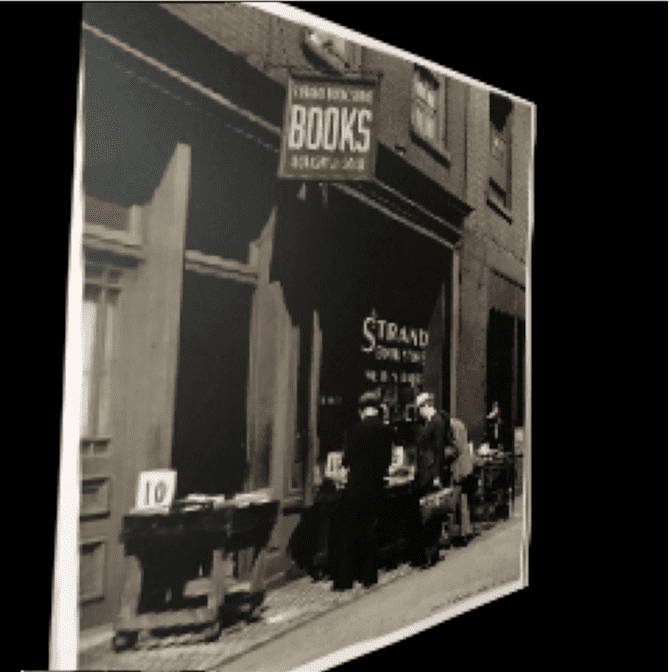}}\hfill
		\vspace{-10pt}
		\caption{\textbf{DMP convergence:} (From left to right) source image, target image, iterative evolution of warped images. Given a good initialization, which facilitates convergence and boosts matching performance, our approach successfully estimates the correspondence fields.}\vspace{-10pt}\label{img:3}
\end{figure*}

\begin{figure}
	\centering
\includegraphics[width=0.8\linewidth]{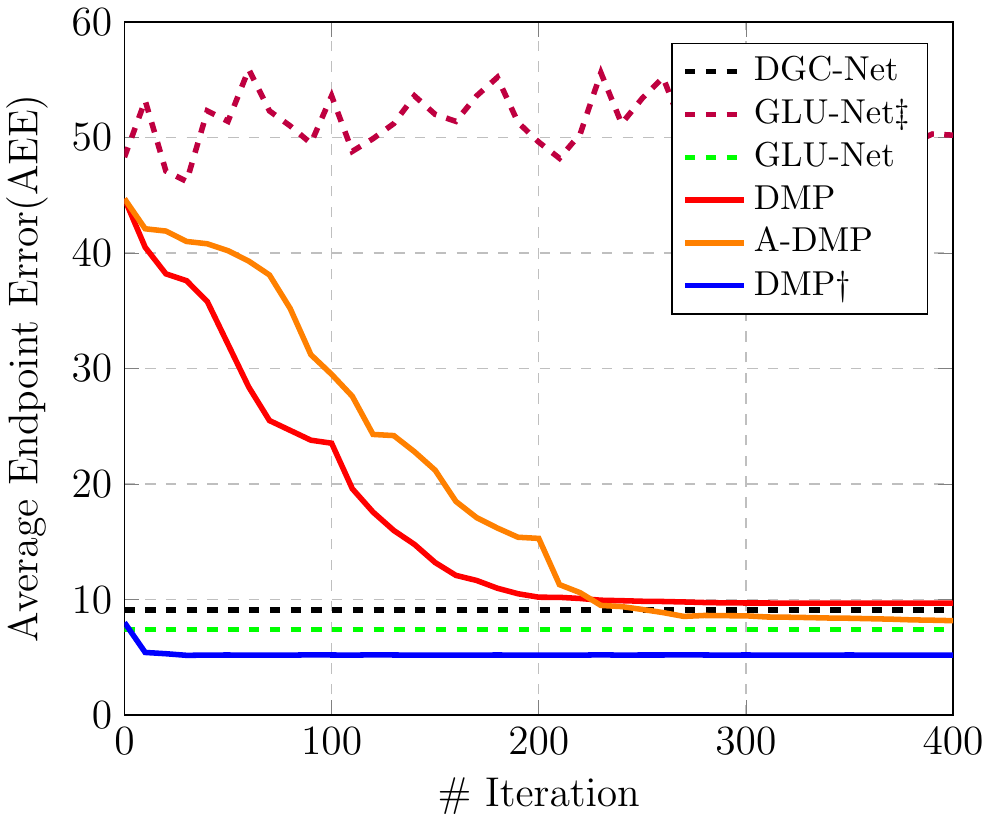}
  \caption{\textbf{Convergence analysis of DMP.} Starting from \emph{untrained} network, our DMP converges to better correspondences by optimizing the network with a well-designed loss function on a single pair of images. Compared to the \emph{learning}-based methods, such as DGC-Net~\cite{melekhov2019dgc} and GLU-Net~\cite{truong2020glu} that are \emph{pre-trained} with an intensive learning on large training datasets, our DMP has shown a competitive performance after convergence, which can be expedited with pre-trained initialization (denoted by DMP$\dagger$). More details can be found in supplementary material.}\vspace{-15pt}\label{img:4}
\end{figure}

\subsection{Formulation}\label{sec:3_2}
In this session, we argue that the matching prior does not necessarily need to be learned from an intensive learning; instead, an image pair-specific matching prior can be captured by solely minimizing the data term on a \emph{single pair of images}, like what is done by traditional optimization-based methods~\cite{liu2010sift,kim2017dctm}, with an \emph{untrained} network for matching, as shown in~\figref{img:2} (c), which can be formulated as 
\begin{equation}
\begin{split}
&\omega^{*}_m =\underset{\omega}{\mathrm{argmin}}
\ \mathcal{L}_{data}(D^{s}(\mathcal{F}(I^s,I^t;\omega)),{D}^{t}), \\
&F^{*}=\mathcal{F}(I^s,I^t;\omega^{*}_m),
\end{split}
\end{equation}
where $\omega^{*}_m$ are parameters, over-fitted to a single pair of images to allow for generating $F^{*}$ with a pair-specific matching prior, and can be obtained using an optimizer such as gradient descent~\cite{kingma2014adam}, starting from a \emph{random} initialization. Our framework, dubbed Deep Matching Prior (DMP), requires neither massive training image pairs nor ground-truth correspondences that are major bottlenecks of existing learning-based methods~\cite{melekhov2019dgc,truong2020glu,shen2020ransac}, but only requires a single pair of images to be matched, which is competitive when compared to learning-based methods~\cite{melekhov2019dgc,kim2019semantic,truong2020glu,shen2020ransac} and even outperforms, as exemplified in~\figref{img:3} and~\figref{img:4}. Thanks to its inherent error feedback nature, DMP does not suffer from a generalization issue for unseen image pairs, which existing learning-based methods~\cite{melekhov2019dgc,kim2019semantic,truong2020glu,shen2020ransac} frequently fail to avoid.

However, optimizing the existing matching network~\cite{melekhov2019dgc,truong2020glu,shen2020ransac} on a single image pair with simple data term~\cite{liu2010sift,kim2017dctm} is extremely hard to converge due to the lack of samples, high-dimensional search space for dense correspondence and its non-convexity~\cite{kim2017dctm}. 
As shown in~\figref{img:4}, when an existing matching network, e.g., GLU-Net~\cite{yang2019volumetric}, is optimized from a random initialization with a self-supervised loss, e.g., feature matching~\cite{shen2020ransac}, denoted by GLU-Net$\ddagger$, it fails to find meaningful correspondences. 
To overcome these, we present a residual matching network and a confidence-aware contrastive loss tailored to boost the convergence and matching performance. In the following, we describe the network architecture and loss function in detail.

\subsection{Network Architecture}\label{sec:3_4}
To guarantee a meaningful convergence during the optimization, a good initialization for correspondence $F$ should be set, even though our network parameters $\omega$ are randomly initialized. 
To achieve this, we formulate our model, consisting of \emph{feature extraction} networks and \emph{matching} networks, in a residual manner, as illustrated in~\figref{img:5}. \vspace{-10pt}

\paragraph{Feature extraction networks.}
Our model accomplishes dense correspondence using deep features, e.g., VGG~\cite{simonyan2014very} or ResNet~\cite{he2016deep}, which undergo $l$-2 normalization. For the backbone feature, we exploit a model pre-trained on ImageNet~\cite{deng2009imagenet}, as done in almost most literature~\cite{rocco2018neighbourhood,kim2019semantic,melekhov2019dgc,kim2019semantic,truong2020glu,shen2020ransac}. The backbone model could be directly optimized in DMP framework, but we found that using an additional adaptation layer~\cite{lee2019sfnet} to refine the backbone features and optimizing the layer only boosts the performance drastically. This is because it helps to focus on more learning matching networks at early training stages. The learnable parameters $\omega_f$ in the feature extractor are thus from the adaptation layers. At the initialization, we zero-initialize the layer to make it behave like an identity.
\vspace{-10pt}
\begin{figure*}[t]
	\centering
	\renewcommand{\thesubfigure}{}
	\subfigure[]
	{\includegraphics[width=0.9\linewidth]{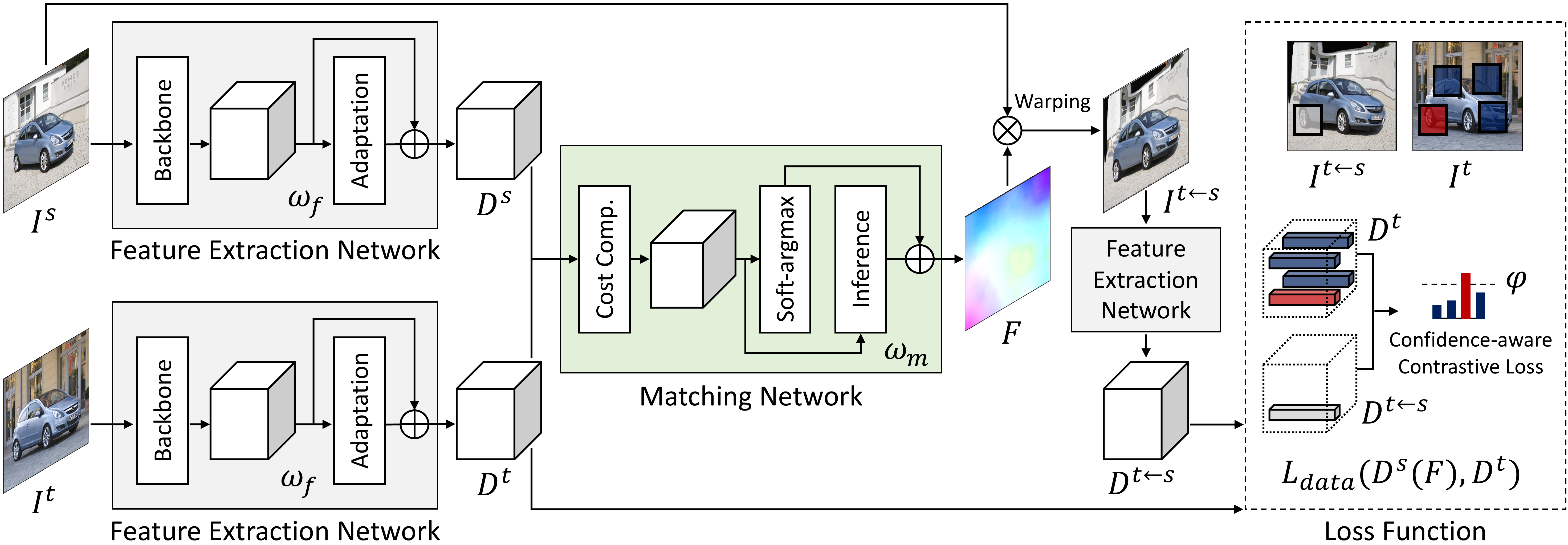}}\\
	\vspace{-10pt}
	\caption{\textbf{Network configuration and loss function of DMP:} Our networks consist of feature extraction and matching networks, which are formulated in a residual manner to guarantee a good initialization for optimization. Note that a single-level version of the networks are illustrated for brevity, while the full model is formulated in a pyramidal fashion. A confidence-aware contrastive loss enables joint learning of feature extraction and matching networks by rejecting ambiguous matches while accepting confident matches through a thresholding.}\vspace{-10pt}\label{img:5}
\end{figure*}

\paragraph{Matching networks.}
Our matching networks consist of correlation map computation and inference modules. The correlation maps are first computed using the inner product between features as
\begin{equation}
\mathcal{C}(i,l)={D^{s}(i)}\cdot D^{t}(l),
\end{equation}
where $l$ is defined within a search space in the target. 

Based on the correlation $\mathcal{C}$, correspondence $F_\mathrm{s}$ is estimated through an inference module with parameters $\omega_m$ as
\begin{equation}
F_\mathrm{s} = \mathcal{F}(\mathcal{C};\omega_m). 
\end{equation}
A randomly-initialized $\omega_m$, however, gives a noisy correspondence at the initial phase, making the optimization difficult. To overcome this, we enforce the inference module to estimate a residual of current best matches from given correlation volume, which is achieved by a soft-argmax operation~\cite{Kendall_2017_ICCV} over $\mathcal{C}$, such that
\begin{equation}
F = \mathcal{F}(\mathcal{C};\omega_m) + \Phi(\mathcal{C}),  
\end{equation}
where $\Phi(\cdot)$ represents a soft-argmax operator. 
By setting the parameters of the last layer to generate zeros at the initialization, an optimizer can start at least the current best matches from given correlation volume. As evolving the iterations, the networks provide more regularized matching fields that encode an image pair-specific prior. 
\vspace{-10pt}

\paragraph{Coarse-to-fine formulation.}
Analogous to~\cite{Sun_2018_CVPR,sun2018pwc,jeon2018parn,melekhov2019dgc,truong2020glu}, we also utilize the coarse-to-fine approach through a pyramidal processing. Specifically, we exploit pyramidal features from coarse-to-fine levels to simultaneously provide robustness against deformations and improve fine-grained matching details by minimizing the data term at each layer. At the coarsest level, we consider a global correlation module, while a local correlation module is used at remaining levels. In addition to the use of  soft-argmax for residual flow learning, at each level, previous level's matches are up-sampled to play as a guidance. To handle any input resolution, we employ an adaptive resolution~\cite{truong2020glu}. The details can be found in the supplementary material.

\begin{table*}[t]
	\centering
	\scalebox{0.76}{
	\begin{tabular}{ >{\raggedright}m{0.17\linewidth}|
				>{\centering}m{0.040\linewidth} 
				>{\centering}m{0.040\linewidth}|
				>{\centering}m{0.045\linewidth} 
				>{\centering}m{0.045\linewidth}
				>{\centering}m{0.045\linewidth}
				>{\centering}m{0.045\linewidth}
				>{\centering}m{0.045\linewidth}|
				>{\centering}m{0.045\linewidth}|
				>{\centering}m{0.045\linewidth}|
				>{\centering}m{0.045\linewidth}
				>{\centering}m{0.045\linewidth}
				>{\centering}m{0.045\linewidth}
				>{\centering}m{0.045\linewidth}
				>{\centering}m{0.045\linewidth}|
				>{\centering}m{0.045\linewidth}|
				>{\centering}m{0.045\linewidth}
				}
			\hlinewd{0.8pt}
			
			\multirow{2}{*}{Methods} & Pre- & Test- & \multicolumn{7}{ c |}{Hpatches ($240 \times 240$)} & 
			\multicolumn{7}{ c }{Hpatches}\tabularnewline
			\cline{4-17}
			& train & opt. &I &II &III &IV &V &Avg. &PCK &I &II &III &IV &V &Avg. &PCK \tabularnewline
			\hline
			\hline
			CNNGeo~\cite{rocco2017convolutional} &\cmark  &\xmark&9.59&18.55&21.15&27.83&35.19&22.46&-&-&-&-&-&-&-&- \tabularnewline
			DGC-Net~\cite{melekhov2019dgc}  &\cmark  &\xmark &1.74&5.88&9.07&12.14&16.50&9.07&50.01&5.71&20.48&34.15&43.94&62.01&33.26&58.06 \tabularnewline
			GLU-Net~\cite{truong2020glu} &\cmark &\xmark &0.59&4.05&7.64&9.82&14.89&7.40&83.47&1.55&12.66&27.54&32.04&52.47&25.05&78.54 \tabularnewline
			GLU-GOCor~\cite{truong2020gocor} &\cmark &\xmark &- &-&-&-&-&-&-&1.29&10.07&23.86&27.17 &38.41&20.16&81.43 \tabularnewline
			\hdashline
			\multirow{3}{*}{RANSAC-Flow~\cite{shen2020ransac}} &\cmark &\xmark  &0.51&2.36&2.91&4.41&\bf{5.12}&3.06 &-&-&-&-&-&-&-&- \tabularnewline
			&\xmark &\cmark &3.81&5.76&5.92&8.31 &13.24 &7.40 &81.28  &12.41 &18.64 &20.09 &35.81 &35.00 &24.39&22.06  \tabularnewline
			&\cmark &\cmark &1.29&3.52 &4.01&6.88 &10.27 &5.19 &88.23 &10.25 &15.22 &15.19 &30.43 &48.09 &23.84 &26.21 \tabularnewline
			\hline
			\textbf{DMP} &\xmark &\cmark &1.21&5.12&12.31&13.68&16.12&9.69&79.21&3.21&15.54&32.54&38.62&63.43&30.64&63.21
			\tabularnewline
			\textbf{A-DMP} &\xmark &\cmark &1.31&4.81&10.21&10.69&13.88&8.18&82.35 &3.42&14.21&29.90&32.82&55.8&27.22&71.64 \tabularnewline
			RANSAC-\textbf{DMP} &\xmark &\cmark &0.53&\bf{2.21}&2.76&4.62&5.14&3.05 &96.28&4.32&11.21&22.80&31.34&33.64&20.65&75.35 \tabularnewline
            \hdashline
			\multirow{2}{*}{\textbf{DMP}$\dagger$}
			&\cmark &\xmark &1.48&4.67&7.82&9.96&13.68&7.53&79.94&4.24&15.92&27.42&36.77&46.51&26.15&54.84 \tabularnewline
			&\cmark &\cmark &0.77&3.36&5.22&7.32&9.38&5.21&90.89&2.41&9.88&20.64&28.21&30.15&18.23&81.74 \tabularnewline
			\hdashline
			RANSAC-\textbf{DMP}$\dagger$ &\cmark&\cmark &\bf{0.48} &2.24&\bf{2.41}&\bf{4.32}&5.16&\bf{2.92}&\bf{97.52}&\bf{3.57}&\bf{8.59}&\bf{10.18}&\bf{21.21}&\bf{23.81}&\bf{13.47} &\bf{87.62} \tabularnewline
			\hlinewd{0.8pt}
		\end{tabular}}
		\vspace{+5pt}
	\caption{\textbf{Quantitative evaluation on HPatches~\cite{balntas2017hpatches} dataset in terms of AEE and PCK.} 
	Lower AEE and higher PCK (5-pixel ($\%$)) are better. \emph{Pre-train: Pre-training, Test-opt.: Test-time optimization.}}\label{tab:1}\vspace{-5pt}
\end{table*}

\begin{figure*}[t]
	\centering
	\renewcommand{\thesubfigure}{}
	\subfigure[]
	{\includegraphics[width=0.122\linewidth]{figure/fig9/10_src.png}}\hfill
	\subfigure[]
	{\includegraphics[width=0.122\linewidth]{figure/fig9/10_tgt.png}}\hfill
	\subfigure[]
	{\includegraphics[width=0.122\linewidth]{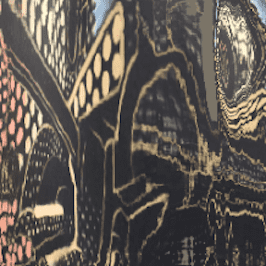}}\hfill
	\subfigure[]
	{\includegraphics[width=0.122\linewidth]{figure/fig9/10_glu.png}}\hfill
	\subfigure[]
	{\includegraphics[width=0.122\linewidth]{figure/fig9/dmp_fig_9.png}}\hfill
	\subfigure[]
	{\includegraphics[width=0.122\linewidth]{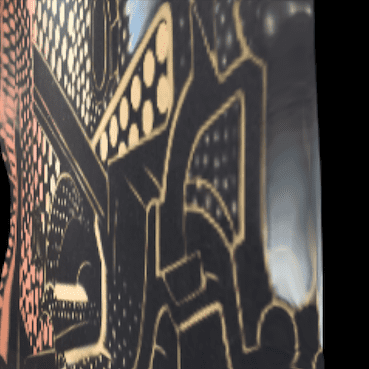}}\hfill
	\subfigure[]
	{\includegraphics[width=0.122\linewidth]{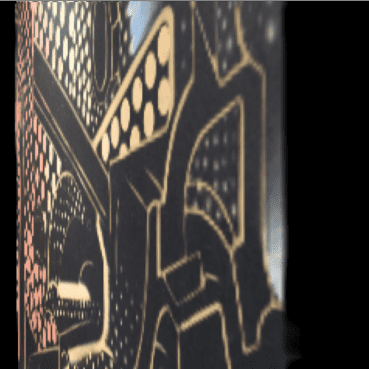}}\hfill
	\subfigure[]
	{\includegraphics[width=0.122\linewidth]{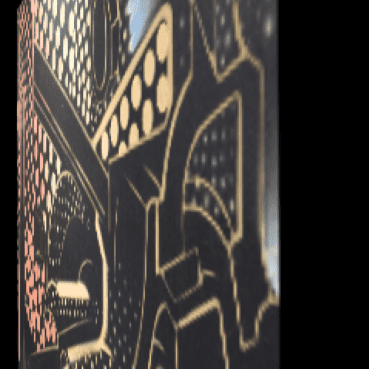}}\hfill\\
	\vspace{-21.5pt}
	\subfigure[(a) Source]
	{\includegraphics[width=0.122\linewidth]{figure/fig9/283_src.png}}\hfill
	\subfigure[(b) Target]
	{\includegraphics[width=0.122\linewidth]{figure/fig9/283_tgt.png}}\hfill
	\subfigure[(c) DGC-Net~\cite{melekhov2019dgc}]
	{\includegraphics[width=0.122\linewidth]{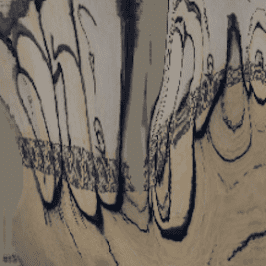}}\hfill
	\subfigure[(d) GLU-Net~\cite{truong2020glu}]
	{\includegraphics[width=0.122\linewidth]{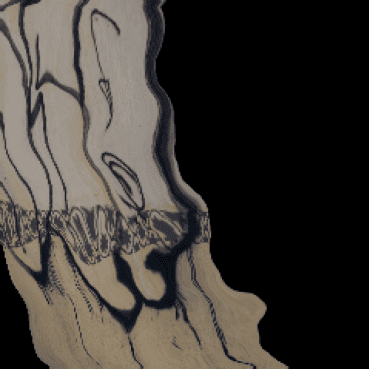}}\hfill
	\subfigure[(e) DMP]
	{\includegraphics[width=0.122\linewidth]{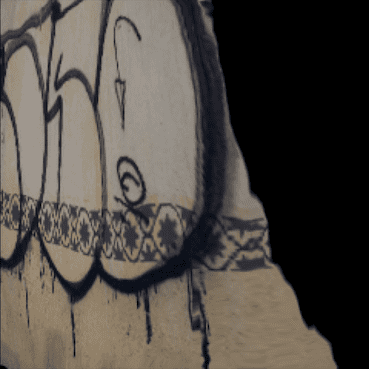}}\hfill
	\subfigure[(f) A-DMP]
	{\includegraphics[width=0.122\linewidth]{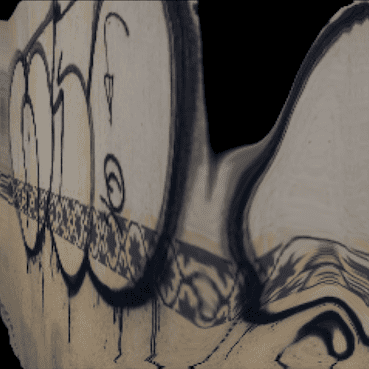}}\hfill
	\subfigure[(g) DMP$\dagger$]
	{\includegraphics[width=0.122\linewidth]{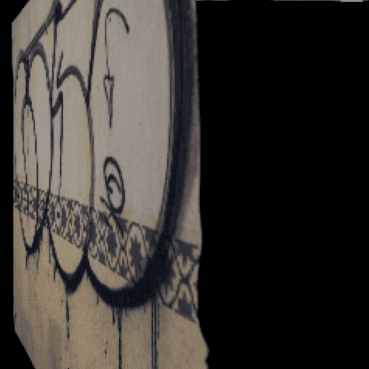}}\hfill
	\subfigure[(h) Ground-truth]
	{\includegraphics[width=0.122\linewidth]{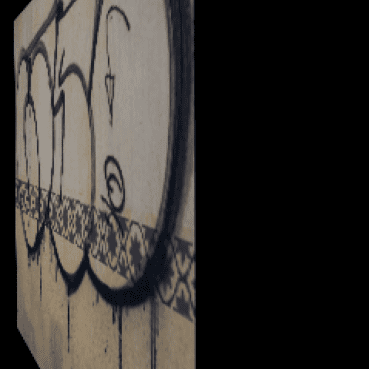}}\hfill
    \caption{\textbf{Qualitative results on HPatches~\cite{balntas2017hpatches} dataset.} The source images were warped to the target images using correspondences.}\vspace{-10pt}\label{img:6}
\end{figure*}

\subsection{Loss function}\label{sec:3_3}
Similarly to traditional optimization-based methods~\cite{liu2010sift,kim2017dctm} and unsupervised learning-based methods~\cite{rocco2018end,kim2019semantic}, the data term in DMP is defined as the similarity function $\mathcal{S}(\cdot,\cdot)$ of features, aggregated across the points $i \in \{1,...,M\}$ as
\begin{equation}\label{eq6}
\mathcal{L}_\mathrm{data}(D^{s}(F),{D}^{t}) = -\frac{1}{M}\sum_{i} \mathcal{S}(D^{t\leftarrow s}(i),{D}^{t}(i)),
\end{equation}
where $D^{t\leftarrow s}$ represents an warped source feature with $F$. Although source feature can be directly warped, we extract the feature directly from warped source image $I^{t\leftarrow s}$ with the final correspondence $F$ at the pyramid that incorporates the information at all pyramidal layers, which demonstrates more stable convergence.

Perhaps the simplest similarity function is to use the inner product between $l$-2 normalized features as $\mathcal{S}(D^{t\leftarrow s}(i),{D}^{t}(i)) = D^{t\leftarrow s}(i)\cdot{D}^{t}(i)$,
accounting for the intuition that the similarities between all pairs of the warped source and target features at the same locations should be maximized. Minimizing this, however, can induce erroneous solutions, e.g., constant features at all the points. To avoid such trivial solutions, we extend this similarity function in a contrastive learning fashion, aiming to maximize the similarities at the same locations while minimizing for the others, such that given
\begin{equation}
\begin{split}
\mathcal{S}_\mathrm{c}=
\left(\frac{ \exp(D^{t\leftarrow s}(i)\cdot{D}^{t}(i) / \tau) }
{  \sum_j \exp(D^{t\leftarrow s}(i)\cdot{D}^{t}(j) / \tau) } \right),
\end{split}
\end{equation}
where $j\in \{1,...,M\}$ and $\tau$ is a temperature hyperparameter, 
by minimizing the loss function $-\mathrm{log}(\mathcal{S}_\mathrm{c})$, both feature extraction and matching networks with parameters $\omega_f$ and $\omega_m$, respectively, can be simultaneously optimized in a manner that the feature extractors embed the warped source and target features at same location to the same representation while the matching networks regress better correspondences that maximize the feature similarity. 

Even though this loss helps to successfully avoid the trivial solution and allows joint learning of feature and matching networks, the erroneous initial estimates may be propagated, without a term to eliminate such estimates. Since our DMP optimizes the networks on a single pair of images, such adverse effects may be critical. To mitigate this, we present a confidence-aware contrastive loss that enables rejecting such ambiguous matches with a thresholding while accepting the confident matches, defined such that

\begin{equation}\label{eq7}
\mathcal{S}_\mathrm{cac}=-\mathrm{log}\left( \Psi \left(\mathcal{S}_c, \varphi \right) \right),
\end{equation}
where $\varphi$ is a confidence hyperparameter; $\Psi(\mathrm{x},\varphi)$ is a function designed to produce $\mathrm{x}$ if $\mathrm{x}\geq\varphi$ and $1$ otherwise.

\subsection{Extension}\label{sec:3_4}
In this section, we introduce several variants of our DMP to expedite convergence and boost the performance.
\vspace{-10pt}

\paragraph{Test-time augmentation.}
Since DMP solely depends on a single pair of images, the samples that capture matching distribution are rather limited, even though it surprisingly well learns from such limited samples, which inspires the enrichment through a test-time augmentation. 
We present Augmentation-DMP (A-DMP), where original target and a randomly-augmented target are fed concurrently to our model, and used to optimize the networks simultaneously. Given a pair of images having dramatically large geometric variations, the randomly-augmented target would be more similar to a source than original target at an iteration, generating the loss signal to accelerate the convergence. In addition, thanks to our confidence-aware loss, if augmented ones are more difficult to be matched, they are rejected. We apply one of spatial transformations at each iteration such as homography~\cite{detone2016deep} or affine and thin plate spline (TPS)~\cite{rocco2017convolutional} to obtain the augmented targets. 
\vspace{-10pt}

\paragraph{Pre-training.}
Unsurprisingly, well-initialized parameters can guarantee better performance. We also show that by pre-training our networks, similarly to others~\cite{truong2020glu,truong2020gocor}, denoted by DMP$\dagger$, our DMP can start with a well learned initialization, thus yielding much better results. Note that \cite{shen2020ransac} also attempted to pre-train and fine-tune their networks on a single pair of images, but as will be seen in our experiments, it failed to achieve the satisfactory performance, which proves that our well-designed networks and loss functions are essential for such a framework. 
In this work, we follow the training procedure identical to~\cite{truong2020glu} and use the same dataset and hyperparameters.

\section{Experiments}
\subsection{Implementation Details}\label{sec:4_1}
For backbone feature extractor, we used VGG-16~\cite{simonyan2014very} and ResNet-101~\cite{he2016deep} pre-trained on ImageNet~\cite{deng2009imagenet}. Specifically, for geometric matching, we used VGG-16, while for semantic matching, we additionally used ResNet-101. 
For the loss function, instead of using all the samples, we randomly sampled $M=$ 256 feature vectors at each iteration, considering the trade-off between performance and complexity. We set the maximum number of iterations as 2k and 0.3k for DMP (or A-DMP) and DMP$\dagger$, respectively. In addition, we set the temperature $\tau$ and confidence $\varphi$ as 0.1 and 0.01, respectively, following ablation study in~\secref{sec:4_5}. We used Adam optimizer~\cite{kingma2014adam} fixed for all experiments, but the advanced optimizers may improve performance~\cite{reddi2019convergence,liu2019variance}. 

\subsection{Experimental Settings}\label{sec:4_2}
In this section, we conduct comprehensive experiments for geometric matching and semantic matching, by evaluating our approach through comparisons to state-of-the-art methods including, CNNGeo~\cite{rocco2017convolutional}, PARN~\cite{jeon2018parn}, NC-Net~\cite{rocco2018neighbourhood}, DGC-Net~\cite{melekhov2019dgc}, SAM-Net~\cite{kim2019semantic}, GLU-Net~\cite{truong2020glu}, GOCor~\cite{truong2020gocor}, RANSAC-Flow~\cite{shen2020ransac}, and SCOT~\cite{liu2020semantic}. For the geometric matching task, we evaluate our method on Hpatches~\cite{balntas2017hpatches} and ETH3D~\cite{schops2017multi} datasets, while for the semantic matching task, we evaluate our method on TSS~\cite{taniai2016joint} and PF-PASCAL~\cite{ham2016proposal} datasets. We also include the results of our variants for all the experiments and analyze the influence of different components of our method. 
\begin{figure}[t]
	\centering
	\renewcommand{\thesubfigure}{}
	\subfigure[]
	{\includegraphics[width=0.25\linewidth]{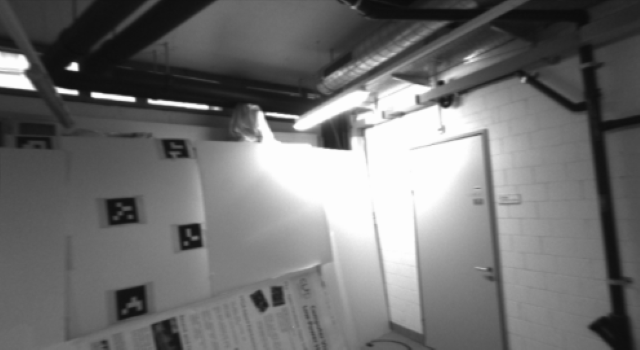}}\hfill
	\subfigure[]
	{\includegraphics[width=0.25\linewidth]{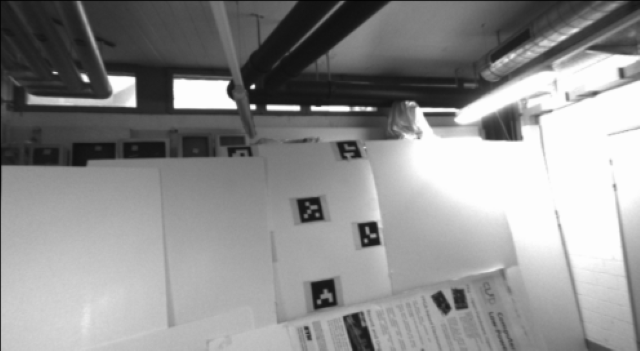}}\hfill
	\subfigure[]
	{\includegraphics[width=0.25\linewidth]{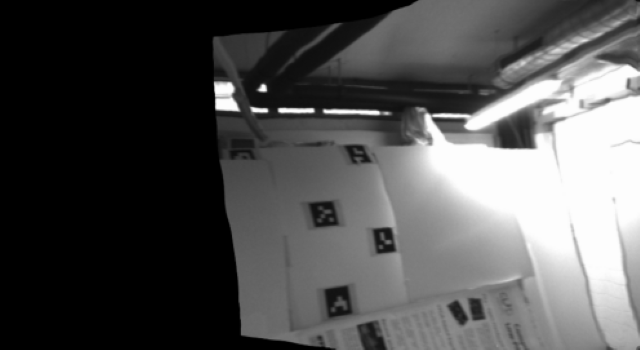}}\hfill
	\subfigure[]
	{\includegraphics[width=0.25\linewidth]{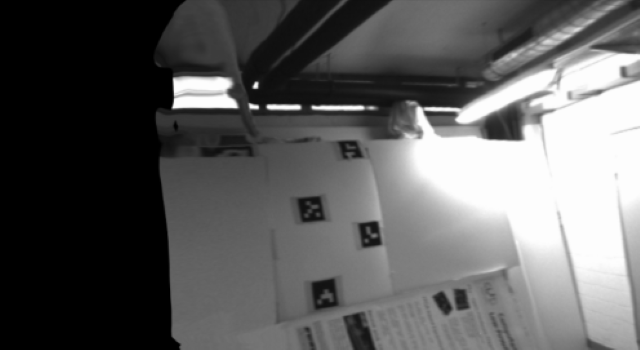}}\hfill\\
	\vspace{-21.5pt}
	\subfigure[(a) Source]
	{\includegraphics[width=0.25\linewidth]{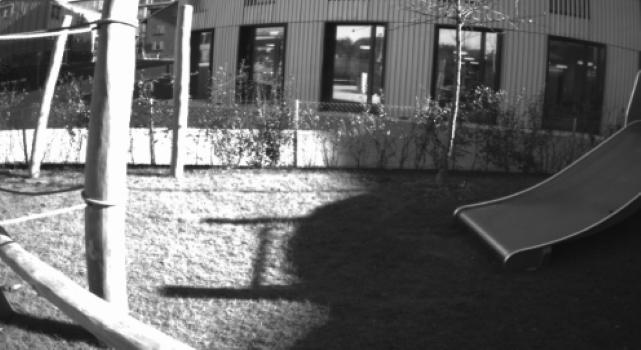}}\hfill
	\subfigure[(b) Target]
	{\includegraphics[width=0.25\linewidth]{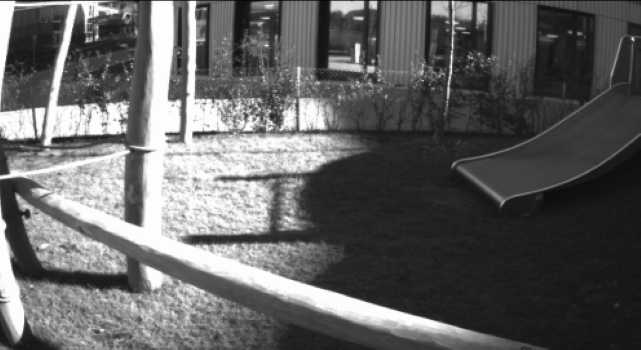}}\hfill
	\subfigure[(c) GLU-Net~\cite{truong2020glu}]
	{\includegraphics[width=0.25\linewidth]{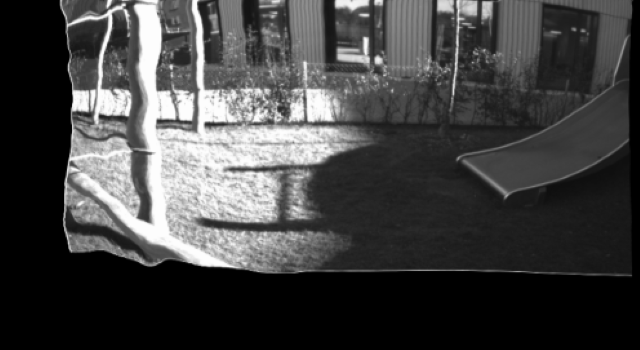}}\hfill
	\subfigure[(d) DMP]
	{\includegraphics[width=0.25\linewidth]{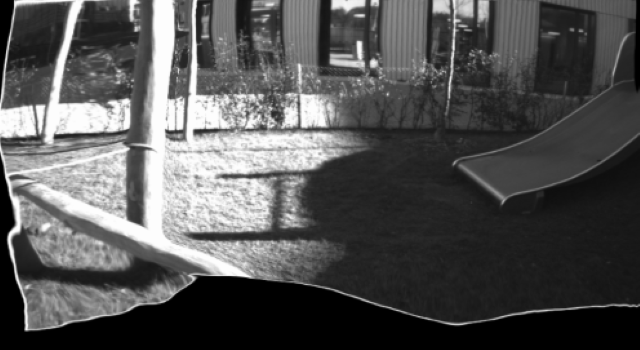}}\hfill\\
    \caption{\textbf{Qualitative results on ETH3D~\cite{schops2017multi} dataset.}}\vspace{-10pt}\label{img:7}
\end{figure}

\subsection{Geometric Matching Results}\label{sec:4_3}
\paragraph{HPatches.}
We first evaluate our method on Hpatches~\cite{balntas2017hpatches} which consists of images with different views of the same scenes. Each sequence contains a source and five target images with different viewpoints with corresponding ground-truth flows. We use images of high resolutions ranging from 450 $\times$ 600 to 1,613 $\times$ 1,210, whereas the down-scaled images (240 $\times$ 240) are also used for the evaluation as in~\cite{melekhov2019dgc}. For the evaluation metric, we use the Average Endpoint Error (AEE), computed by averaging the euclidean distance between the ground-truth and estimated flow, and Percentage of Correct Keypoints (PCK), computed as the ratio of estimated keypoints within the threshold from ground-truths to the total number of keypoints.

\tabref{tab:1} summarizes the quantitative results, and \figref{img:6} visualizes the qualitative results. We also provide an ablation study for RANSAC-Flow~\cite{shen2020ransac} to validate the effect of test-time optimization, which deteriorates performance as studied in~\cite{shen2020ransac}. However, DMP is competitive against the state-of-the-art learning-based methods~\cite{melekhov2019dgc,truong2020glu,shen2020ransac} thanks to the feedback signals during the test-time optimization. Leveraging RANSAC~\cite{fischler1981random} within our network allows outstanding performance, and with the networks pre-trained, DMP$\dagger$ even outperforms GOCor~\cite{truong2020gocor} by a large margin. 
\vspace{-10pt}
\begin{figure}
	\centering
	\renewcommand{\thesubfigure}{}
	\subfigure[(a) AEE]
	{\includegraphics[width=0.495\linewidth]{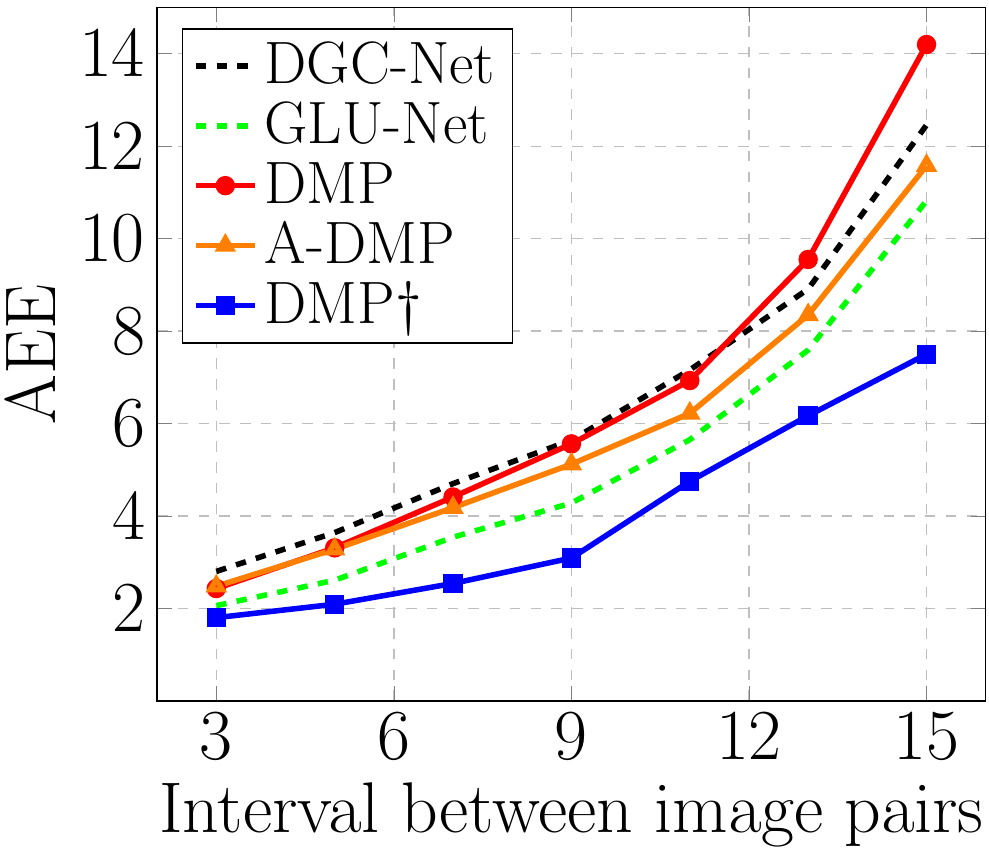}}\hfill
	\subfigure[(b) PCK-5px]
	{\includegraphics[width=0.495\linewidth]{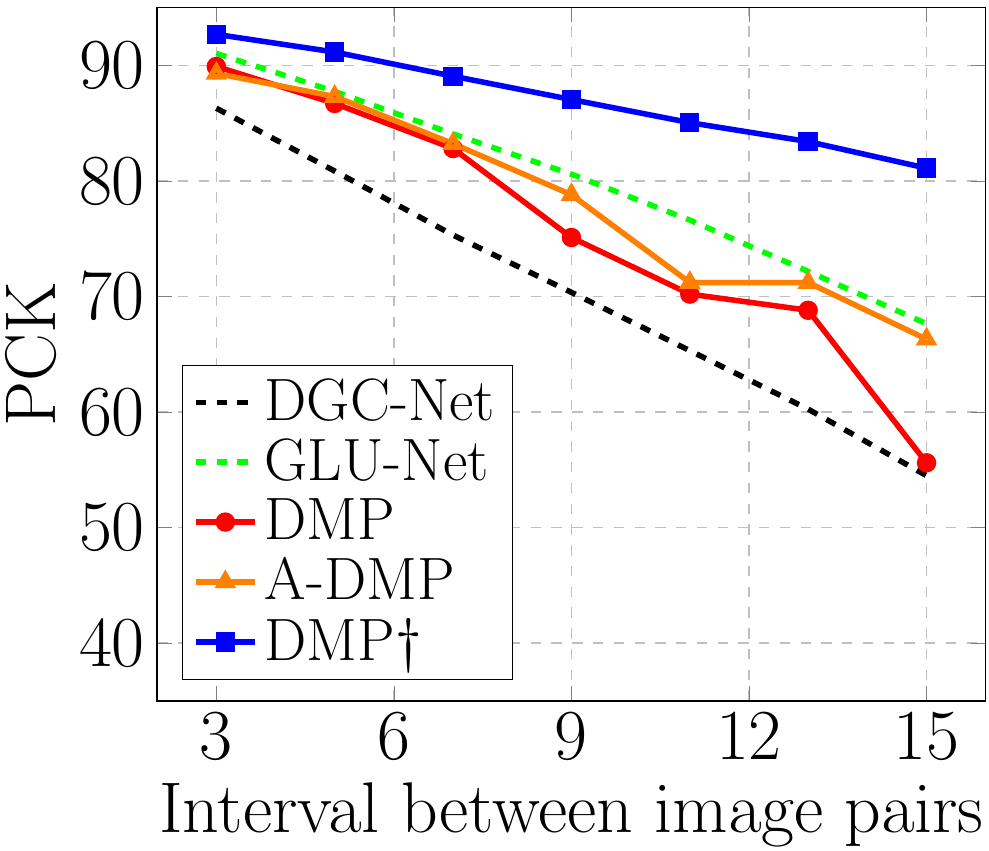}}\hfill\\
    \caption{\textbf{Quantitative results on ETH3D~\cite{schops2017multi} dataset.} AEE and PCK are computed on image pairs sampled at different intervals.}
\label{ETH3d}\vspace{-5pt}\label{img:8}
\end{figure}

\begin{table}[t]
	\centering
	\scalebox{0.76}{
		\begin{tabular}{ >{\raggedright}m{0.28\linewidth}|
				>{\raggedright}m{0.21\linewidth}|
				>{\centering}m{0.08\linewidth}
				>{\centering}m{0.08\linewidth}
				>{\centering}m{0.08\linewidth}|
				>{\centering}m{0.08\linewidth}|
				>{\centering}m{0.11\linewidth}}
			\hlinewd{0.8pt}
			\multirow{2}{*}{Methods}  
			&\multicolumn{1}{ c |}{Backbone} 
			&\multicolumn{4}{ c |}{TSS} 
			&\multirow{2}{*}{PF-PA.} \tabularnewline
			\cline{3-6}
			&\multicolumn{1}{ c |}{feature} &FG &JO &PA &Avg. & \tabularnewline
			\hline
			\hline
			CNNGeo~\cite{rocco2017convolutional} &ResNet-101 &90.3 &76.4 &56.5 &74.3 &69.5 \tabularnewline
			PARN~\cite{jeon2018parn} &VGG-16 &87.6 &71.6 &68.8 &76.0 &49.1  \tabularnewline
			NC-Net~\cite{rocco2018neighbourhood} &ResNet-101 &94.5 &81.4 &57.1 &77.7 &78.9 \tabularnewline
			SAM-Net~\cite{kim2019semantic} &VGG-16 &94.4 &75.5 &78.3 &82.7 &80.2 \tabularnewline
			SCOT~\cite{liu2020semantic} &ResNet-101 &95.3 &81.3 &57.7 &78.1 &88.8 \tabularnewline
			GLU-Net~\cite{truong2020glu} &VGG-16 &93.2 &73.3 &71.1 &79.2 &79.7 \tabularnewline
			GOCor~\cite{truong2020gocor} &VGG-16 &95.0 &78.9 &81.3 &85.1 &- \tabularnewline
			\hline
			\multirow{2}{*}{\textbf{DMP}} &VGG-16 &93.9 &77.2 &72.1 &81.7 &81.5  \tabularnewline
			&ResNet-101  &94.6 &78.1 &74.2 &82.3 &85.3 \tabularnewline
			\hdashline
			\multirow{2}{*}{\textbf{A-DMP}} &VGG-16 &94.1 &80.1 &74.3 &82.8 &82.4 \tabularnewline
			& ResNet-101  &94.4 &81.4 &75.2 &86.7 &85.4\tabularnewline
			\hdashline
			\multirow{2}{*}{\textbf{DMP}$\dagger$} &VGG-16 &96.3 &83.1 &80.1 &86.5 &86.2 \tabularnewline
			& ResNet-101 &\bf{96.7} &\bf{89.2} &\bf{82.3} &\bf{89.4} &\bf{89.1} \tabularnewline
			\hlinewd{0.8pt}
		\end{tabular}
	}
	\vspace{+5pt}
	\caption{\textbf{Quantitative evaluation on TSS~\cite{taniai2016joint} and PF-PASCAL (PF-PA.)~\cite{ham2016proposal} benchmark.} Higher PCK is better. \emph{FG: FG3DCar, JO: JODS, PA: PASCAL datasets.}}\label{tab:2}\vspace{-5pt}
\end{table}

\paragraph{ETH3D.}
We further evaluate our method on ETH3D~\cite{schops2017multi} dataset. Unlike Hpatches~\cite{balntas2017hpatches}, ETH3D consists of real 3D scenes, where the image transformations are not constrained to homography transformation, thus more challenging. We follow the protocol of~\cite{truong2020glu}, where we sample the image pairs at different intervals to evaluate on varying magnitude of geometric transformations. We evaluate on 7 intervals in total, each interval containing approximately 500 image pairs, and employ the standard evaluation metrics, AEE and PCK. \figref{img:7} and \figref{img:8} show quantitative and quantitative results. Our methods yield highly competitive results, and DMP$\dagger$, significantly outperforming other methods, attains state-of-the-art performance .

\subsection{Semantic Matching Results}\label{sec:4_4}
\begin{figure}[t]
	\centering
	\renewcommand{\thesubfigure}{}
	\subfigure[]
	{\includegraphics[width=0.25\linewidth]{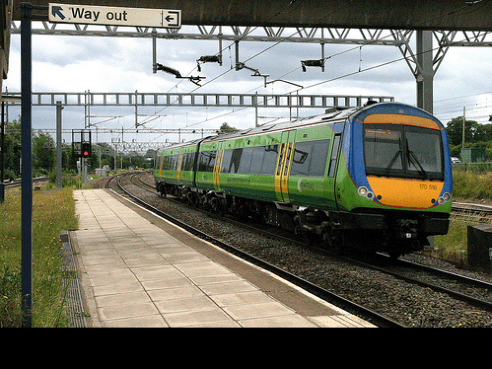}}\hfill
	\subfigure[]
	{\includegraphics[width=0.25\linewidth]{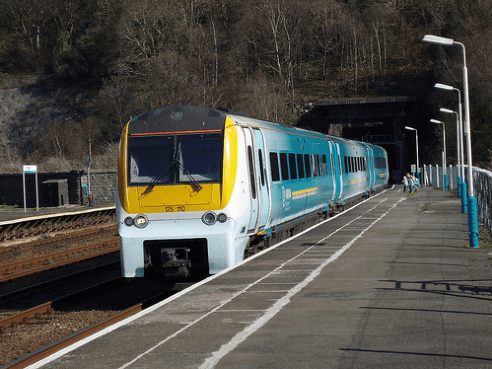}}\hfill
	\subfigure[]
	{\includegraphics[width=0.25\linewidth]{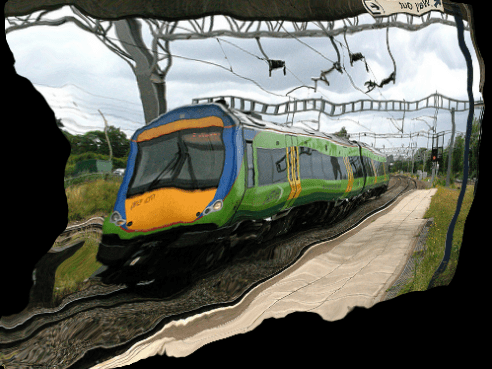}}\hfill
	\subfigure[]
	{\includegraphics[width=0.25\linewidth]{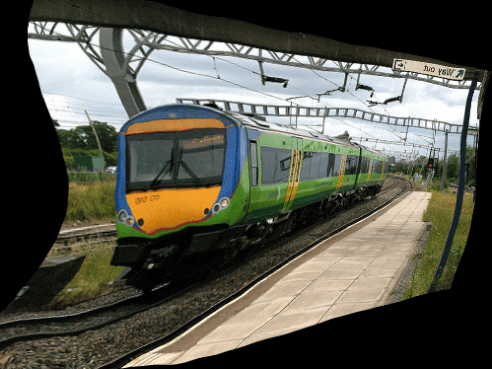}}\hfill\\
	\vspace{-21.5pt}
	\subfigure[(a) Source]
	{\includegraphics[width=0.25\linewidth]{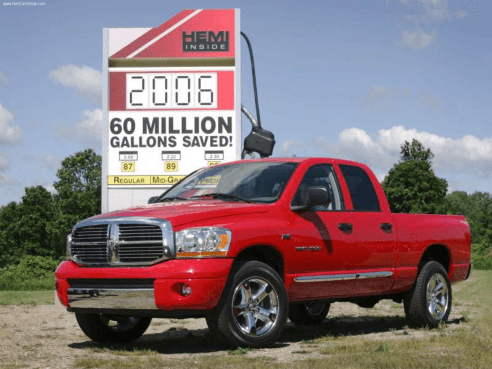}}\hfill
	\subfigure[(b) Target]
	{\includegraphics[width=0.25\linewidth]{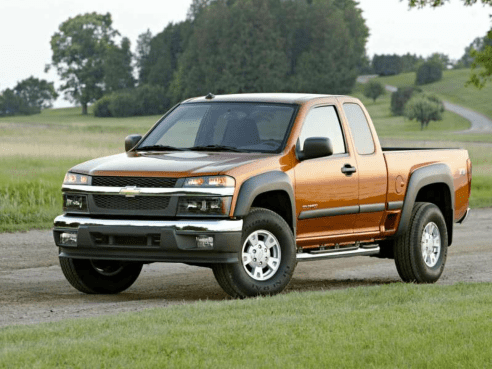}}\hfill
	\subfigure[(c) GLU-Net~\cite{truong2020glu}]
	{\includegraphics[width=0.25\linewidth]{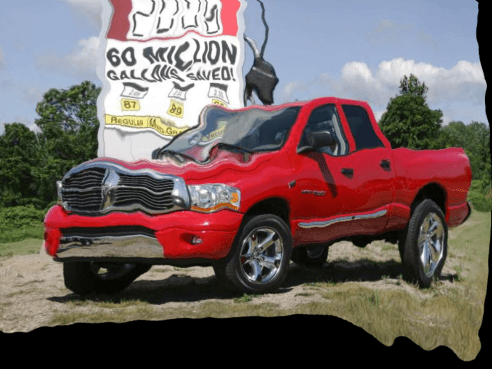}}\hfill
	\subfigure[(d) DMP]
	{\includegraphics[width=0.25\linewidth]{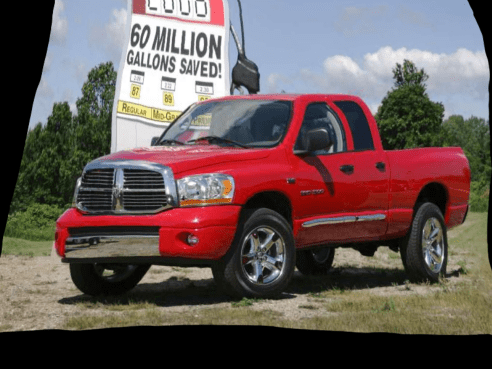}}\hfill\\
	\caption{\textbf{Qualitative results on TSS~\cite{taniai2016joint} benchmark.}}\label{img:9}\vspace{-10pt}
\end{figure}

\paragraph{TSS.}
For semantic matching, we evaluate our method on TSS~\cite{taniai2016joint} benchmark, which contains 3 groups (FG3DCar, JODS, and PASCAL) of 400 image pairs of 7 object categories, foreground masks and dense ground-truth flow. We employ the PCK to measure the precision. For TSS, we set the threshold as $\alpha$ = 0.05. Following~\cite{truong2020glu}, we address the reflections by inferring the flow on both the source and flipped target and the source and original target. We then output the flow field with smaller mean horizontal magnitude. \tabref{tab:2} summarizes the PCK values. \figref{img:9} visualizes the qualitative comparisons. As shown in the results, our method produces highly competitive results compared to other learning-based methods. Interestingly, DMP$\dagger$ already attains state-of-the-art performance with VGG-16~\cite{simonyan2014very} as feature backbone, but using the ResNet-101~\cite{he2016deep} further boosts the performance.
\vspace{-10pt}

\begin{figure}[t]
	\centering
	\renewcommand{\thesubfigure}{}
	\subfigure[(a) Source]
	{\includegraphics[width=0.248\linewidth]{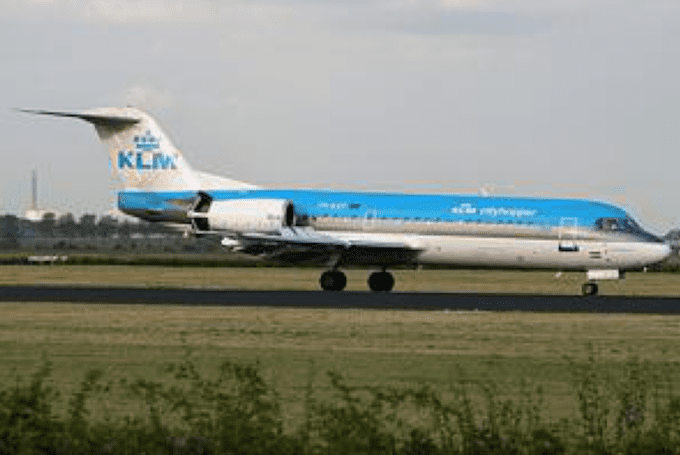}}\hfill
	\subfigure[(b) Target]
	{\includegraphics[width=0.248\linewidth]{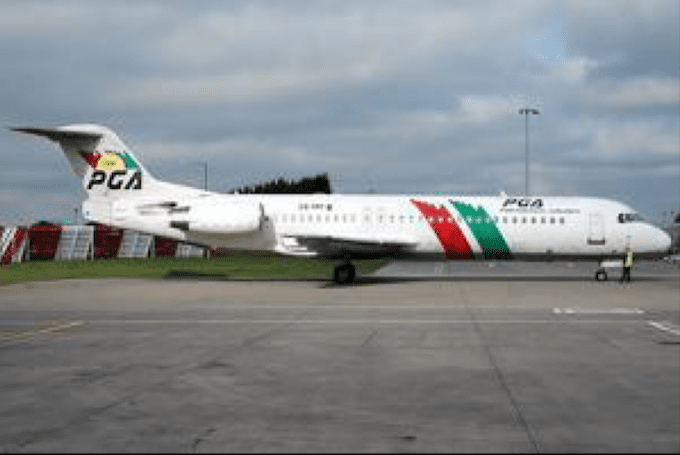}}\hfill
	\subfigure[(c) NC-Net~\cite{rocco2018neighbourhood}]
	{\includegraphics[width=0.25\linewidth]{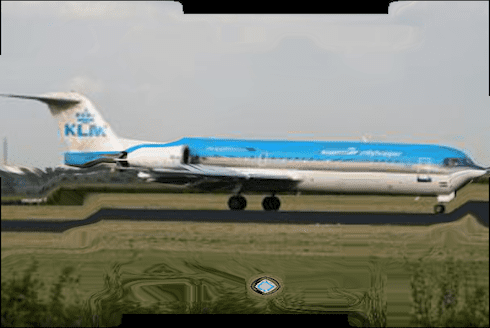}}\hfill
	\subfigure[(d) GLU-Net~\cite{truong2020glu}]
	{\includegraphics[width=0.252\linewidth]{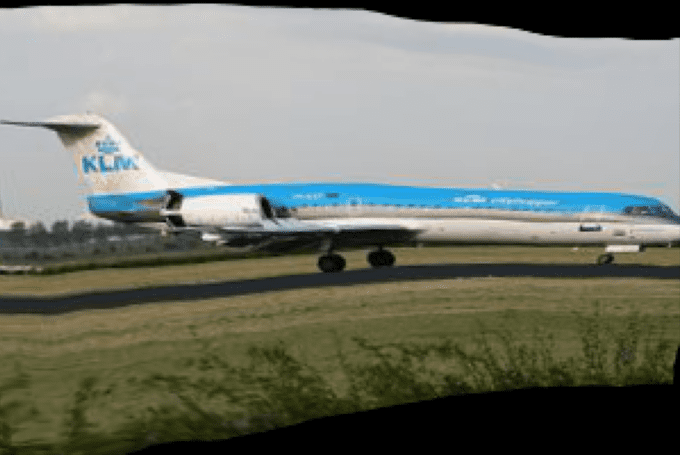}}\hfill\\
	\vspace{-5pt}
	\subfigure[(e) DMP]
	{\includegraphics[width=0.25\linewidth]{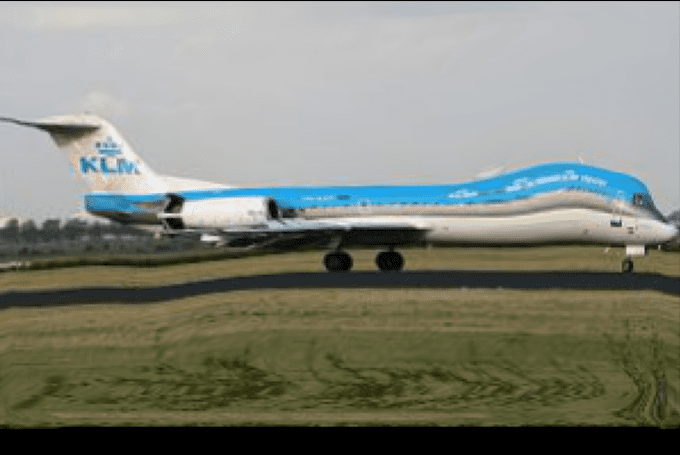}}\hfill
	\subfigure[(f) A-DMP]
	{\includegraphics[width=0.25\linewidth]{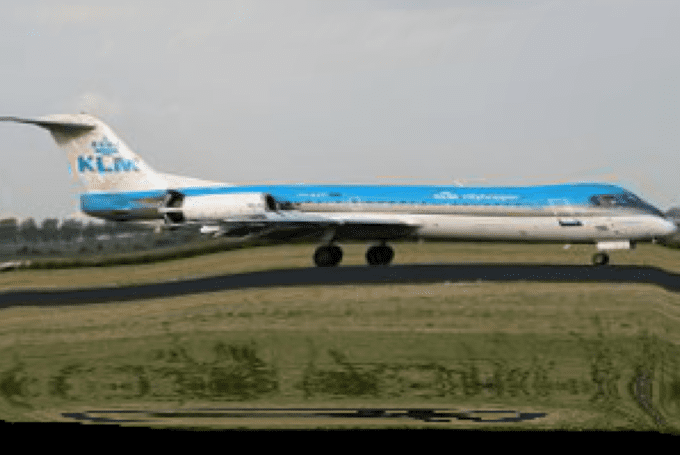}}\hfill
	\subfigure[(g) DMP$\dagger$-VGG]
	{\includegraphics[width=0.25\linewidth]{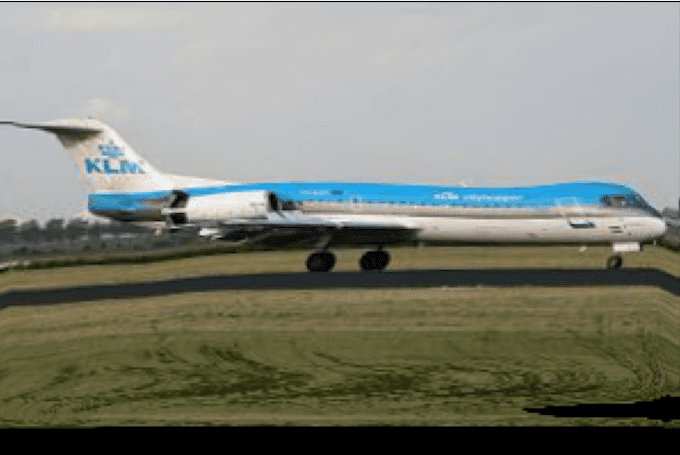}}\hfill
	\subfigure[(h) DMP$\dagger$-ResN.]
	{\includegraphics[width=0.25\linewidth]{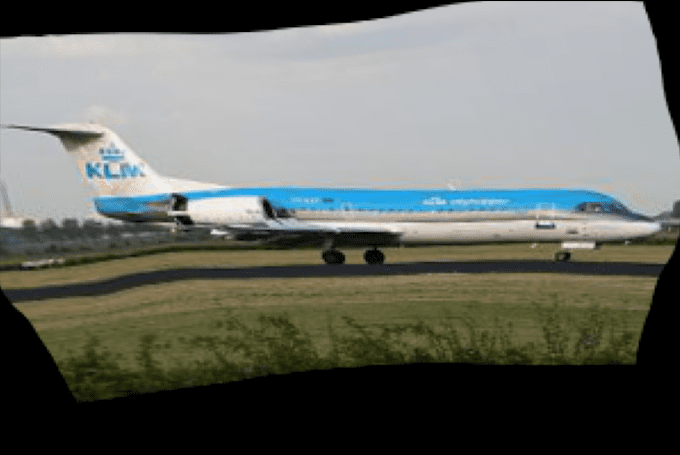}}\hfill\\
    \caption{\textbf{Qualitative results on PF-PASCAL~\cite{ham2016proposal} benchmark.}}\vspace{-10pt}\label{img:10}
\end{figure}

\paragraph{PF-PASCAL.}
We further evaluate on PF-PASCAL~\cite{ham2016proposal} dataset, containing 1,351 image pairs over 20 object categories with keypoint annotations. Following the common practice~\cite{rocco2018neighbourhood,han2017scnet}, we use PCK for the evaluation metric, and the results are reported with the PCK threshold $\alpha=$ 0.1. 
\tabref{tab:2} shows the quantitative results and \figref{img:10} visualizes the qualitative comparisons. Similar to the experiments on TSS~\cite{taniai2016joint}, our DMP provides the highest matching accuracy. 

\subsection{Ablation Study}\label{sec:4_5}
We show an ablation analysis on both architecture and loss function in our model. We measure the AEE over all the scenes of the HPatches~\cite{balntas2017hpatches}, and each ablation experiment is conducted under the same experimental setting. We further report an analysis on computational complexity. 
\vspace{-10pt}

\paragraph{Network architecture.}
\tabref{tab:3} shows the comparisons with different architectural components in our model. The baseline only consists backbone feature and inference module, optimized using contrastive loss without confidence threshold. Since optimization-based model is highly sensitive to initialization, exploiting residual correspondence module in matching networks dramatically helps the convergence and boosts performance in comparison to the baseline. We found that jointly exploiting both adaptation feature and residual correspondence showed apparent improvements. In addition, our model was not sensitive to the number of samples chosen for the loss computation. 
\vspace{-10pt}

\begin{figure}[t]
\centering
\includegraphics[width=0.8\linewidth]{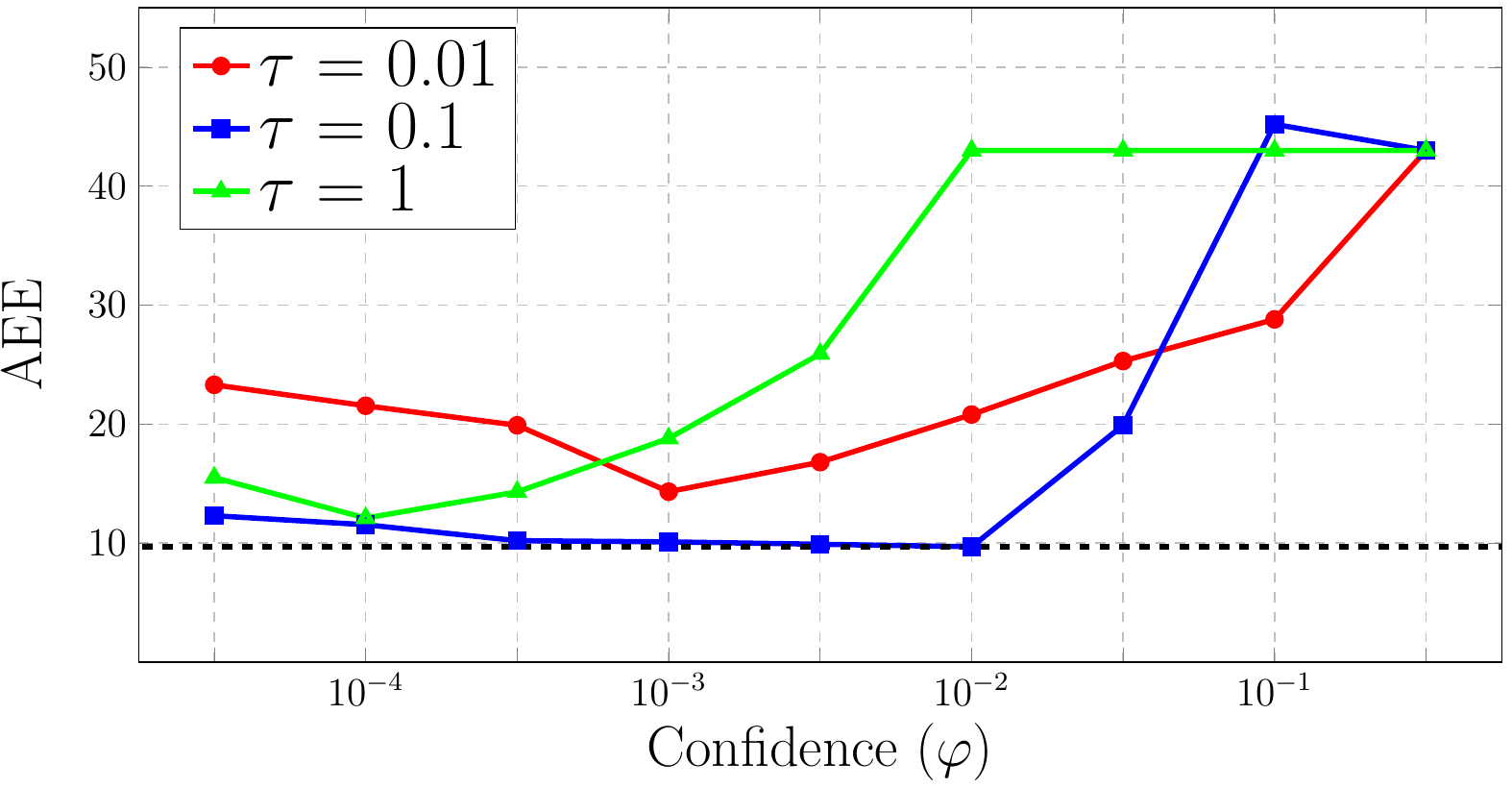}
\caption{\textbf{Ablation study on temperature and confidence.}}
\label{ablation_loss}\vspace{-10pt}\label{img:11}
\end{figure}

\paragraph{Loss function.}
As shown in~\tabref{tab:3}, the performance improvement by confidence threshold was apparent. In addition, \figref{ablation_loss} plots the accuracy as varying both the temperature ($\tau$) and confidence ($\varphi$). We found that the temperature controlling the sharpness of the softmax function is highly influential to the convergence and performance of our model. We thus found from extensive experiments that setting the temperature and confidence as 0.1 and 0.01, respectively, reports the best results. If the confidence is set too high, the loss signal does not occur and learning stops.  
\vspace{-10pt}

\paragraph{Computational complexity.}
In experiments, we found that given good initialization, DMP, including all the variants, converges within 100 iterations. 
However, given difficult image pairs to be matched, DMP struggles to find the good correspondences, and we thus let it iterate 2k times. The average runtime of 100 iterations of DMP and DMP$\dagger$ is 3-5 seconds, while A-DMP is 6-10 seconds on a single GPU Geforce GTX 2080 Ti. More details can be found in supplementary material. A natural next step, which we leave for future work, is to reduce the runtime for the optimization by advanced techniques, e.g., early-stopping. 

\begin{table}[]
\centering
\scalebox{0.7}{
\begin{tabular}{>{\centering}m{0.2\linewidth}|
                >{\centering}m{0.2\linewidth}|
				>{\centering}m{0.2\linewidth}|
				>{\centering}m{0.2\linewidth}|
				>{\centering}m{0.15\linewidth}}
\hlinewd{0.8pt}
$\#$ samples &Adaptation & Residual & Confidence & \multirow{2}{*}{AEE} \tabularnewline
($M$) &feature & correspond. & ($\varphi$) & \tabularnewline
\hline
\hline
$256$ &\xmark & \xmark &\xmark &33.1     \tabularnewline
\hline
$256$ &\cmark & \xmark &\xmark &32.8     \tabularnewline
$256$ &\xmark & \cmark &\xmark &18.6     \tabularnewline
$256$ &\cmark & \cmark &\xmark & 12.3    \tabularnewline
\hline
$128$ &\cmark & \cmark &\cmark & 9.84    \tabularnewline
$256$ &\cmark & \cmark &\cmark & \bf{9.69}    \tabularnewline
$512$ &\cmark & \cmark &\cmark & 9.67    \tabularnewline
\hlinewd{0.8pt}
\end{tabular}}
\vspace{+5pt}
\caption{\textbf{Ablation study of DMP.}}\label{tab:3}\vspace{-5pt}
\end{table}

\section{Conclusion}
In this paper, we have proven that, for the first time, a matching prior must not necessarily be learned from large training data and have introduced DMP that captures an image pair-specific matching prior by optimizing the untrained matching networks on a single pair of images. Tailored for such test-time optimization for matching, we have developed a residual matching network and a confidence-aware contrastive loss. Our experiments have shown that 
although our framework requires neither a large training data nor an intensive learning, it is competitive against the latest learning-based methods and even outperforms.  

\newpage
\textbf{Acknowledgments} This research was supported by the MSIT, Korea, under the ICT Creative Consilience program (IITP-2021-2020-0-01819) and Regional Strategic
Industry Convergence Security Core Talent Training Business (IITP-2019-0-01343) supervised by the IITP and National Research Foundation of Korea (NRF-2021R1C1C1006897).

{\small
\bibliographystyle{ieee_fullname}
\bibliography{egbib}
}
\newpage
\clearpage
\appendix
\begin{center}
	\textbf{\Large Appendix}
\end{center}
\renewcommand\thesection{\Alph{section}}
\renewcommand{\thefigure}{\arabic{figure}}
\renewcommand{\theHfigure}{A\arabic{figure}}
\setcounter{figure}{0}
In this section, we provide more details of DMP and more results on the Hpatches~\cite{balntas2017hpatches}, ETH3D~\cite{schops2017multi}, TSS~\cite{taniai2016joint}, and PF-PASCAL~\cite{ham2016proposal}.

\begin{figure*}[t!]
	\centering
	\renewcommand{\thesubfigure}{}
	\subfigure[]
	{\includegraphics[width=1\linewidth]{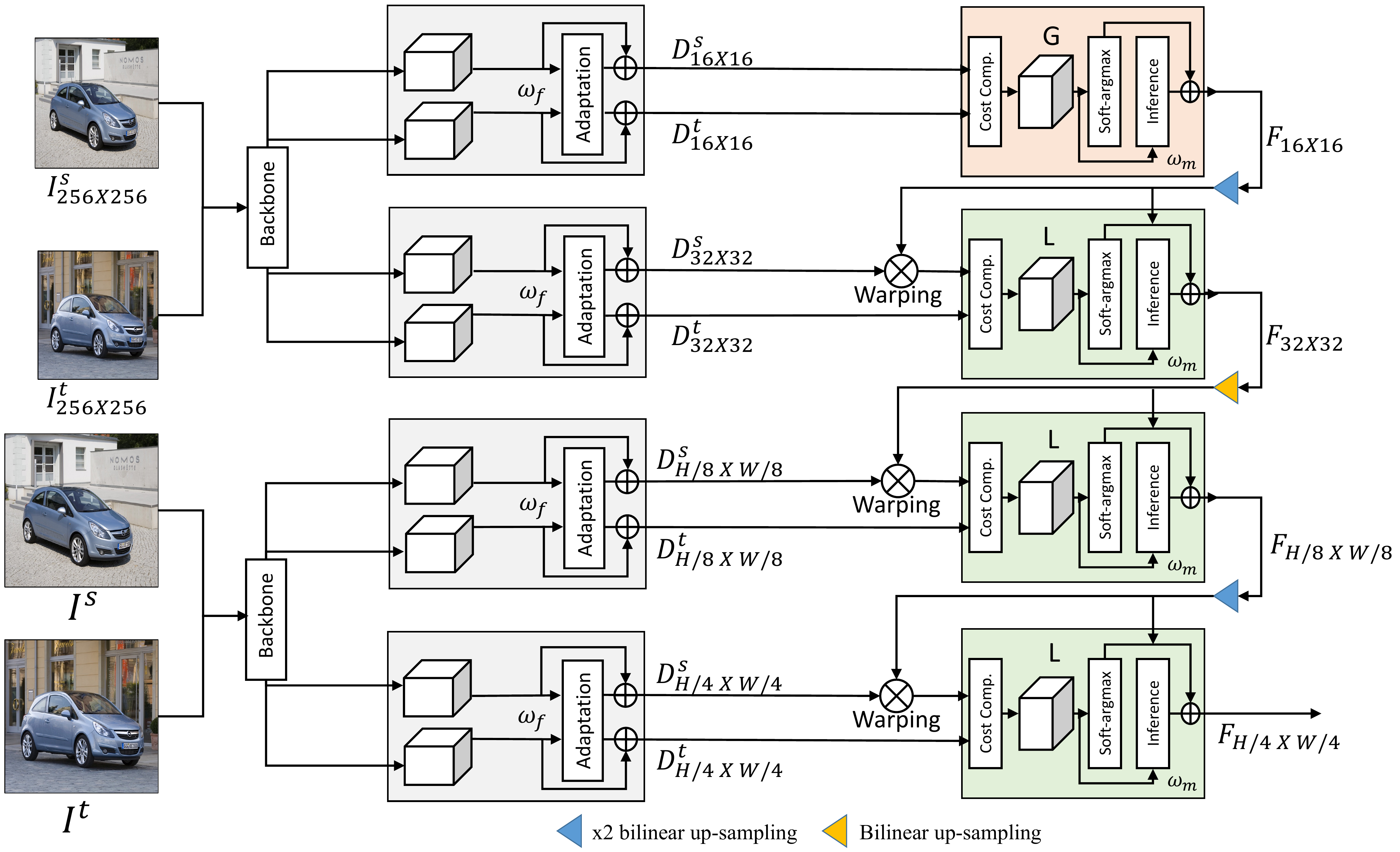}}\\
	\vspace{-10pt}
\caption{\textbf{Overview of DMP architecture.} Overview of our proposed iterative architecture, which consists of feature extraction network and matching network. Source and target images are first fed into feature backbone network to obtain deep features. Each pyramidal features are then fed into adaptation layers and the refined features are obtained. Subsequently, the refined features are fed into a matching network and the estimated flow is up-sampled to warp the next level feature. The final output consists of refined features from target image and the flow field of size $\frac{H}{4} \times \frac{W}{4}$. }	\label{FIG:1}
\end{figure*}

\section{Network Architecture}\label{sec:1}
As shown in~\figref{FIG:1}, our network consists of two parts: {\it feature extraction} networks to extract deep features and {\it matching} networks. 
Note that in the paper, a single-level version of the networks are
illustrated for brevity, while the full model is formulated in a pyramidal fashion. In this section, we will explain our full model. \vspace{-10pt}

\paragraph{Feature extraction network.} Here, we explain feature extraction networks in detail. We employ an adaptive resolution startegy introduced by~\cite{truong2020glu} to let the network take any resolution, which we down-sample the original input images to $256 \times 256$.  We then extract features from both the original and down-sampled resolutions using pre-trained backbone networks and freeze them during the optimization and training. After the feature extraction, we additionally use an adaptation layer to refine the features. Adaptation layers are random initialized and separated for each pyramidal feature map in a residual fashion~\cite{he2016deep}. 
As described in the paper, the backbone model could be directly optimized in DMP framework, but we found that using an additional
adaptation layer to refine the backbone features and optimizing the layer only boosts the performance drastically.

Specifically, we compute the residuals by adding 3 $\times$ 3, 5 $\times$ 5, 7 $\times$ 7 and 9 $\times$ 9 convolutional layers with padding of 1, 2, 3 and 4, respectively, on top of each pyramidal level. We set stride to 1 to ensure that the spatial resolution is preserved. Given VGG-16 as backbone network, identical to~\cite{truong2020glu}, we employ the activation after \verb|Conv5-3| and \verb|Conv4-3| for the resized (256 $\times$ 256) input images, and \verb|Conv4-3| and \verb|Conv3-3| for the original resolution image, which outputs spatial resolution of $16 \times 16$, $32 \times 32$, $\frac{H}{8} \times \frac{W}{8}$ and $\frac{H}{4} \times \frac{W}{4}$, respectively. The number of feature channels of each adaptaion layer are thus 512, 256, 256, and 128, respectively. On the contrary, if ResNet-101 is used as backbone network, we employ the activation after \verb|Conv3| and \verb|Conv4| for the resized input, whereas for the original resolution we employ \verb|Conv2| and \verb|Conv3|. The number of feature channels of each adaptaion layer are thus 1024, 512, 512, and 256, respectively. 
It should be noted that for geometric matching task, we use VGG features, while for semantic matching task, we use both ResNet and VGG, which by default, unless mentioned, all our models use VGG-16 features. 
\vspace{-10pt}
	
\paragraph{Matching network.} We then provide additional details of matching networks, which consists of two parts: cost computation and inference modules. For the global correlation, we compute the pairwise inner product between features from coarsest level. For the local correlation, we employ $l=4$ for the search space in the target. As in~\cite{melekhov2019dgc,truong2020glu} we feed global correlation into a inference module, which consists of 5 feed-forward convolutional blocks with a 3 $\times$ 3 filter. The number of output channels of each layers are 128, 128, 96, 64, and 32, respectively. For the remaining levels, we use an inference module designed for the local correlation volume which infers the flow field similar to the one in PWC-Net~\cite{sun2018pwc}. The numbers of output channels at each layer are 128, 128, 96, 64, and 32, respectively and the size spatial kernel of is also 3 $\times$ 3. The final output of the inference module is computed by feeding into a linear 2D convolution. The soft-argmax~\cite{kendall2017end} computes an output by averaging all the spatial positions with weighted corresponding probabilities. The temperature for the soft-argmax is set to 0.02. 

The flow field inferred at each level is up-sampled using bilinear interpolation. From experiments, we observed that using transposed convolution degraded the performance. We thus employed bilinear interpolation at every pyramidal layer.  
\begin{figure}
	\centering
\includegraphics[width=0.9\linewidth]{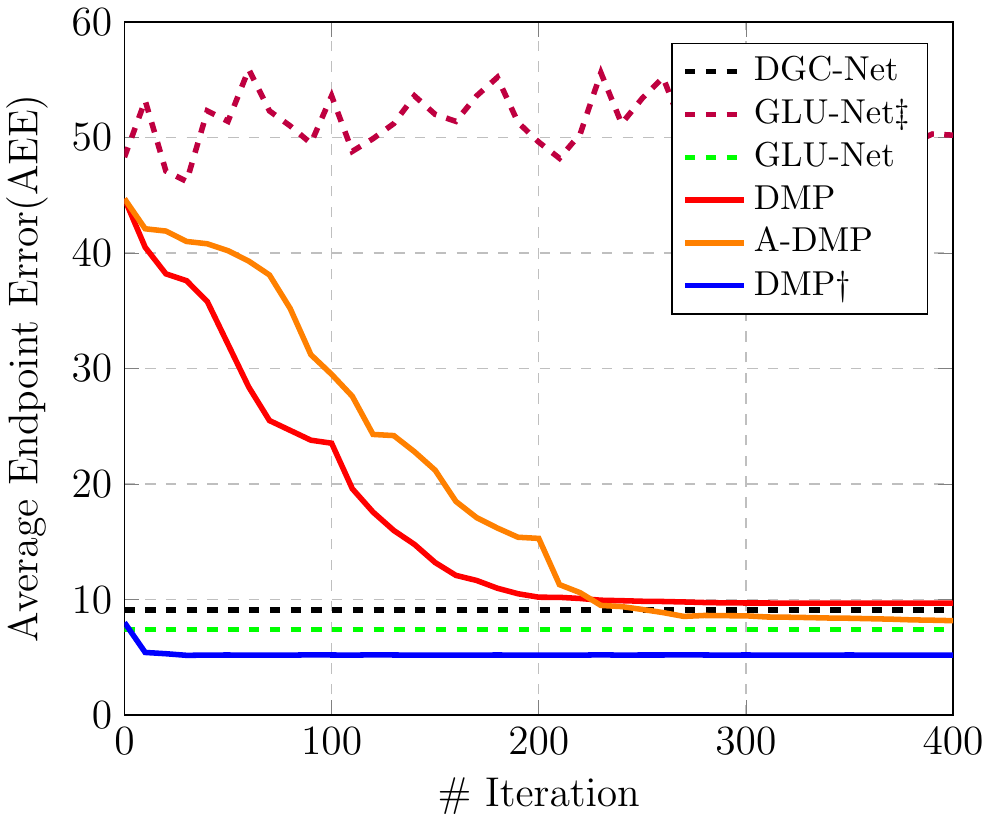}
  \caption{\textbf{Convergence analysis of DMP.}}\vspace{-10pt}\label{fig2:supp}
\end{figure}

\section{Convergence Analysis Details}\label{sec:2}
In the paper, we showed the comparison of AEE over iteration between models as shown in~\figref{fig2:supp} (Fig. 4 in the paper). 
Here, we describe the details for this experiment. For a fair comparison, we iterated 2k times for all the test-time optimization methods, which include GLU-Net$\ddagger$, DMP, A-DMP and DMP$\dagger$. We evaluated each method on Hpatches~\cite{balntas2017hpatches} benchmark, which consists of 295 target images, and averaged the AEE at every 10-th iteration. Note that in~\figref{fig2:supp}, we only show the range of 0-400 for x-axis as the AEE for all the methods except GLU-Net$\ddagger$ converge. To conduct experiment on GLU-Net$\ddagger$, we simply replaced the model to GLU-Net within our optimization implementation under the identical experimental setting to DMP test-time optimization. We did not find noticeable differences when we attempted optimizing with different hyperparameter settings e.g., learning rate. We conducted experiment on GLU-Net$\ddagger$ to show that the several choices we made, including architecture and loss, were critical for the untrained network to guarantee a meaningful convergence.

\section{Limitations}\label{limit}
In this session, we would like to discuss the limitations of DMP and its variants. One limitation that all the methods, including DMP and its variants, is the time they take to converge. Although with good initialization, the optimization time required to obtain correspondences significantly reduces, our approach fundamentally isn't applicable for real-time applications. Furthermore, even though DMP attained competitive results for standard benchmarks by optimizing from untrained networks, it fails to find accurate correspondences given difficult images, i.e., ETH3D interval 15. To overcome, we pre-trained DMP to provide strong initialization, but this may result in weakening of DMP's advantage, an ability to avoid generalization issues. Although designed to address difficult cases, A-DMP suffers from doubled optimization time. RANSAC-DMP successfully avoids this challenge, but the use of RANSAC often yields unstable results that may lead to failure to find correspondences. 

We proposed, for the first time, to find correspondences between a pair of images by test-time optimization, and we believe that further improvements could be made in this direction. 

\section{Implementation and Experimental Details}\label{sec:3}
We first pre-process the input images by centering the mean and normalizing the values using the mean and standard deviation of ImageNet~\cite{deng2009imagenet}. For DMP and A-DMP, we initially set the learning rate to 3$e^{-3}$ and divide it by 2 at every 300 iterations. For DMP$\dagger$ and variants that exploit RANSAC~\cite{shen2020ransac}, we use learning rate of 1$e^{-5}$. We use Adam optimizer~\cite{kingma2014adam} with $\beta_1 = 0.9$ and~$\beta_2=0.999$. We implemented our model using $\texttt{PyTorch}$\cite{paszke2017automatic}. 

\begin{figure}[t]
	\centering
	\renewcommand{\thesubfigure}{}
	\subfigure[(a) Original]
	{\includegraphics[width=0.243\linewidth]{supplefig/188.png}}\hfill
	\subfigure[(b) Affine]
	{\includegraphics[width=0.243\linewidth]{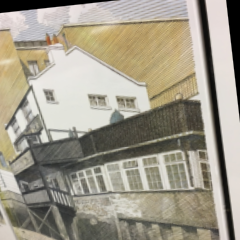}}\hfill
	\subfigure[(c) Homography]
	{\includegraphics[width=0.243\linewidth]{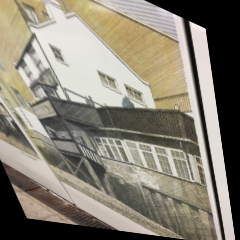}}\hfill
	\subfigure[(d) TPS]
	{\includegraphics[width=0.243\linewidth]{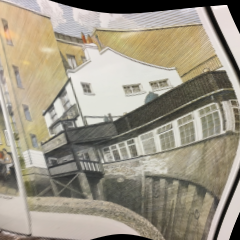}}\hfill\\
	\caption{\textbf{Example of the synthetic images}~\cite{rocco2017convolutional}.\label{synth} }\vspace{-10pt}
\end{figure}

To obtain the randomly-augmented target for A-DMP, we use the same kind of geometric transformation to GLU-Net and DGC-Net. 
Specifically, Rocco et al.~\cite{rocco2017convolutional} generates synthetic data using affine and thin-plate spline transformation which we additionally use homography transformation as in DGC-Net as shown in~\figref{synth}. To conduct experiments on variants of ours that utilize RANSAC to obtain coarsely aligned pair of images, we followed the protocol of~\cite{shen2020ransac} to obtain a pair of coarsely aligned images first and then fed the aligned images into our network.

We additionally showed an ablation study on RANSAC-Flow~\cite{shen2020ransac} in the paper, to validate the effect of test-time optimization. We first obtained coarsely aligned input images and then implemented using the full loss function provided in~\cite{shen2020ransac} for the test-time optimization and iterated 2000 times with identical hyperparameter setting to RANSAC-Flow trained on Mega-Depth~\cite{li2018megadepth}. We did not find drastic diffference when the matchability loss was not included within the total loss. For evaluating test-time optimization of RANSAC-Flow on original resolution of Hpatches, we up-sampled the estimated flow using bilinear interpolation and calculated the AEE and PCK. 
\section{More Results}\label{sec:4}
In this section, we provide additional qualitative examples on the Hpatches~\cite{balntas2017hpatches}, ETH3D~\cite{schops2017multi}, TSS~\cite{taniai2016joint}, and PF-PASCAL~\cite{ham2016proposal}.

We first show more qualitative results of convergence process of DMP. Given good initialization, DMP guarantees a meaningful convergence, which also indicates that once the warped image is similar enough to the target image, the convergence process is boosted and DMP can successfully correct the errors in the flow fields during the optimization to find the optimal flow field. As shown in~\figref{img:3}, the convergence is boosted when the warped image is similar to the target image.

For geometric matching task, DMP shows highly competitive results , nearly approximating the ground-truth flow as shown in~\figref{img:4}. Note that our variants estimate extremely accurate flow fields, demonstrating the superiority of our approach. ~\figref{img:5} shows the qualitative comparison on ETH3D~\cite{schops2017multi} dataset. All the results are from the highest intervals, which demand addressing extreme viewpoint changes. Note that our approaches, compared to GLU-Net~\cite{truong2020glu} which obtains satisfactory results, successfully estimate the correspondence field between images with extreme appearance variations.

Semantic matching task requires estimating correspondence fields between images with intra-class variations. Our works, compared to GLU-Net~\cite{truong2020glu}, consistently obtain sharp and extremely accurate warped images as shown in~\figref{img:6} and~\figref{img:7}. We obtain results with fine details preserved and accurately aligned, which demonstrate the superiority of our approaches on semantic matching task.

\begin{figure*}[t!]
	\centering
	\renewcommand{\thesubfigure}{}
	\subfigure[]
	{\includegraphics[width=0.122\linewidth]{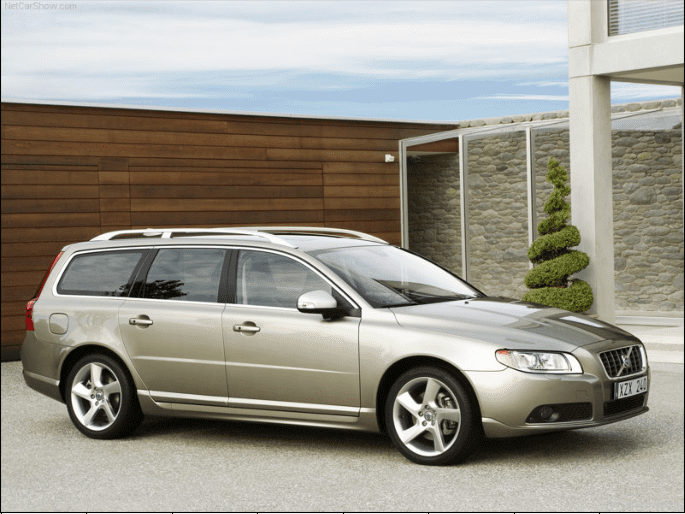}}\hfill
	\subfigure[]
	{\includegraphics[width=0.122\linewidth]{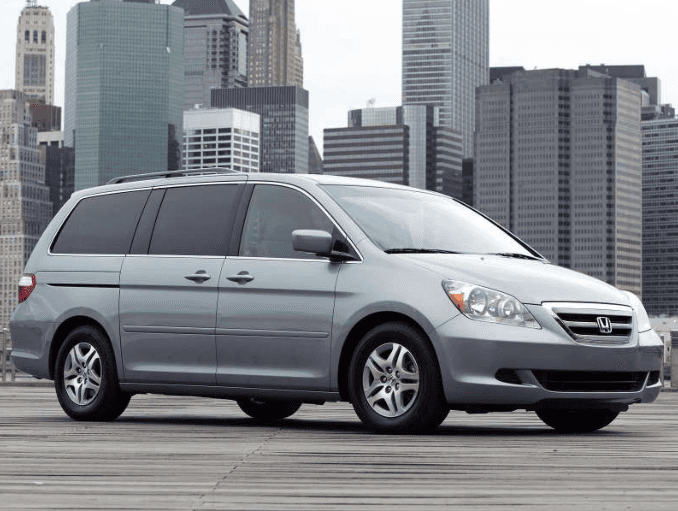}}\hfill
	\subfigure[]
	{\includegraphics[width=0.122\linewidth]{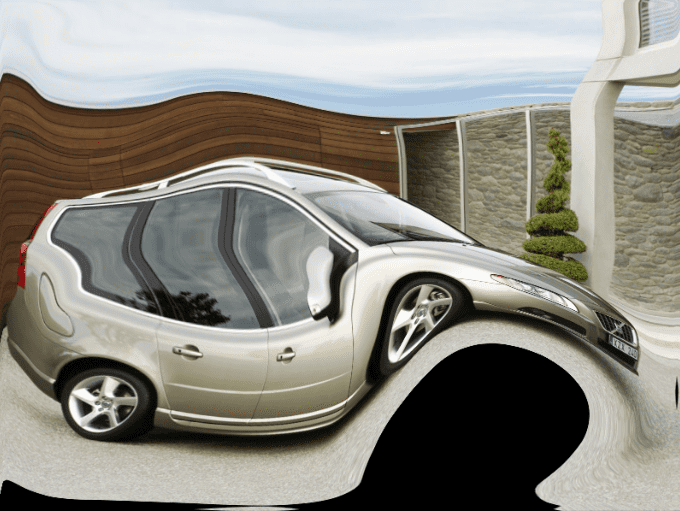}}\hfill
	\subfigure[]
	{\includegraphics[width=0.122\linewidth]{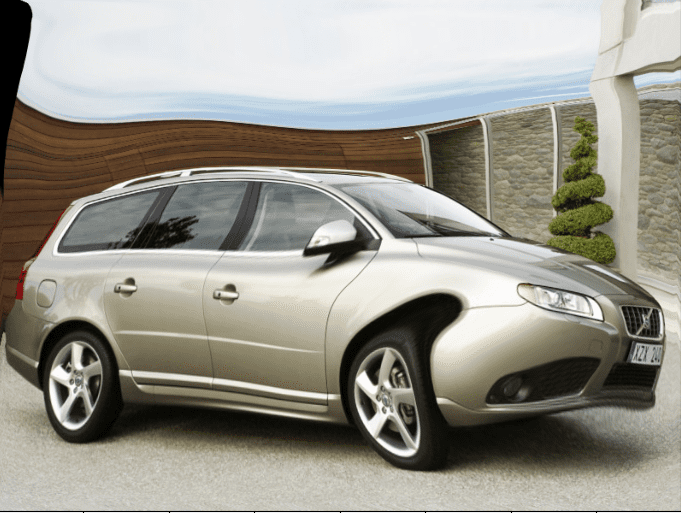}}\hfill
	\subfigure[]
	{\includegraphics[width=0.122\linewidth]{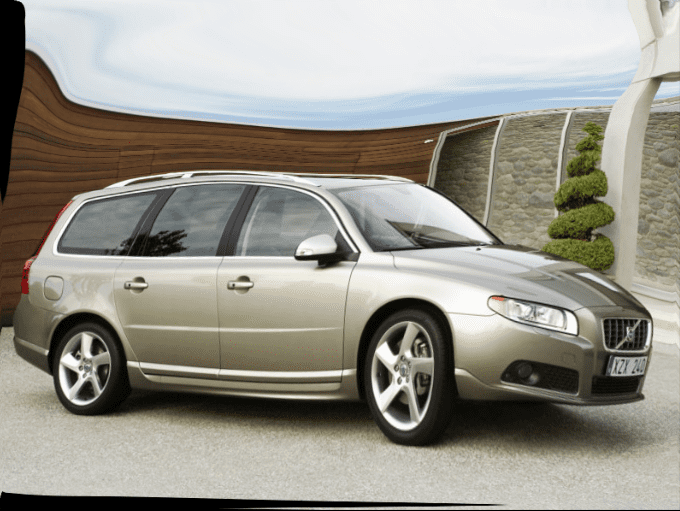}}\hfill
	\subfigure[]
	{\includegraphics[width=0.122\linewidth]{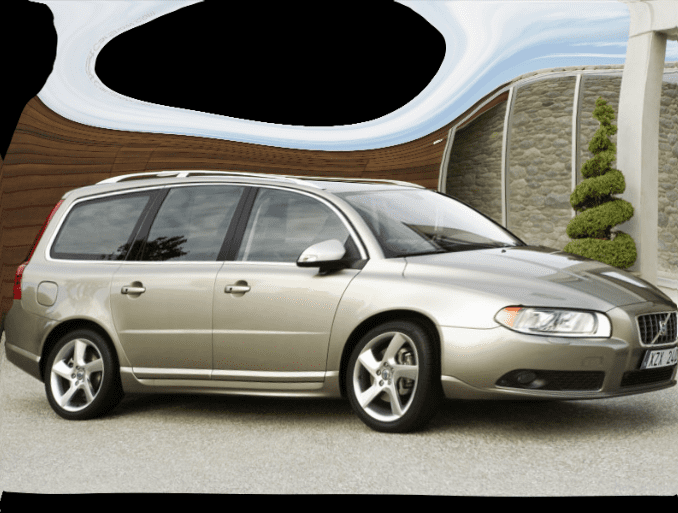}}\hfill
	\subfigure[]
	{\includegraphics[width=0.122\linewidth]{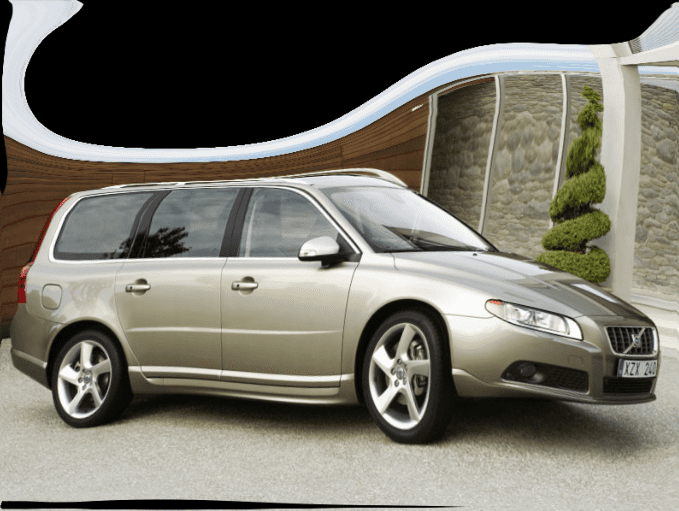}}\hfill
	\subfigure[]
	{\includegraphics[width=0.122\linewidth]{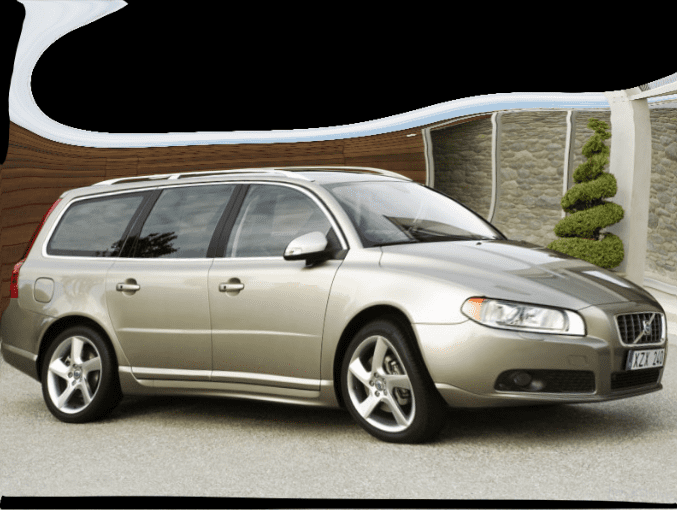}}\hfill\\
	\vspace{-21.5pt}
	\subfigure[]
	{\includegraphics[width=0.122\linewidth]{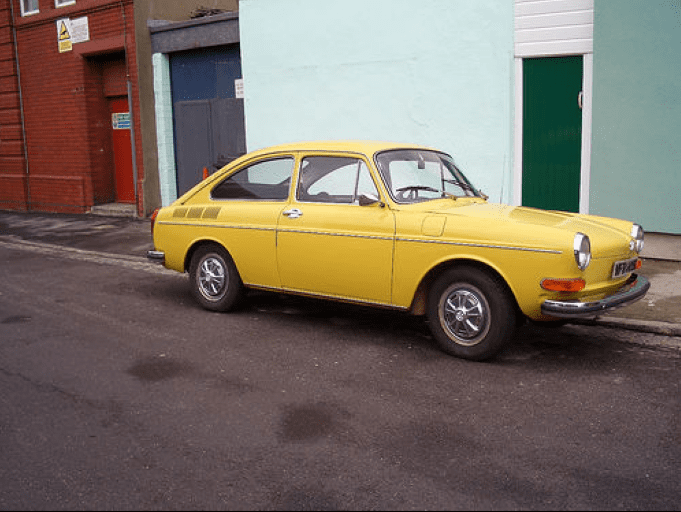}}\hfill
	\subfigure[]
	{\includegraphics[width=0.122\linewidth]{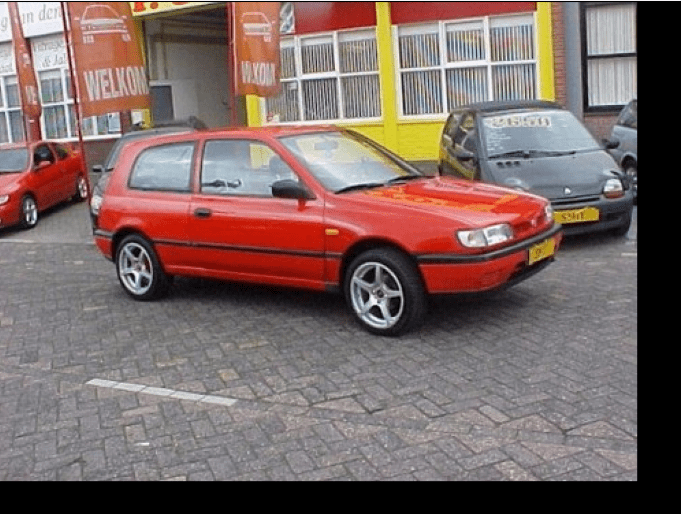}}\hfill
	\subfigure[]
	{\includegraphics[width=0.122\linewidth]{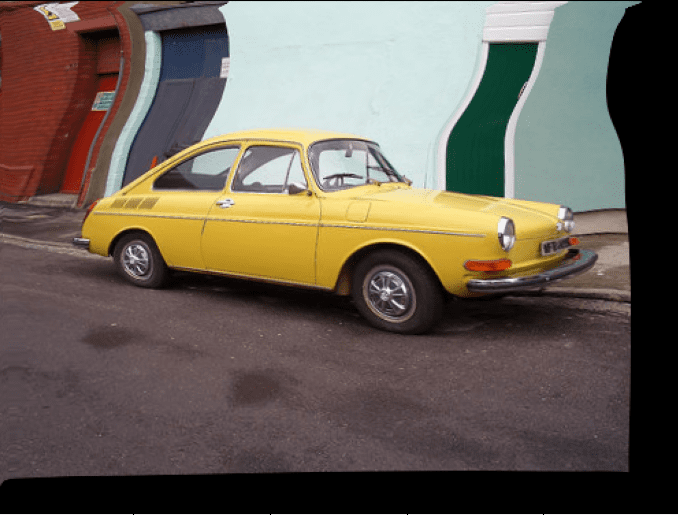}}\hfill
	\subfigure[]
	{\includegraphics[width=0.122\linewidth]{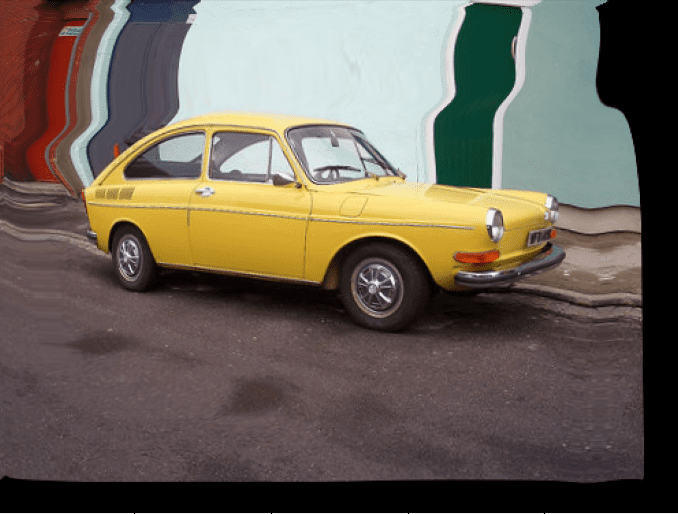}}\hfill
	\subfigure[]
	{\includegraphics[width=0.122\linewidth]{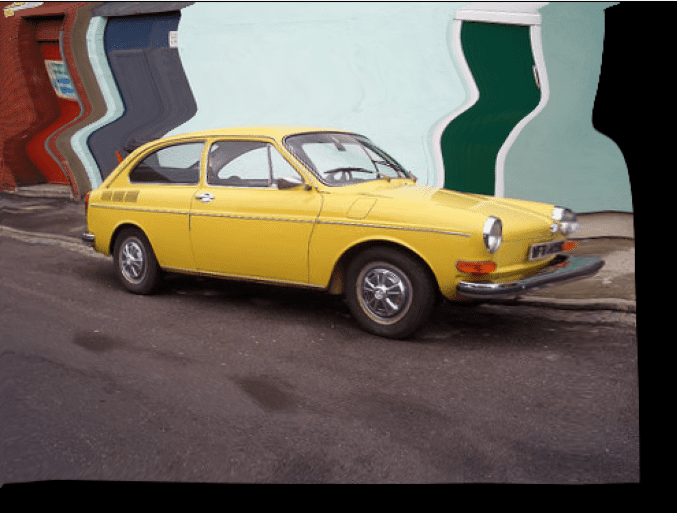}}\hfill
	\subfigure[]
	{\includegraphics[width=0.122\linewidth]{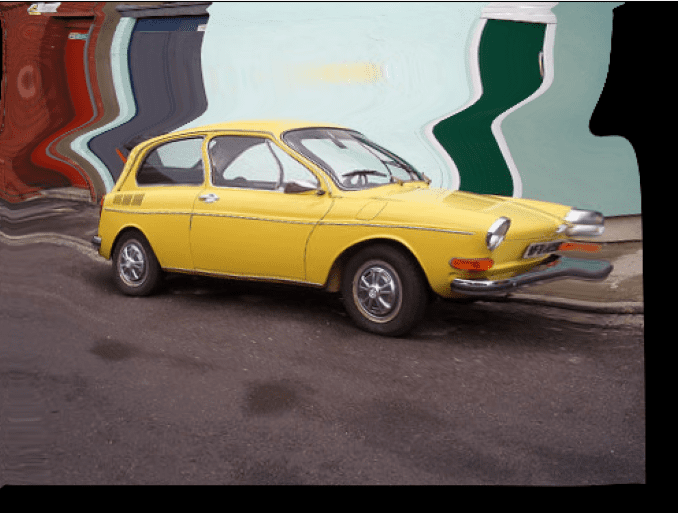}}\hfill
	\subfigure[]
	{\includegraphics[width=0.122\linewidth]{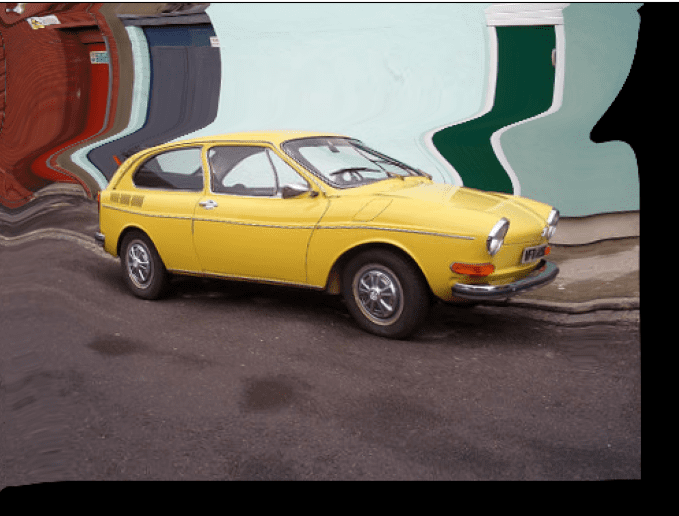}}\hfill
	\subfigure[]
	{\includegraphics[width=0.122\linewidth]{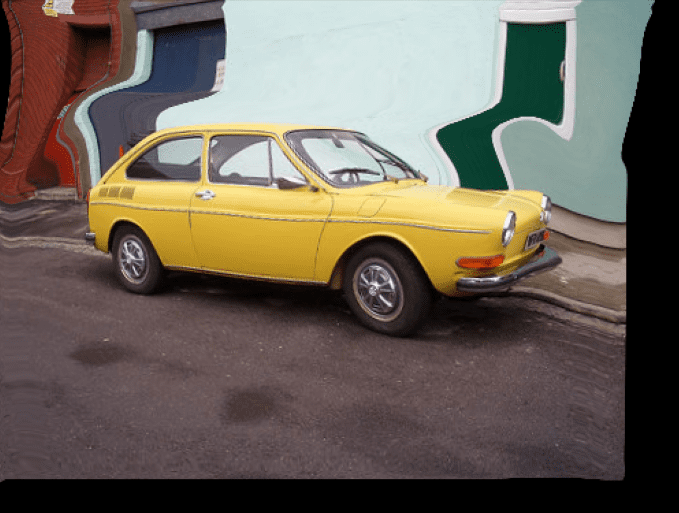}}\hfill\\
	\vspace{-21.5pt}
	\subfigure[]
	{\includegraphics[width=0.122\linewidth]{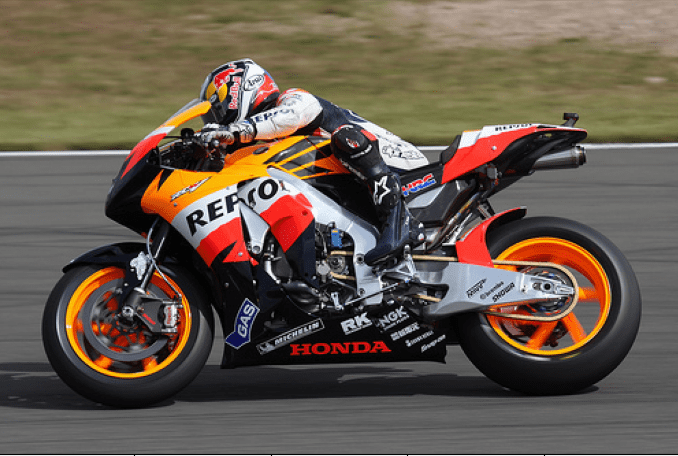}}\hfill
	\subfigure[]
	{\includegraphics[width=0.122\linewidth]{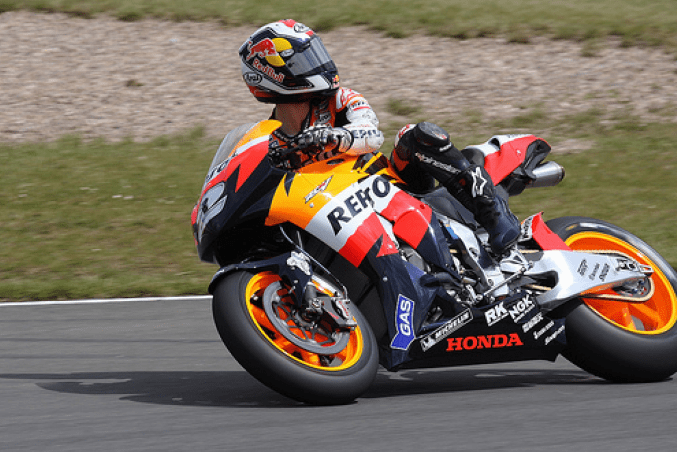}}\hfill
	\subfigure[]
	{\includegraphics[width=0.122\linewidth]{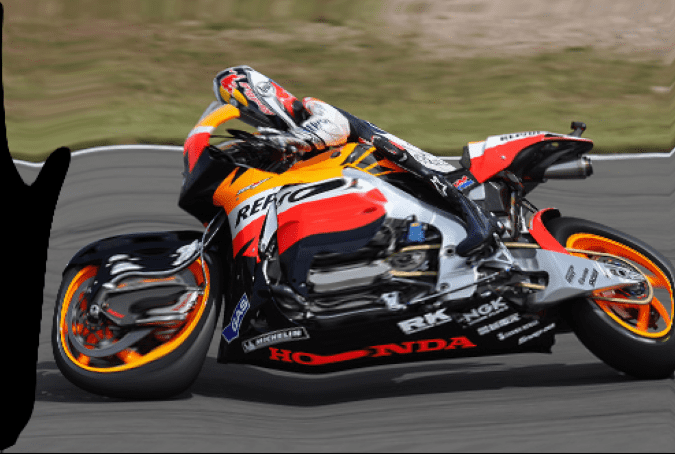}}\hfill
	\subfigure[]
	{\includegraphics[width=0.122\linewidth]{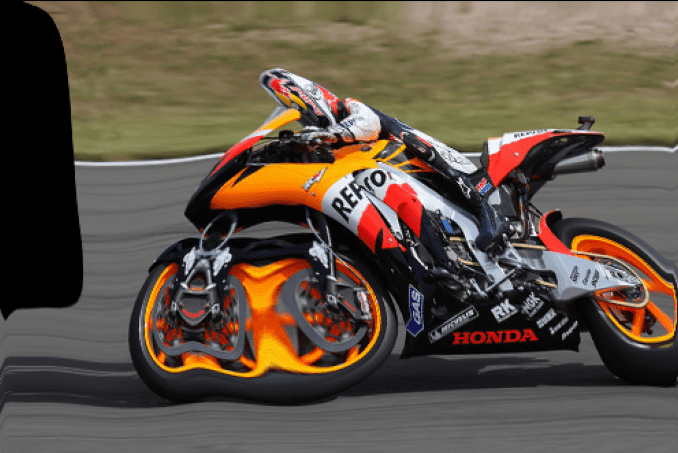}}\hfill
	\subfigure[]
	{\includegraphics[width=0.122\linewidth]{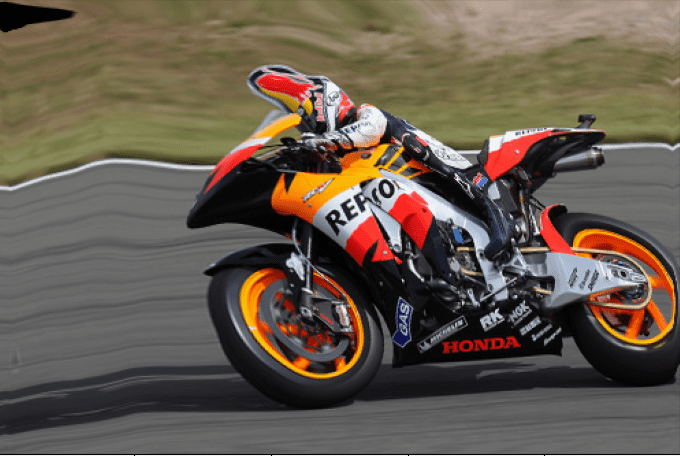}}\hfill
	\subfigure[]
	{\includegraphics[width=0.122\linewidth]{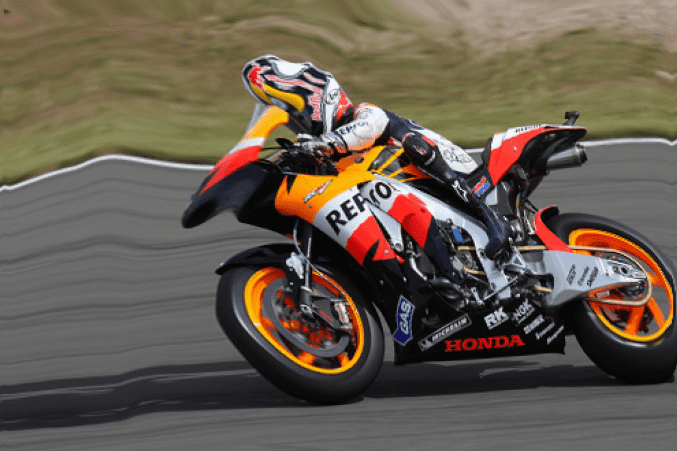}}\hfill
	\subfigure[]
	{\includegraphics[width=0.122\linewidth]{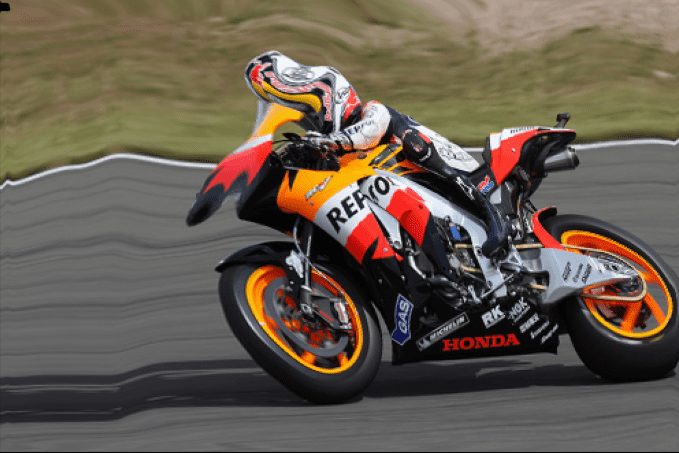}}\hfill
	\subfigure[]
	{\includegraphics[width=0.122\linewidth]{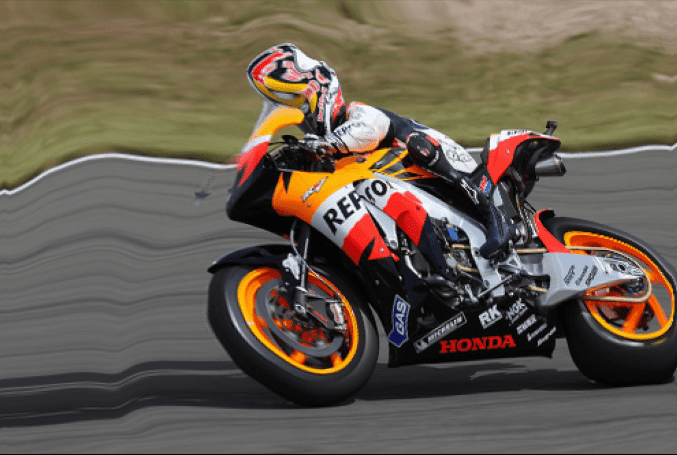}}\hfill\\
	\vspace{-21.5pt}
	\subfigure[]
	{\includegraphics[width=0.122\linewidth]{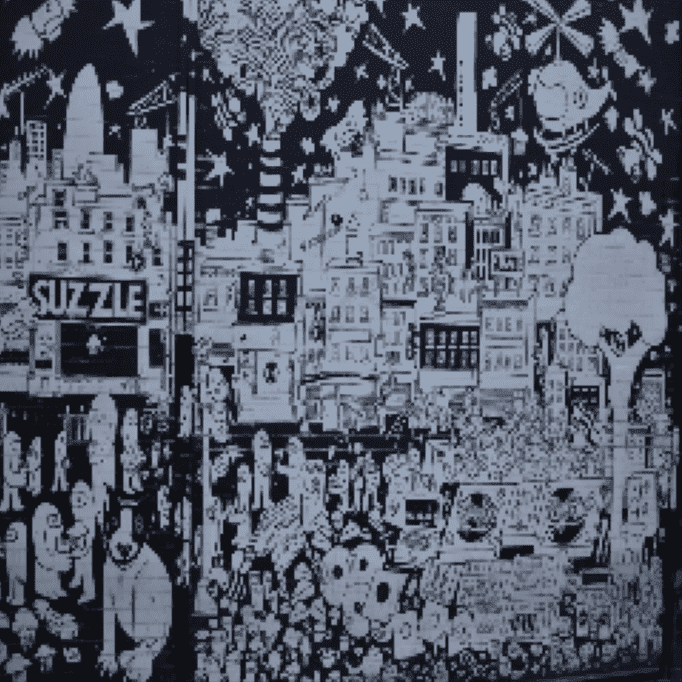}}\hfill
	\subfigure[]
	{\includegraphics[width=0.122\linewidth]{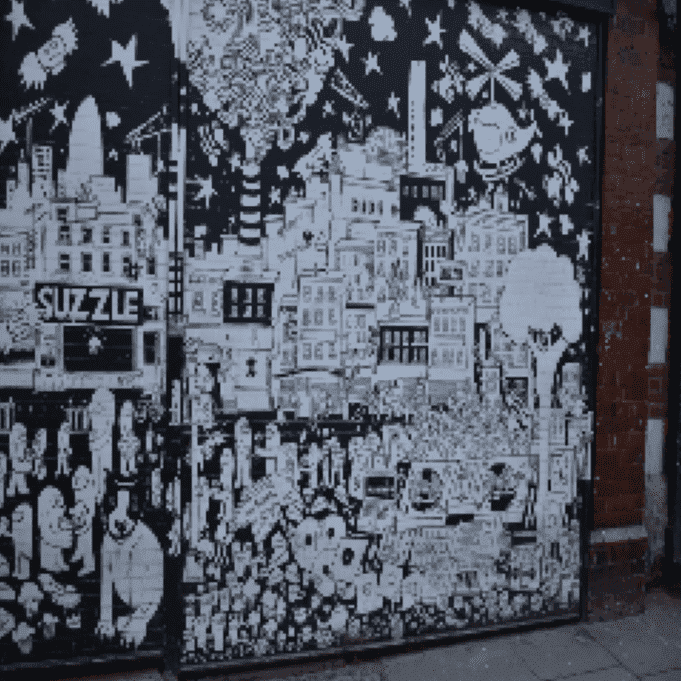}}\hfill
	\subfigure[]
	{\includegraphics[width=0.122\linewidth]{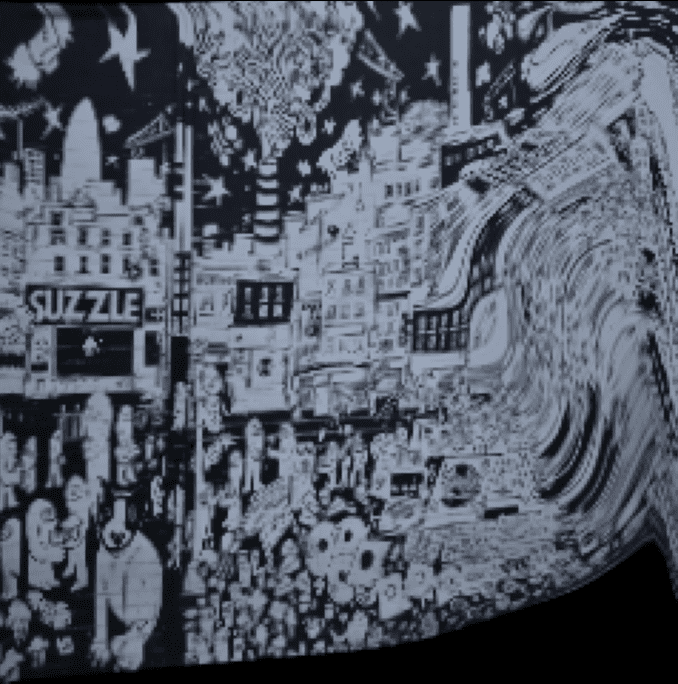}}\hfill
	\subfigure[]
	{\includegraphics[width=0.122\linewidth]{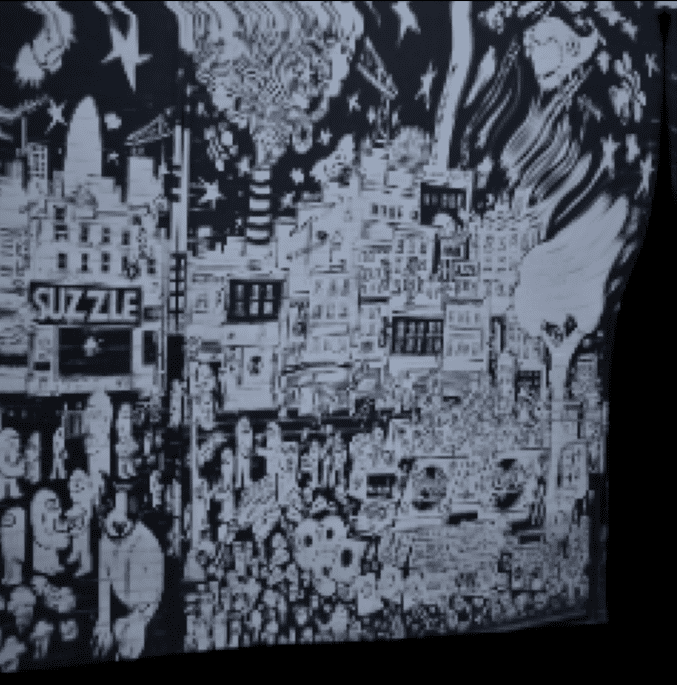}}\hfill
	\subfigure[]
	{\includegraphics[width=0.122\linewidth]{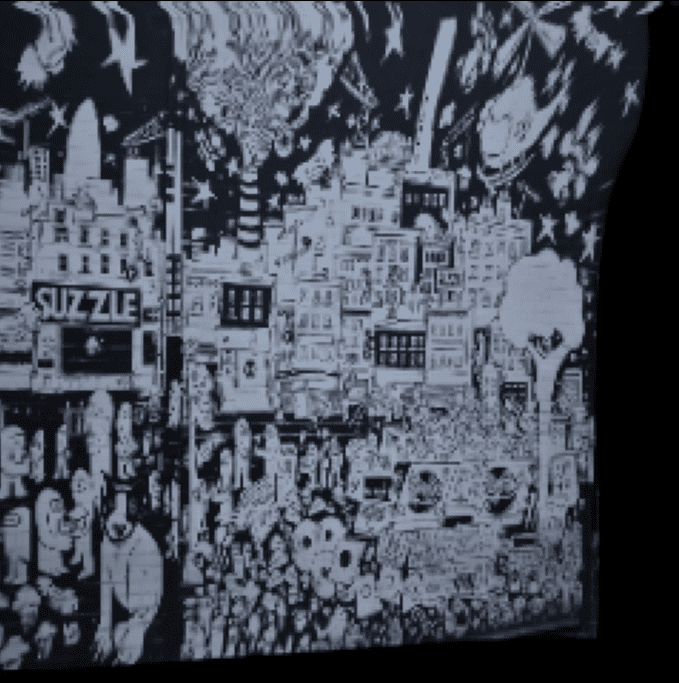}}\hfill
	\subfigure[]
	{\includegraphics[width=0.122\linewidth]{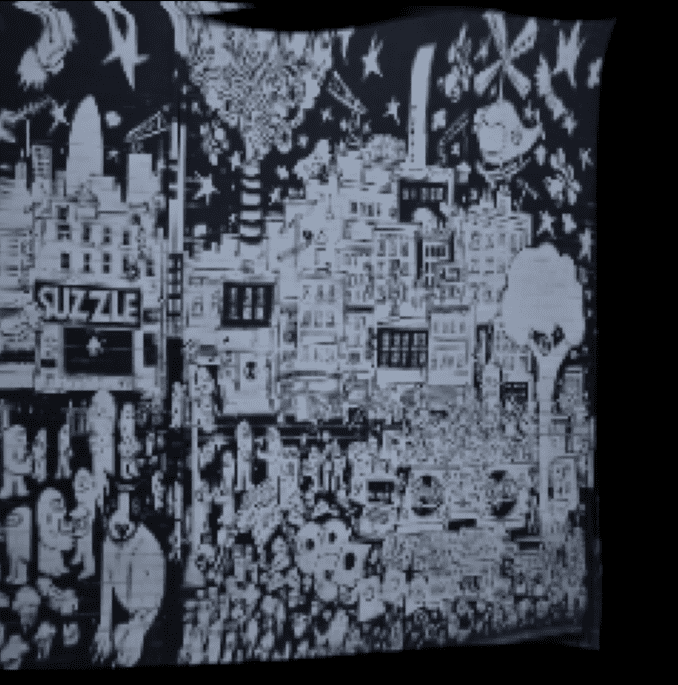}}\hfill
	\subfigure[]
	{\includegraphics[width=0.122\linewidth]{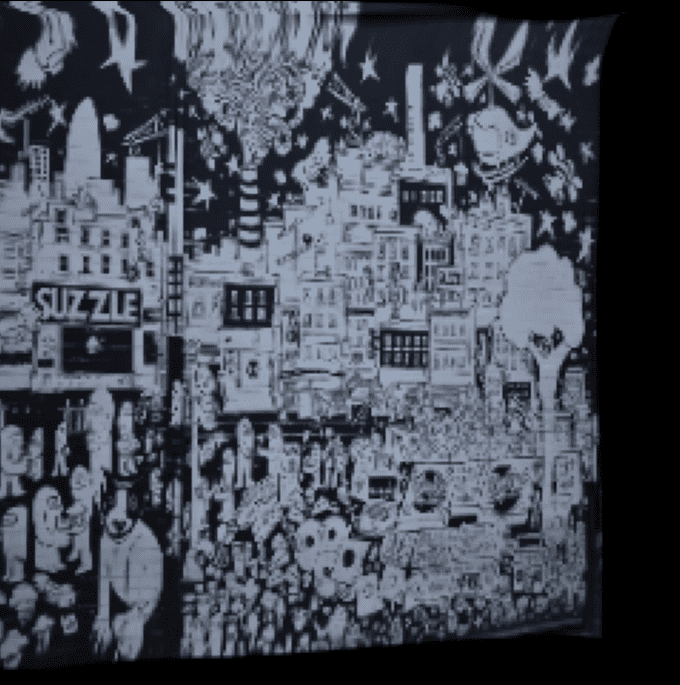}}\hfill
	\subfigure[]
	{\includegraphics[width=0.122\linewidth]{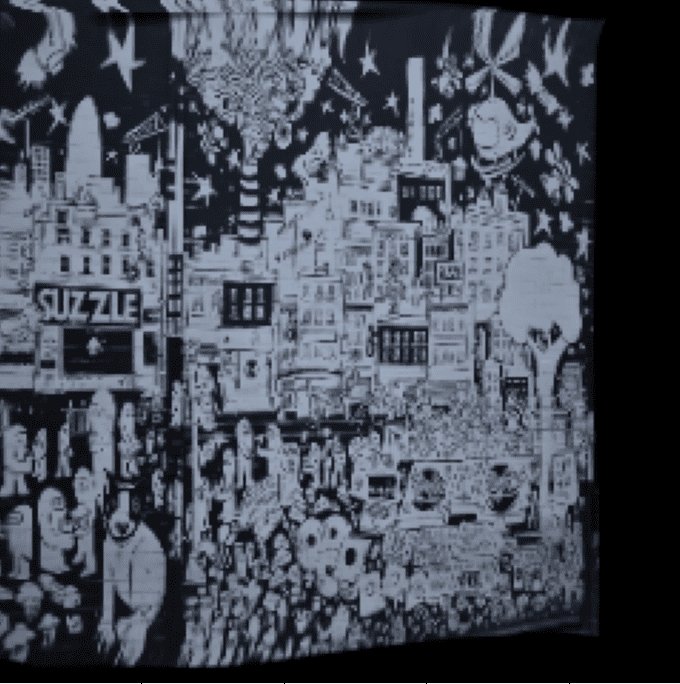}}\hfill\\
	\vspace{-21.5pt}
	\subfigure[]
	{\includegraphics[width=0.122\linewidth]{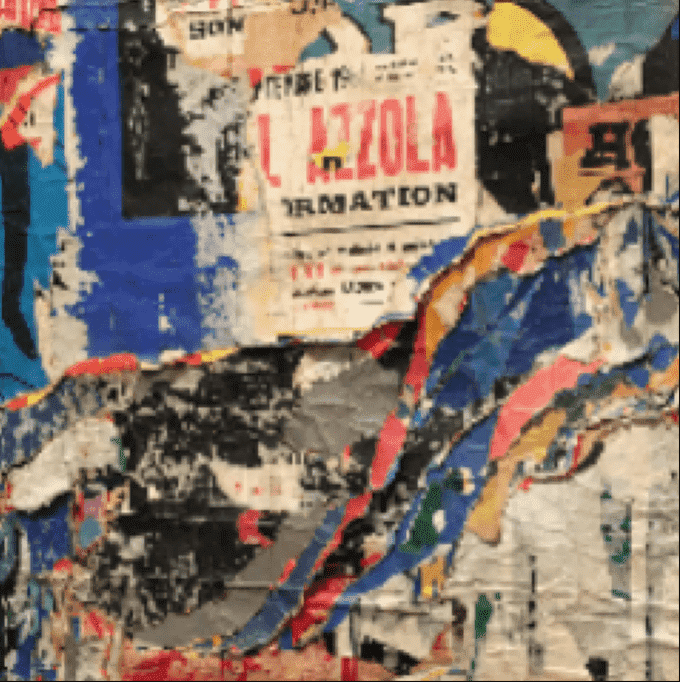}}\hfill
	\subfigure[]
	{\includegraphics[width=0.122\linewidth]{supplefig/iter/test113_1000_1.png}}\hfill
	\subfigure[]
	{\includegraphics[width=0.122\linewidth]{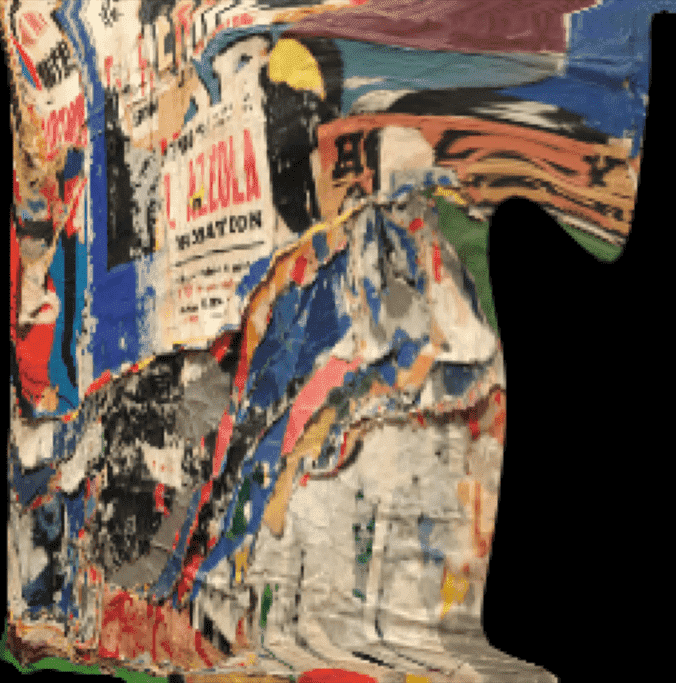}}\hfill
	\subfigure[]
	{\includegraphics[width=0.122\linewidth]{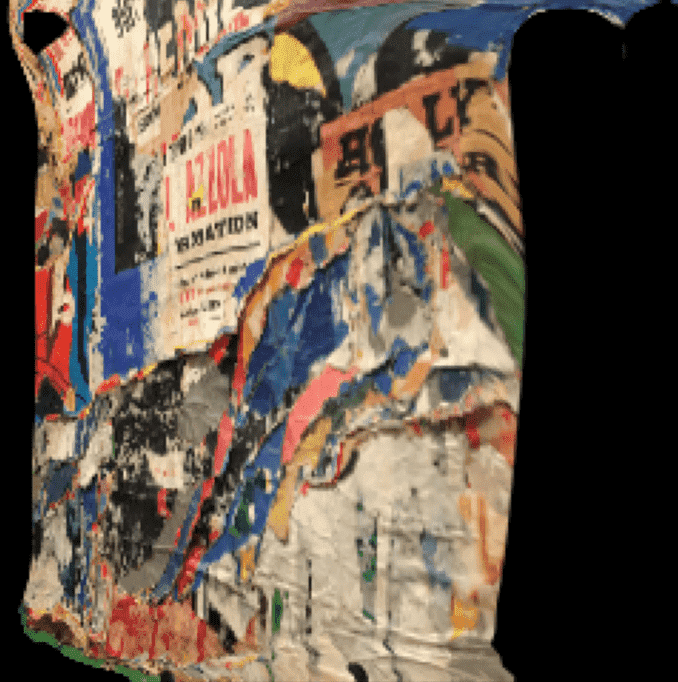}}\hfill
	\subfigure[]
	{\includegraphics[width=0.122\linewidth]{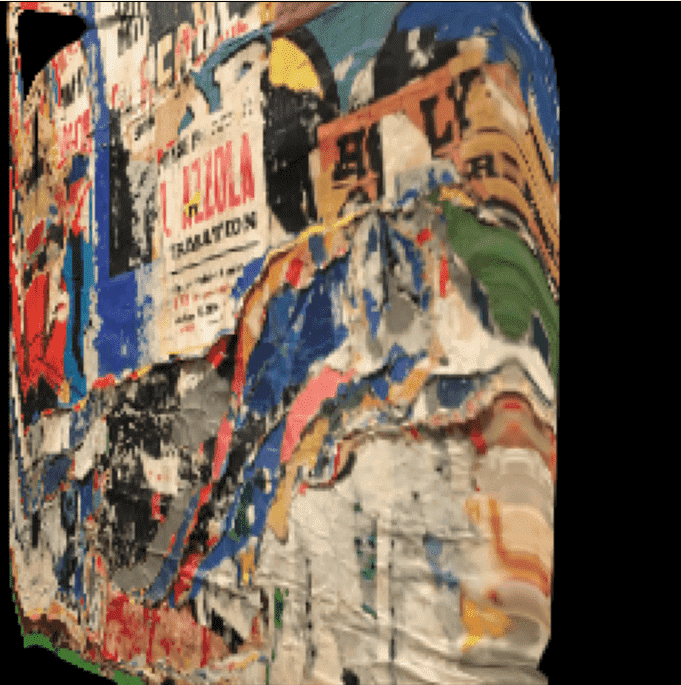}}\hfill
	\subfigure[]
	{\includegraphics[width=0.122\linewidth]{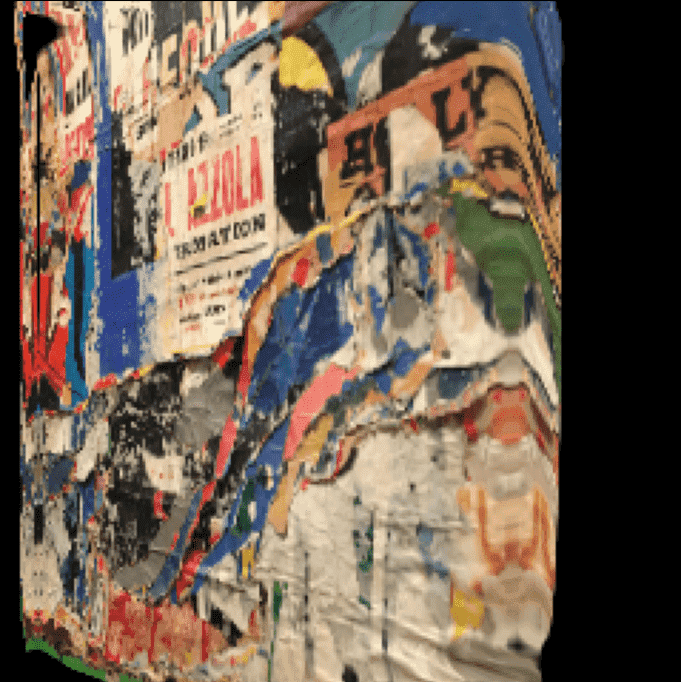}}\hfill
	\subfigure[]
	{\includegraphics[width=0.122\linewidth]{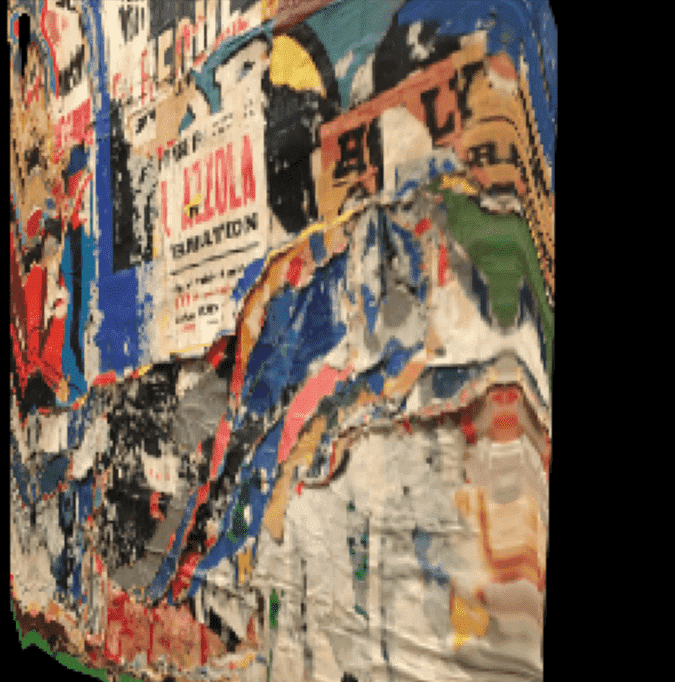}}\hfill
	\subfigure[]
	{\includegraphics[width=0.122\linewidth]{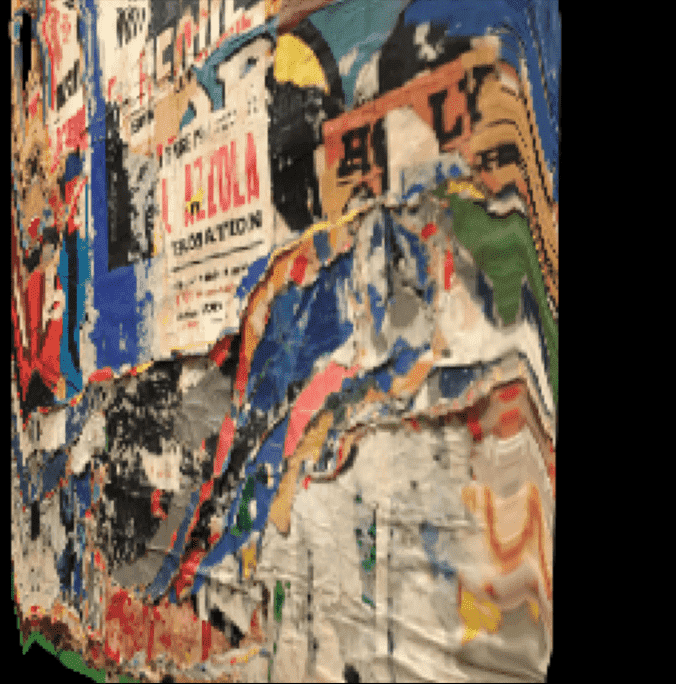}}\hfill\\
	\vspace{-21.5pt}
	\subfigure[]
	{\includegraphics[width=0.122\linewidth]{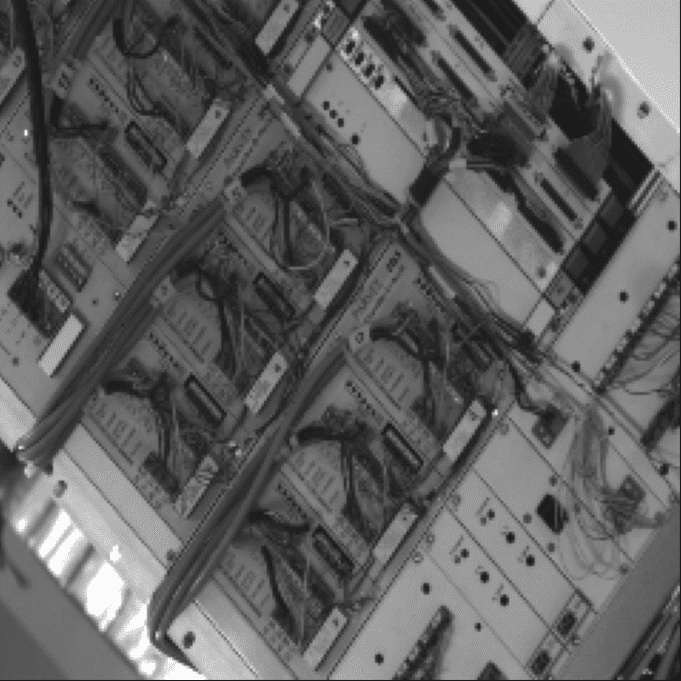}}\hfill
	\subfigure[]
	{\includegraphics[width=0.122\linewidth]{supplefig/iter/test_138_tgt.png}}\hfill
	\subfigure[]
	{\includegraphics[width=0.122\linewidth]{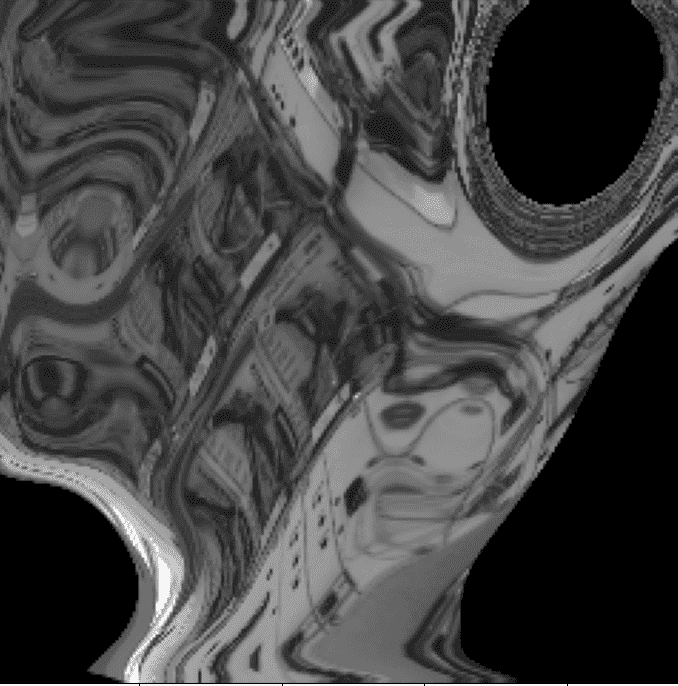}}\hfill
	\subfigure[]
	{\includegraphics[width=0.122\linewidth]{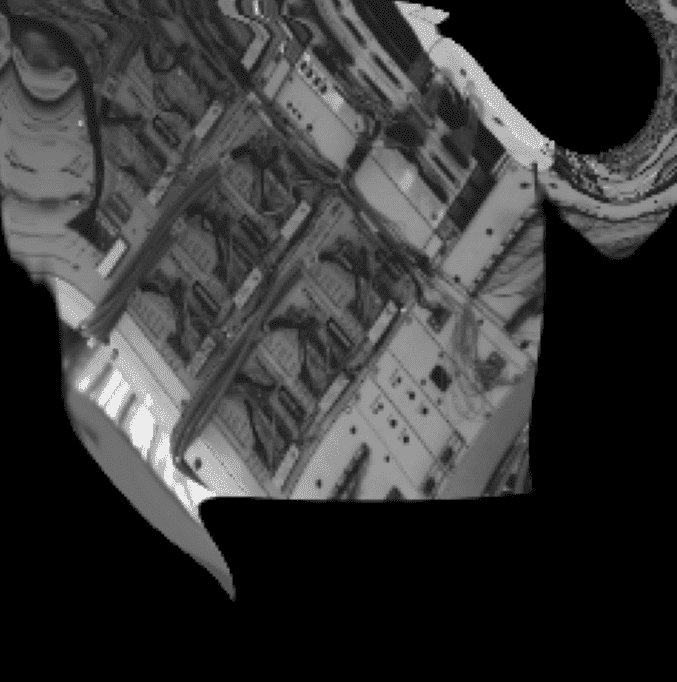}}\hfill
	\subfigure[]
	{\includegraphics[width=0.122\linewidth]{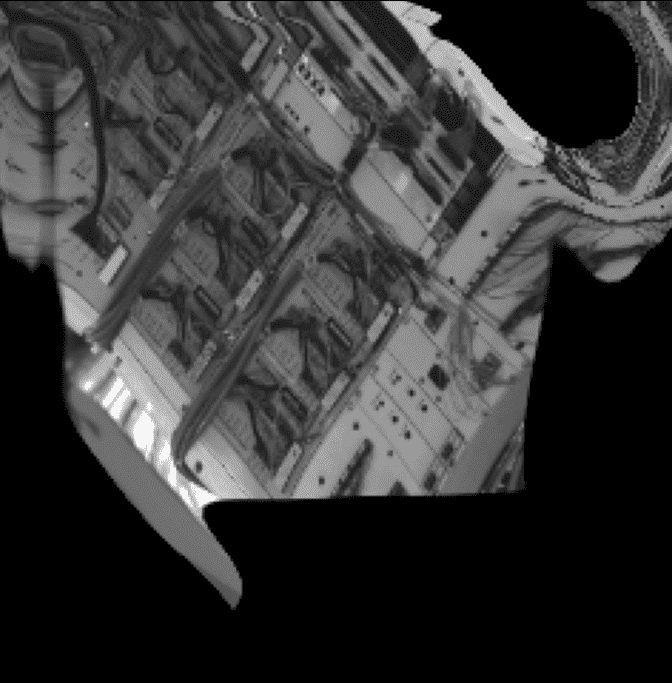}}\hfill
	\subfigure[]
	{\includegraphics[width=0.122\linewidth]{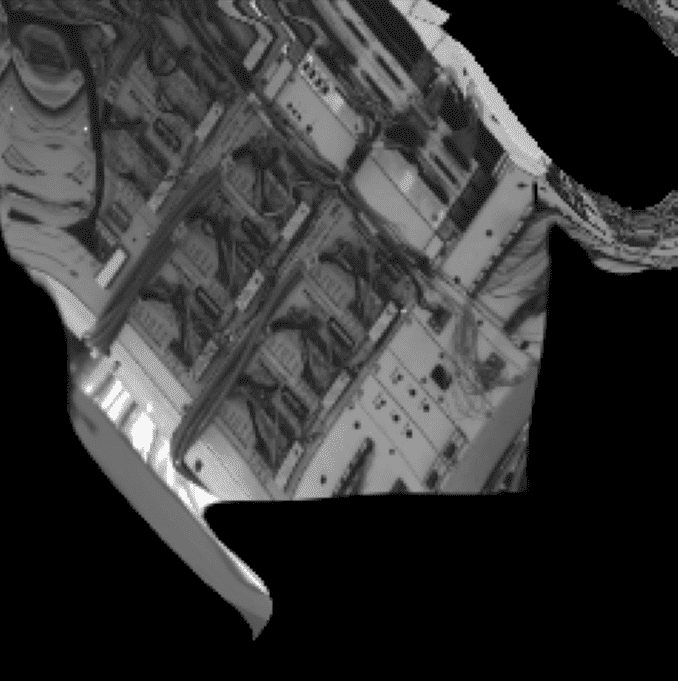}}\hfill
	\subfigure[]
	{\includegraphics[width=0.122\linewidth]{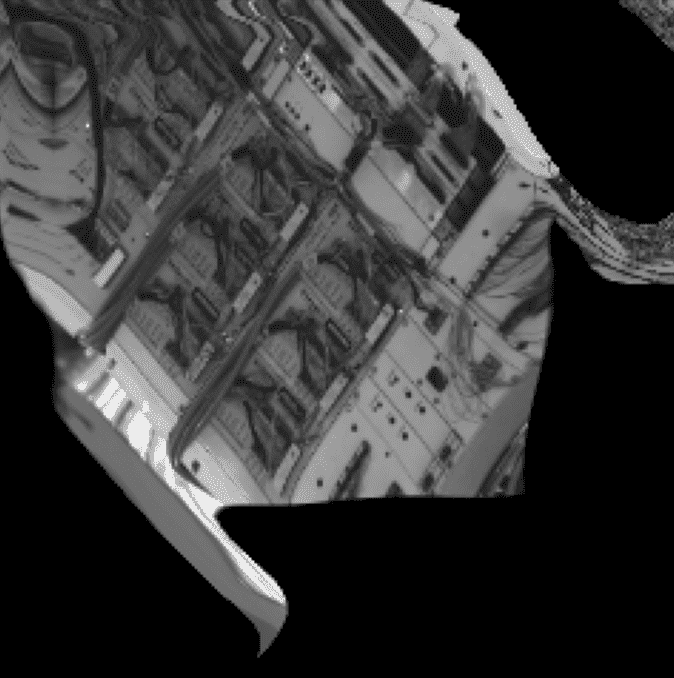}}\hfill
	\subfigure[]
	{\includegraphics[width=0.122\linewidth]{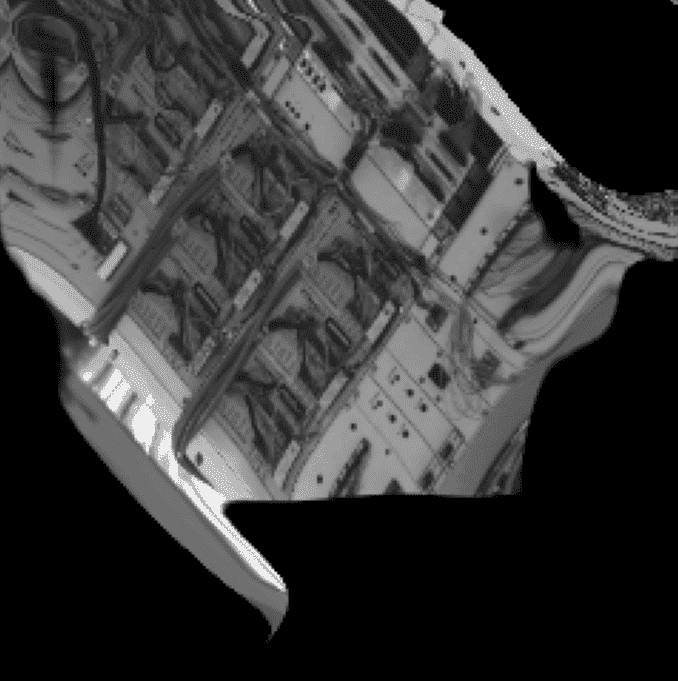}}\hfill\\
	\vspace{-21.5pt}
	\subfigure[]
	{\includegraphics[width=0.122\linewidth]{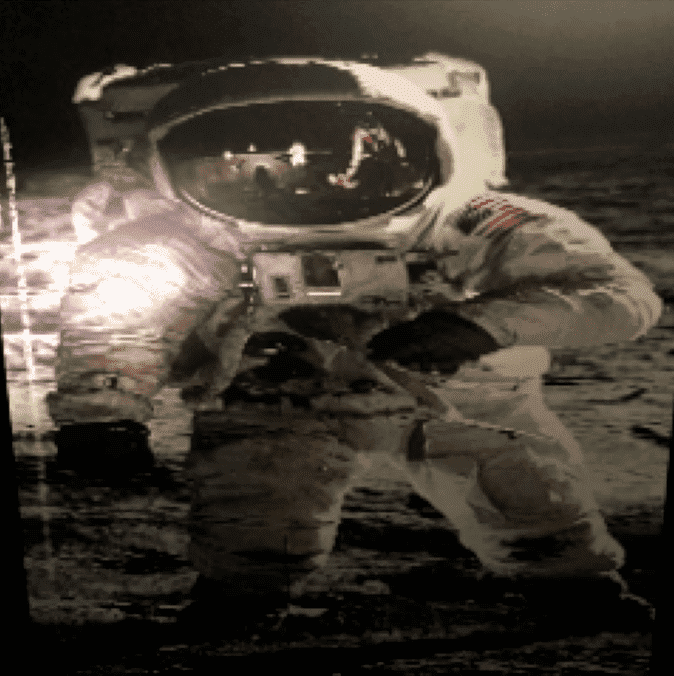}}\hfill
	\subfigure[]
	{\includegraphics[width=0.122\linewidth]{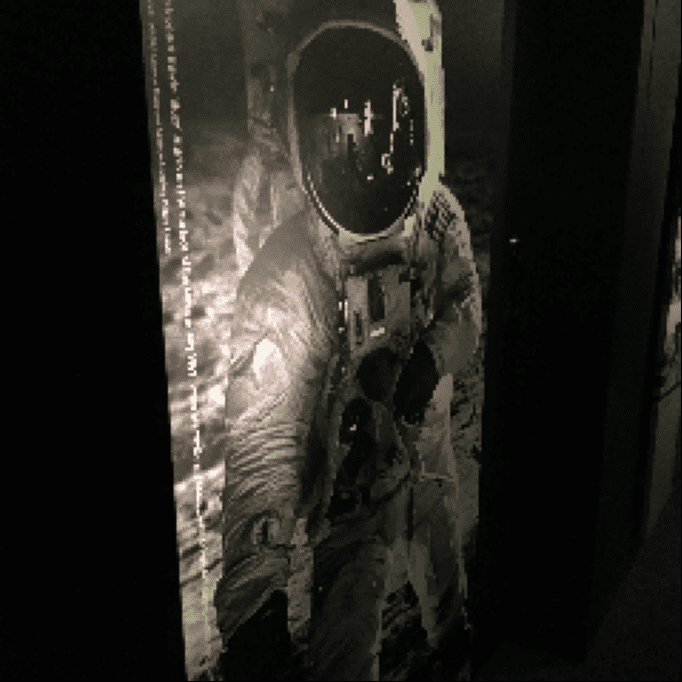}}\hfill
	\subfigure[]
	{\includegraphics[width=0.122\linewidth]{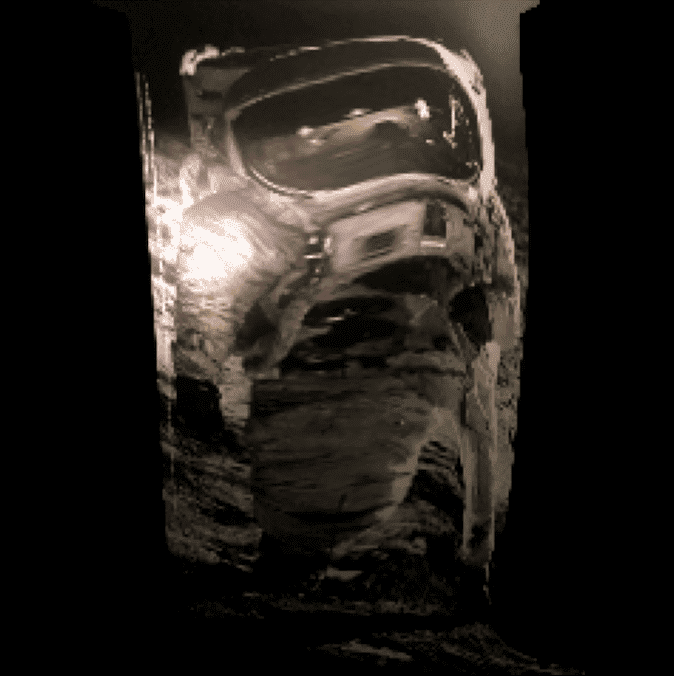}}\hfill
	\subfigure[]
	{\includegraphics[width=0.122\linewidth]{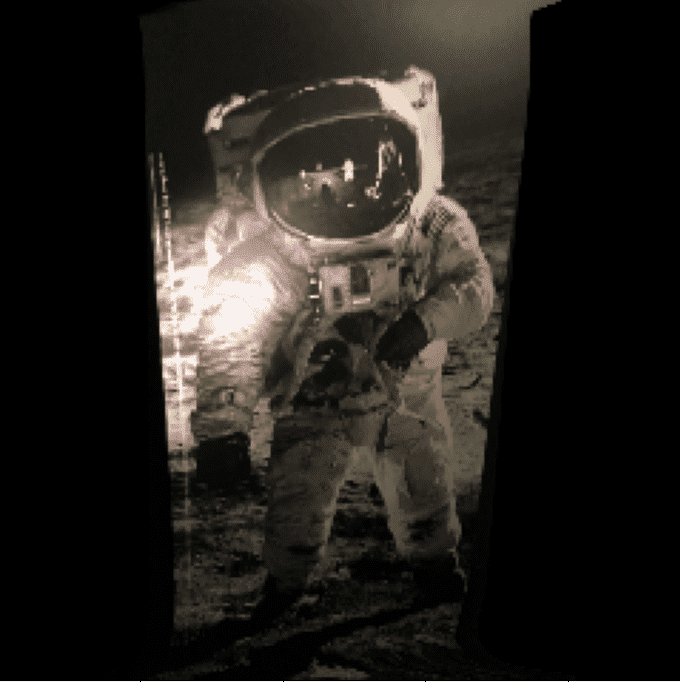}}\hfill
	\subfigure[]
	{\includegraphics[width=0.122\linewidth]{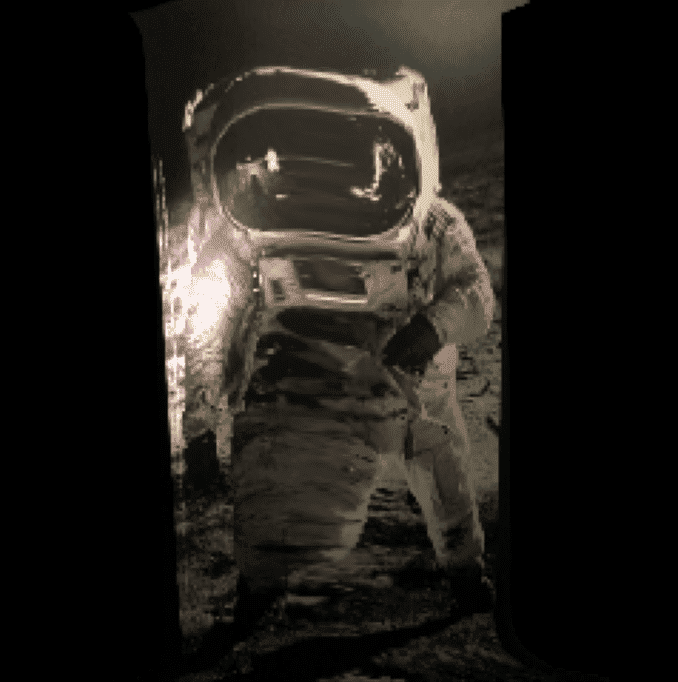}}\hfill
	\subfigure[]
	{\includegraphics[width=0.122\linewidth]{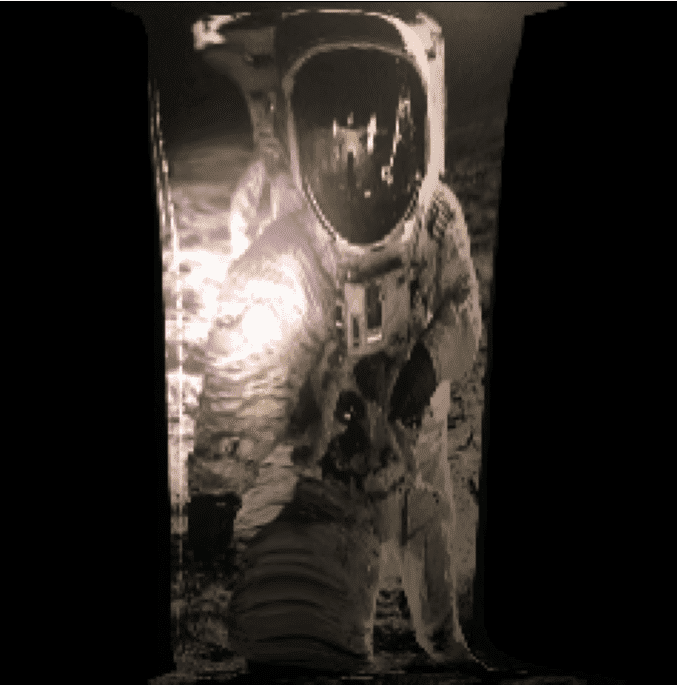}}\hfill
	\subfigure[]
	{\includegraphics[width=0.122\linewidth]{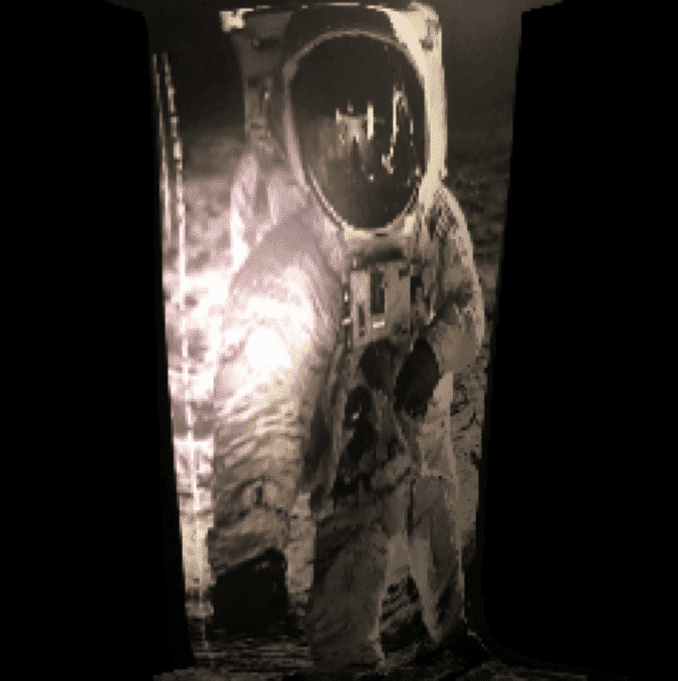}}\hfill
	\subfigure[]
	{\includegraphics[width=0.122\linewidth]{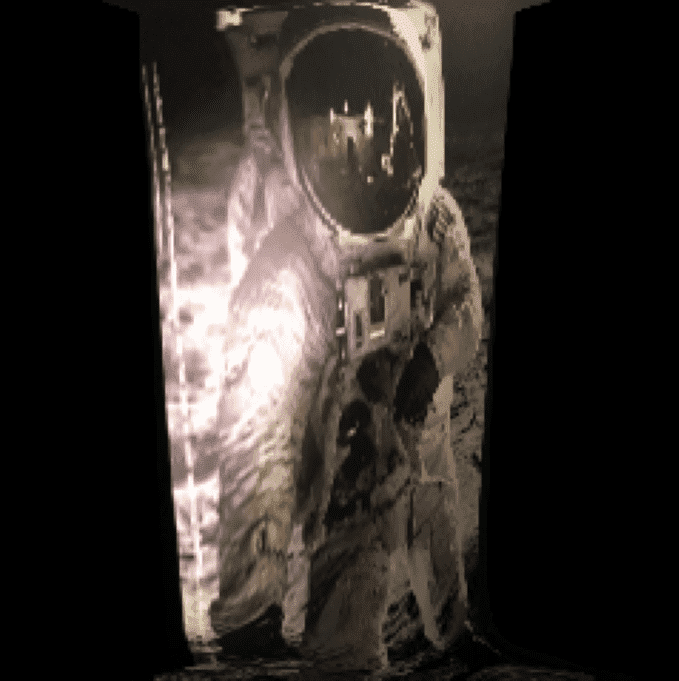}}\hfill\\
	\vspace{-21.5pt}
	\subfigure[(a)]
	{\includegraphics[width=0.122\linewidth]{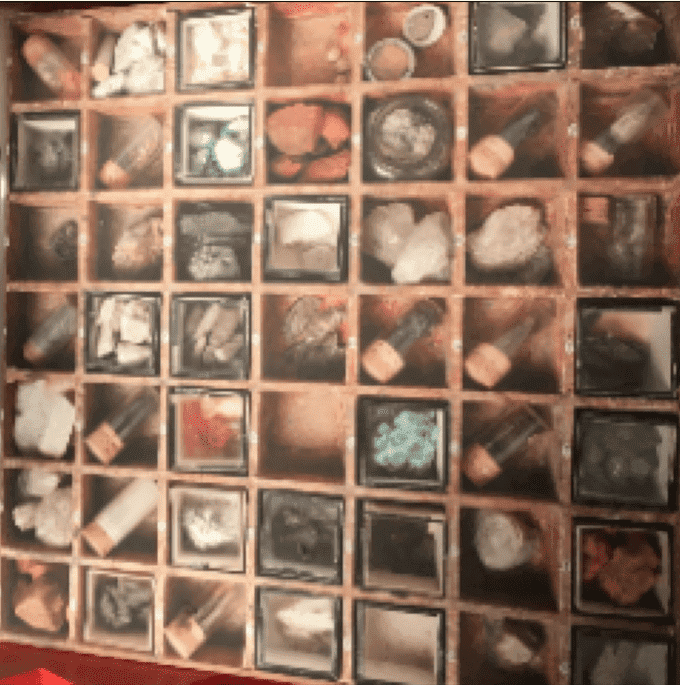}}\hfill
	\subfigure[(b)]
	{\includegraphics[width=0.122\linewidth]{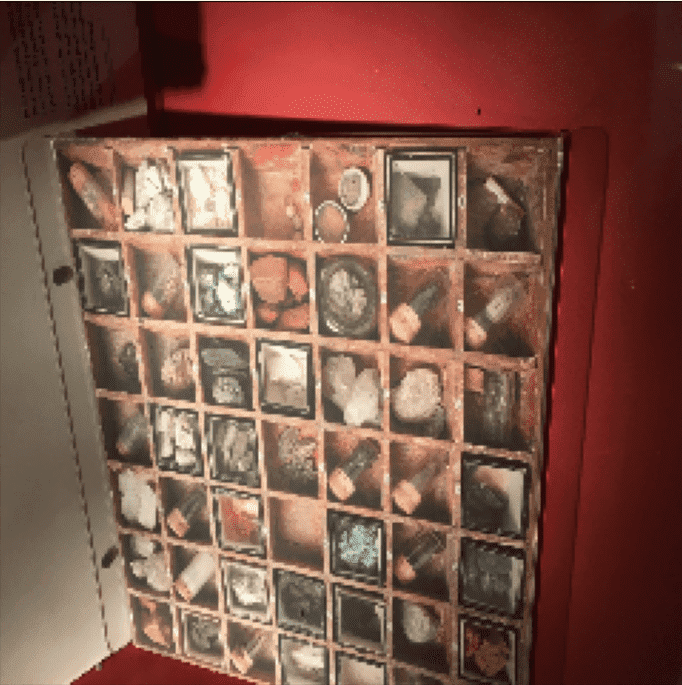}}\hfill
	\subfigure[(c)]
	{\includegraphics[width=0.122\linewidth]{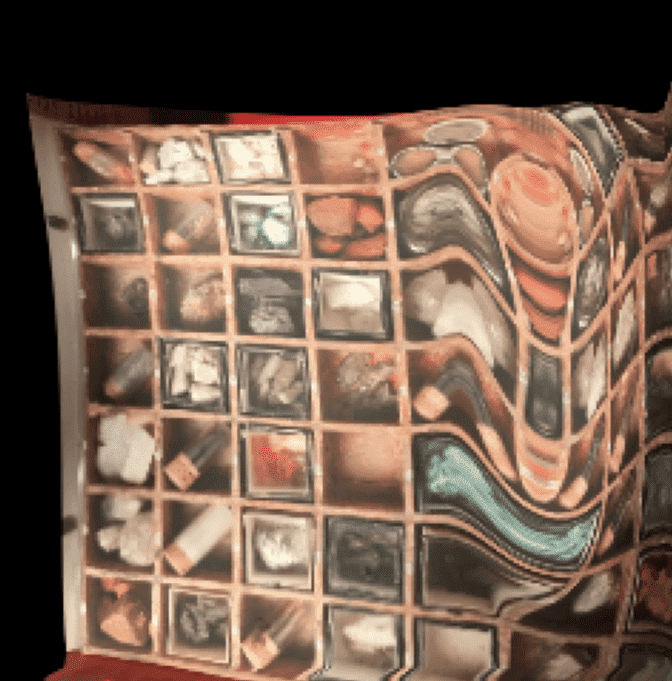}}\hfill
	\subfigure[(d)]
	{\includegraphics[width=0.122\linewidth]{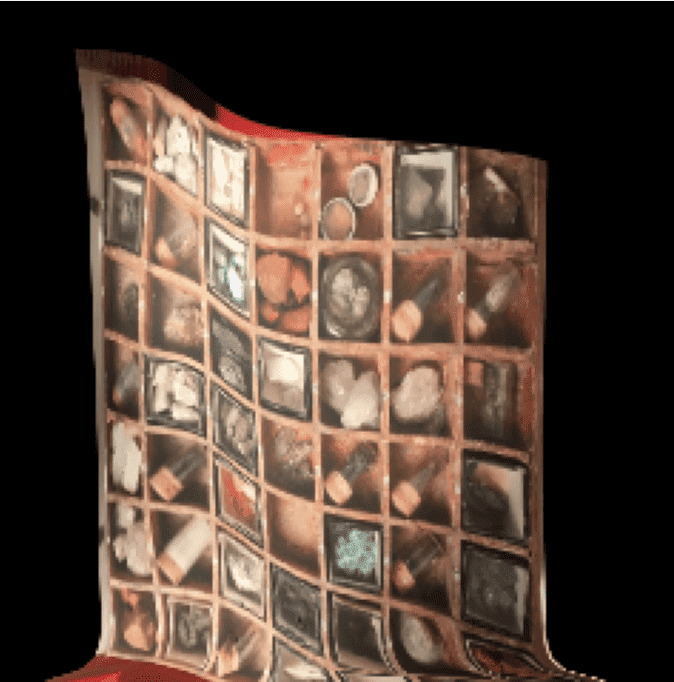}}\hfill
	\subfigure[(e)]
	{\includegraphics[width=0.122\linewidth]{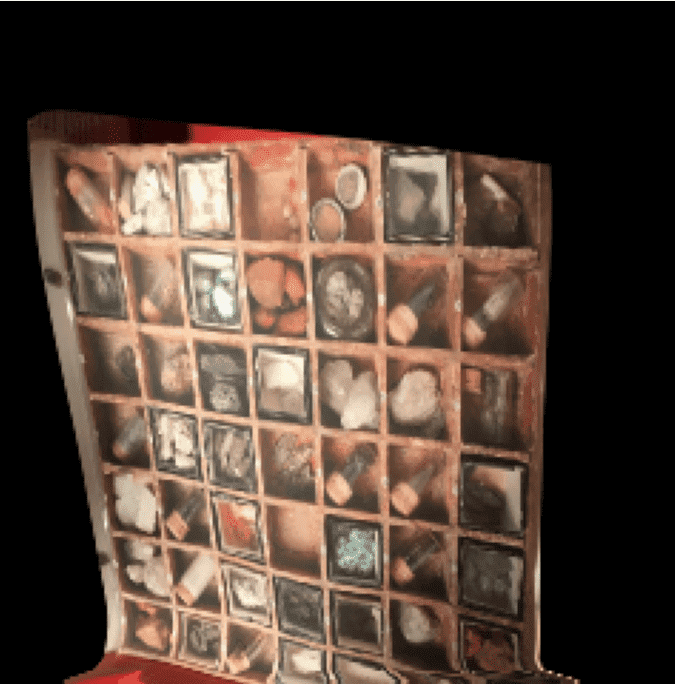}}\hfill
	\subfigure[(f)]
	{\includegraphics[width=0.122\linewidth]{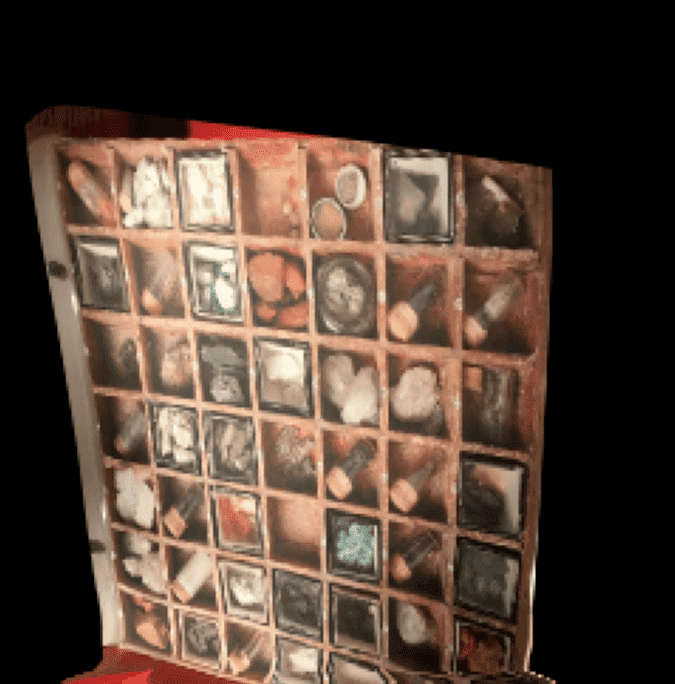}}\hfill
	\subfigure[(g)]
	{\includegraphics[width=0.122\linewidth]{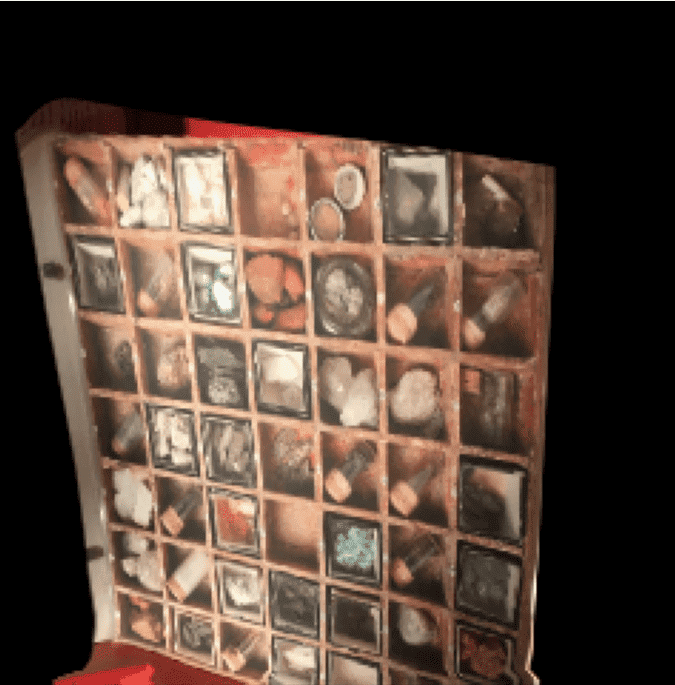}}\hfill
	\subfigure[(h)]
	{\includegraphics[width=0.122\linewidth]{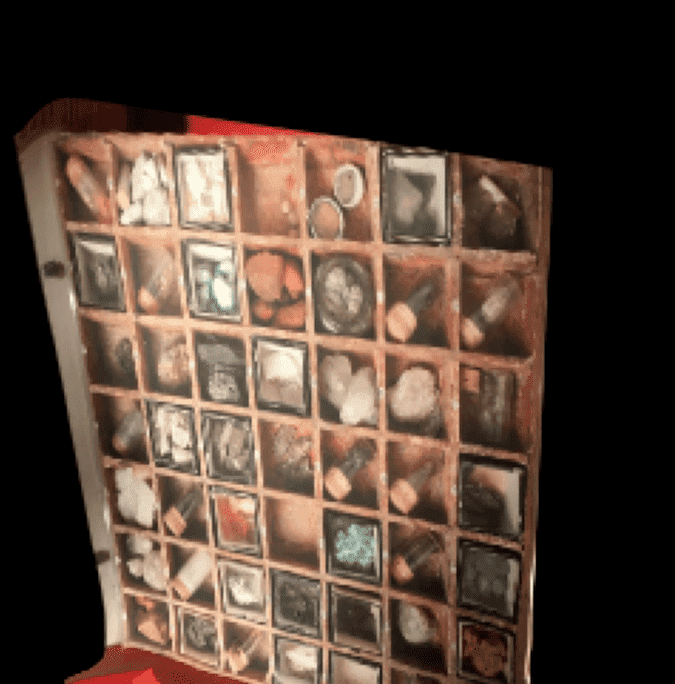}}\hfill\\
	\caption{\textbf{Convergence of DMP.} (a) source image, (b) target image, (c), (d), (e), (f), (g), and (h) iterative evolution of warped images by DMP. The error signal received at each iteration helps to correct the flow field, which the predicted transformation fields become progressively more accurate through iterative estimation.}\label{img:3}\vspace{-10pt}
\end{figure*}

\newpage

\begin{figure*}[t]
	\centering
	\renewcommand{\thesubfigure}{}
	\subfigure[]
	{\includegraphics[width=0.139\linewidth]{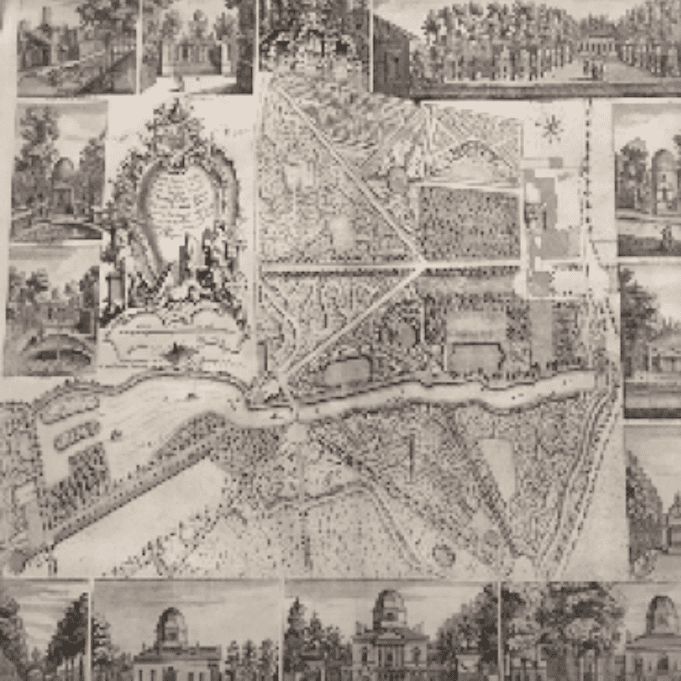}}\hfill
	\subfigure[]
	{\includegraphics[width=0.139\linewidth]{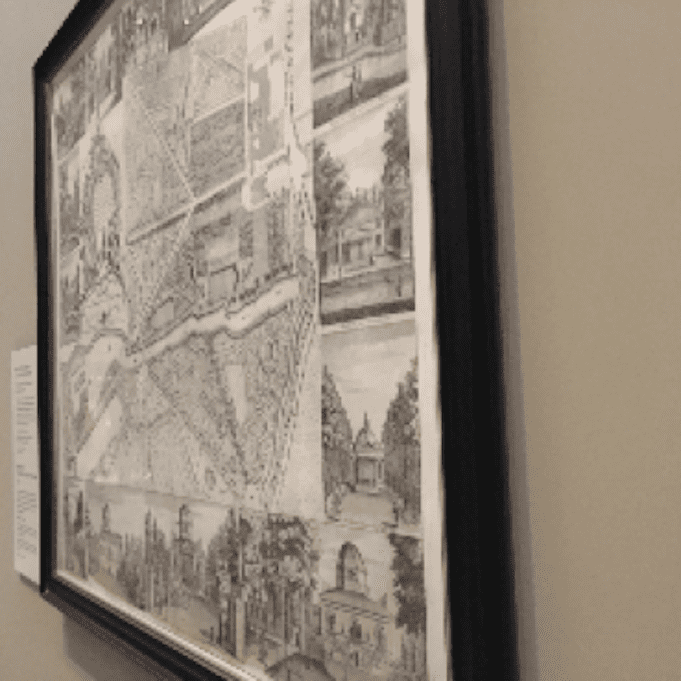}}\hfill
	\subfigure[]
	{\includegraphics[width=0.139\linewidth]{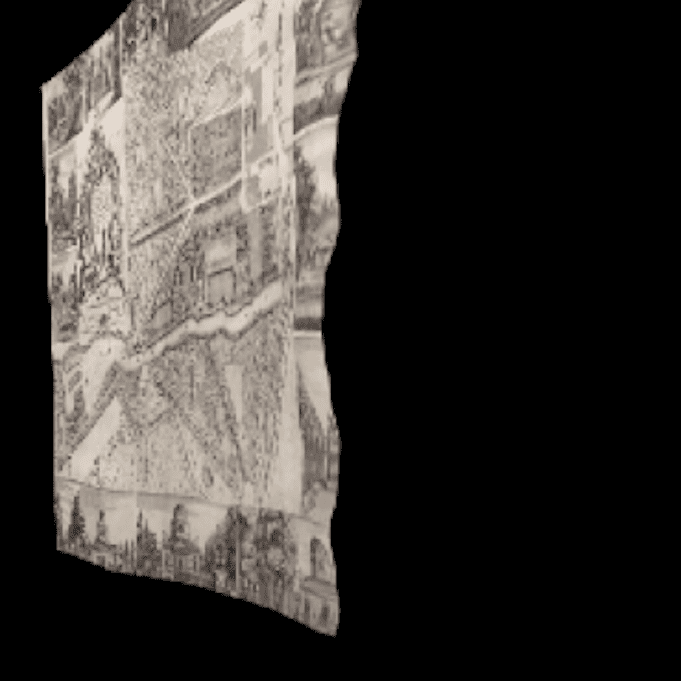}}\hfill
	\subfigure[]
	{\includegraphics[width=0.139\linewidth]{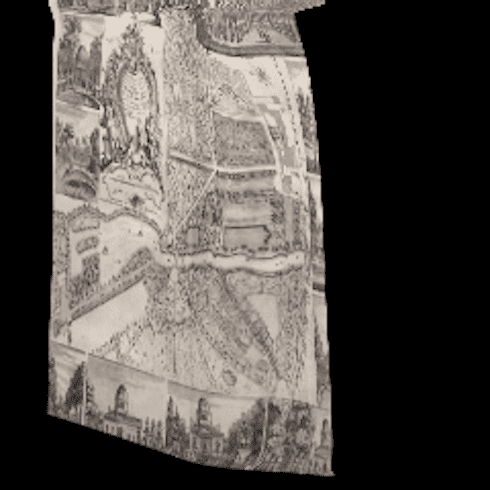}}\hfill
	\subfigure[]
	{\includegraphics[width=0.139\linewidth]{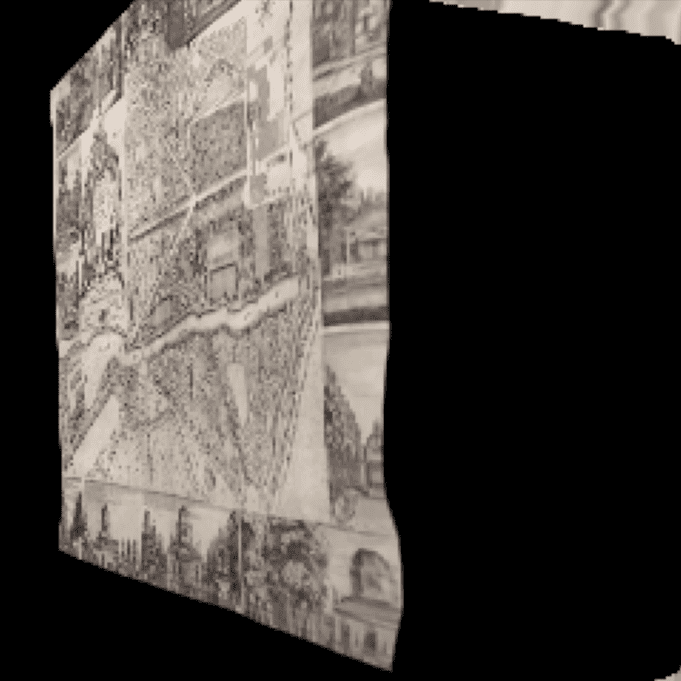}}\hfill
	\subfigure[]
	{\includegraphics[width=0.139\linewidth]{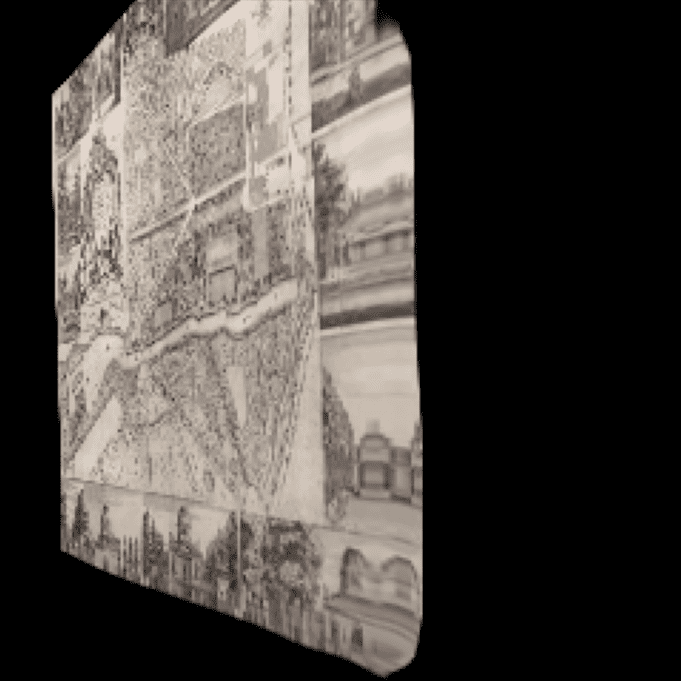}}\hfill
	\subfigure[]
	{\includegraphics[width=0.139\linewidth]{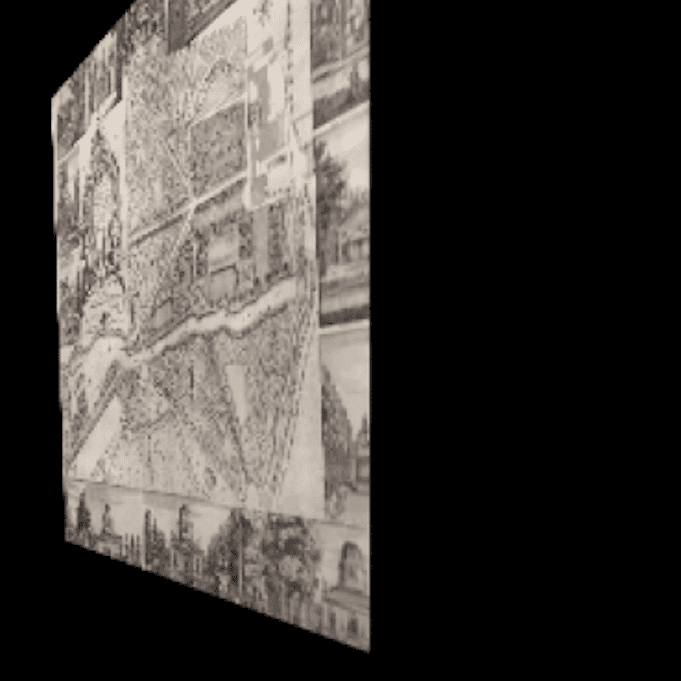}}\hfill\\
	\vspace{-21.5pt}
	\subfigure[]
	{\includegraphics[width=0.139\linewidth]{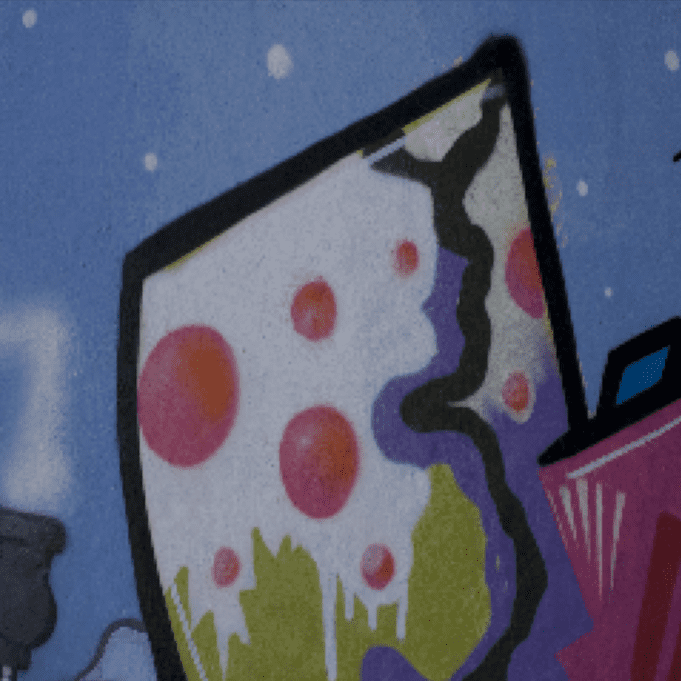}}\hfill
	\subfigure[]
	{\includegraphics[width=0.139\linewidth]{supplefig/hpatch/tgt_2.png}}\hfill
	\subfigure[]
	{\includegraphics[width=0.139\linewidth]{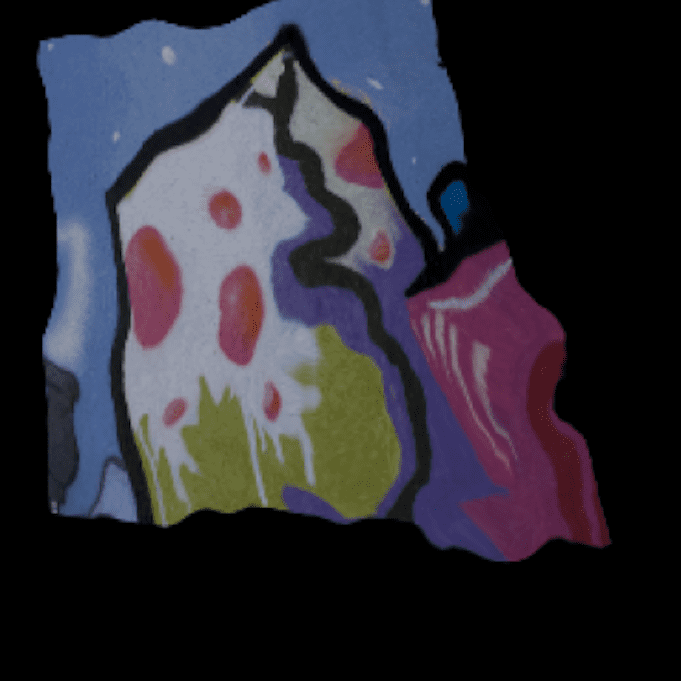}}\hfill
	\subfigure[]
	{\includegraphics[width=0.139\linewidth]{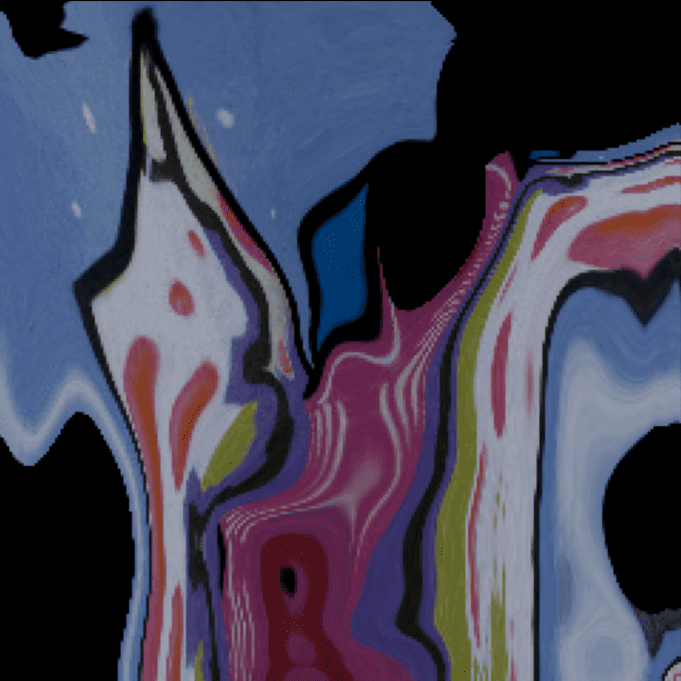}}\hfill
	\subfigure[]
	{\includegraphics[width=0.139\linewidth]{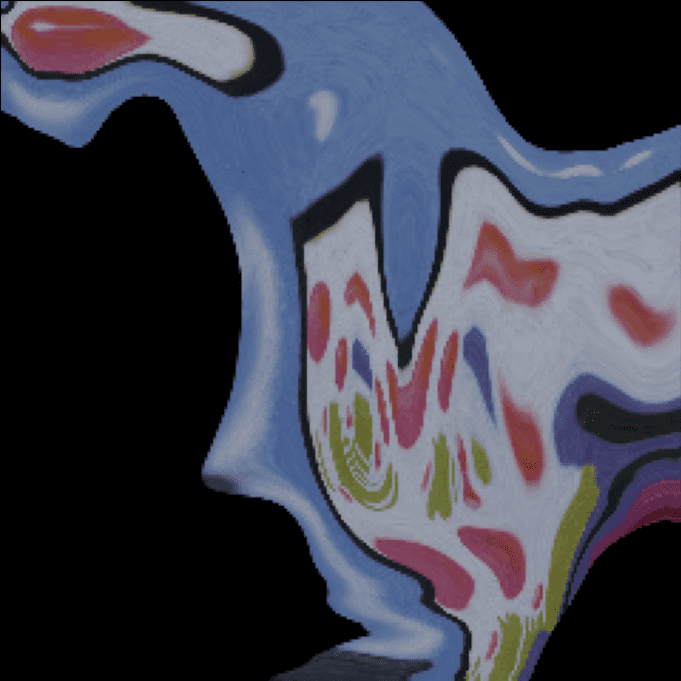}}\hfill
	\subfigure[]
	{\includegraphics[width=0.139\linewidth]{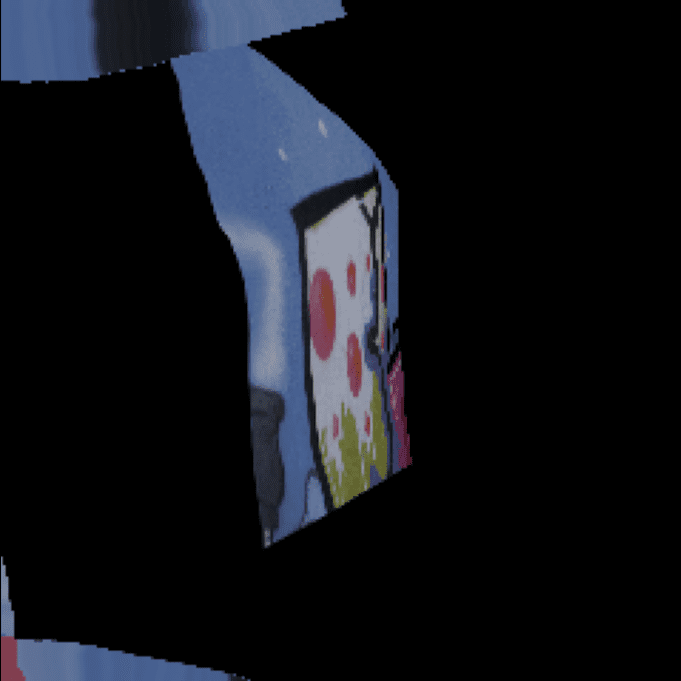}}\hfill
	\subfigure[]
	{\includegraphics[width=0.139\linewidth]{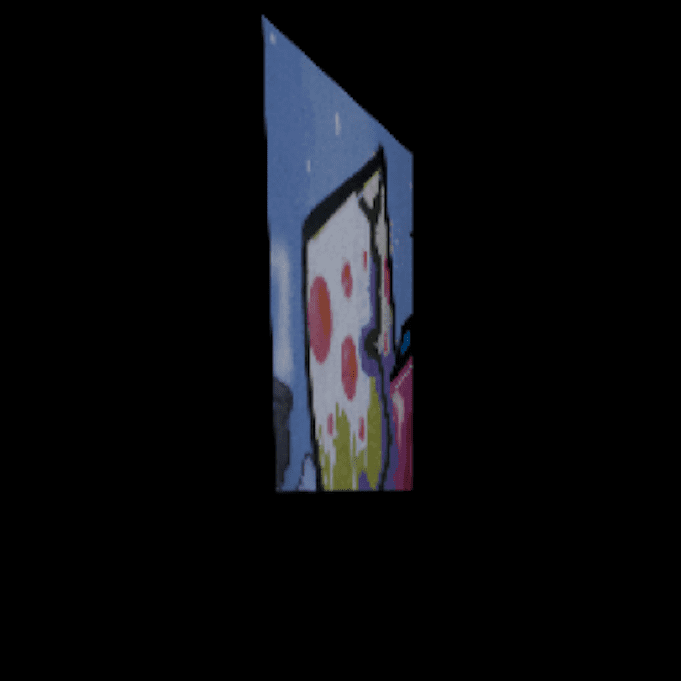}}\hfill\\
	\vspace{-21.5pt}
	\subfigure[]
	{\includegraphics[width=0.139\linewidth]{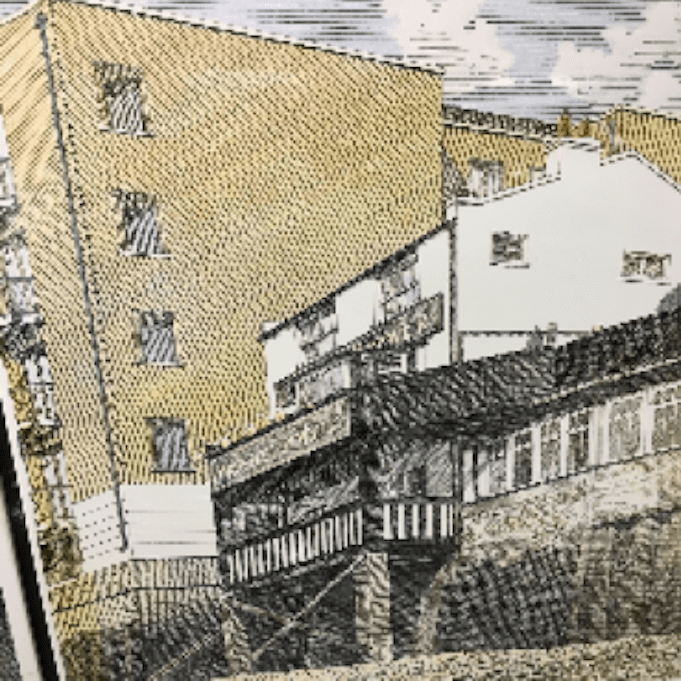}}\hfill
	\subfigure[]
	{\includegraphics[width=0.139\linewidth]{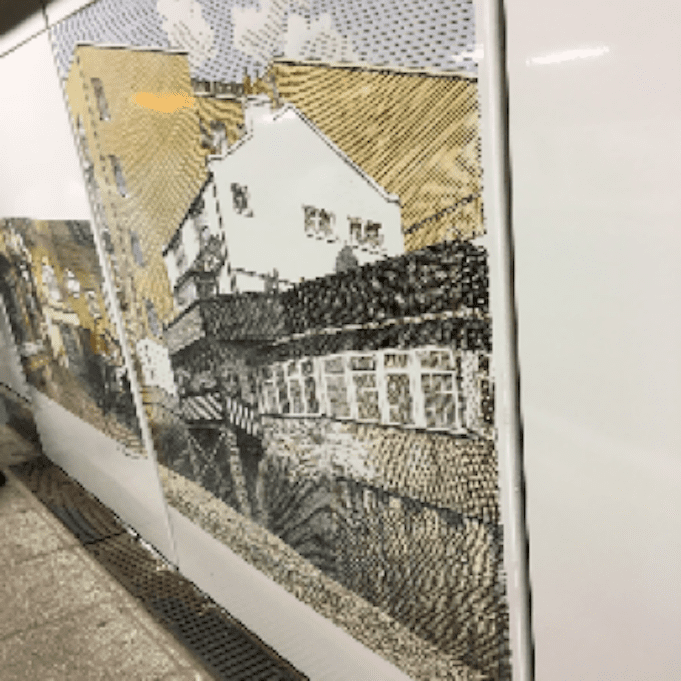}}\hfill
	\subfigure[]
	{\includegraphics[width=0.139\linewidth]{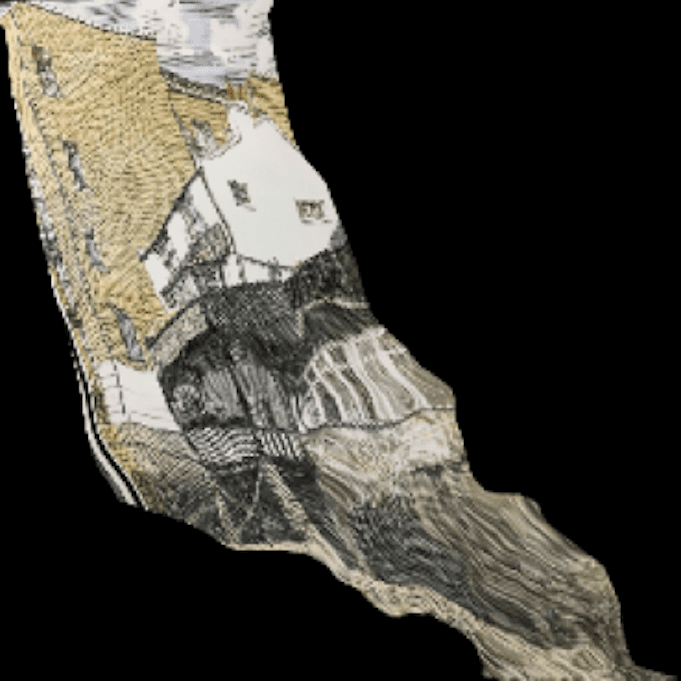}}\hfill
	\subfigure[]
	{\includegraphics[width=0.139\linewidth]{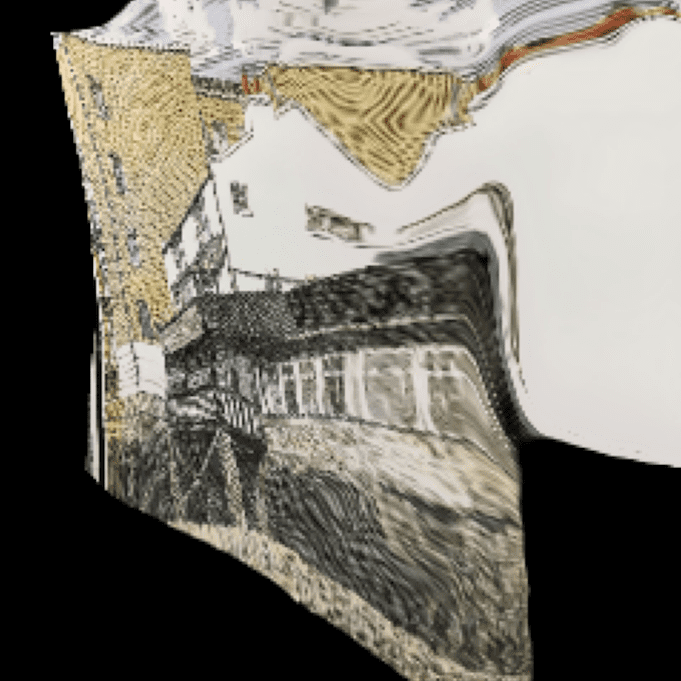}}\hfill
	\subfigure[]
	{\includegraphics[width=0.139\linewidth]{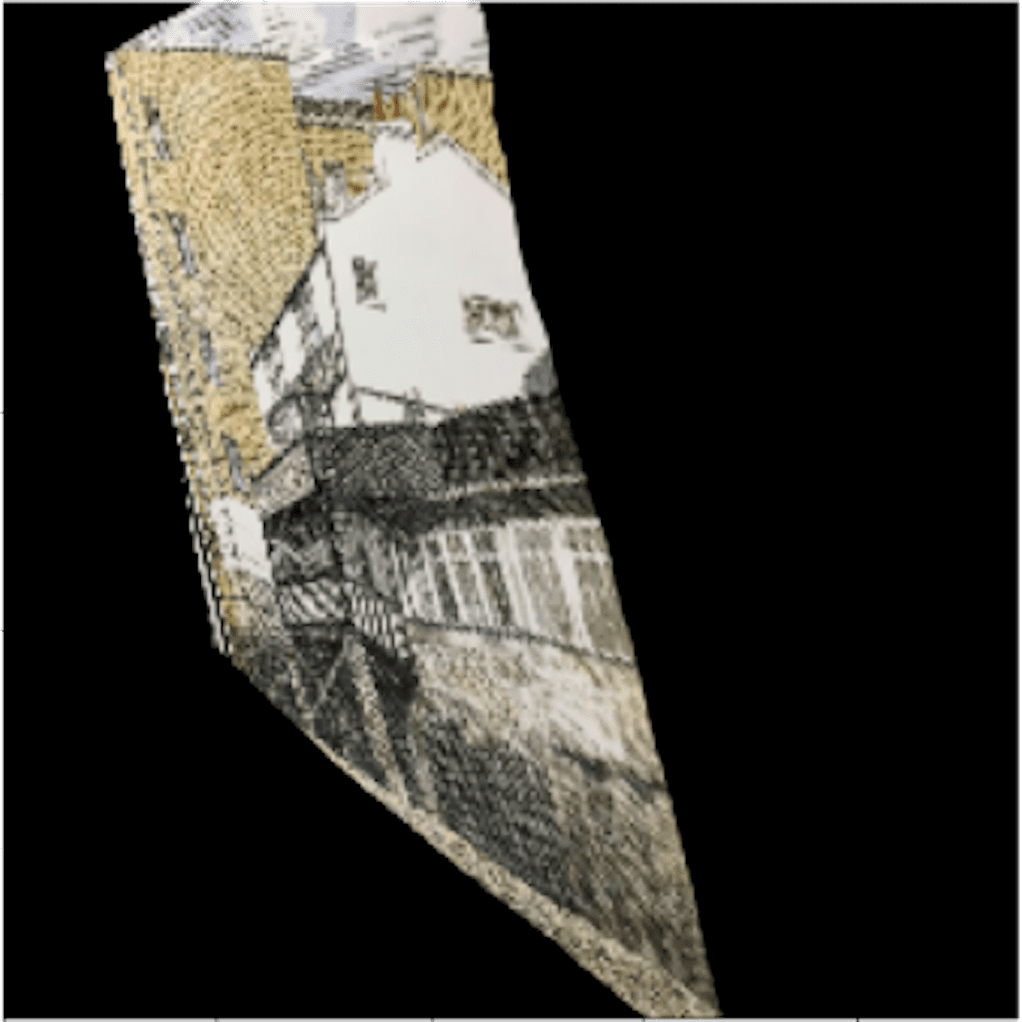}}\hfill
	\subfigure[]
	{\includegraphics[width=0.139\linewidth]{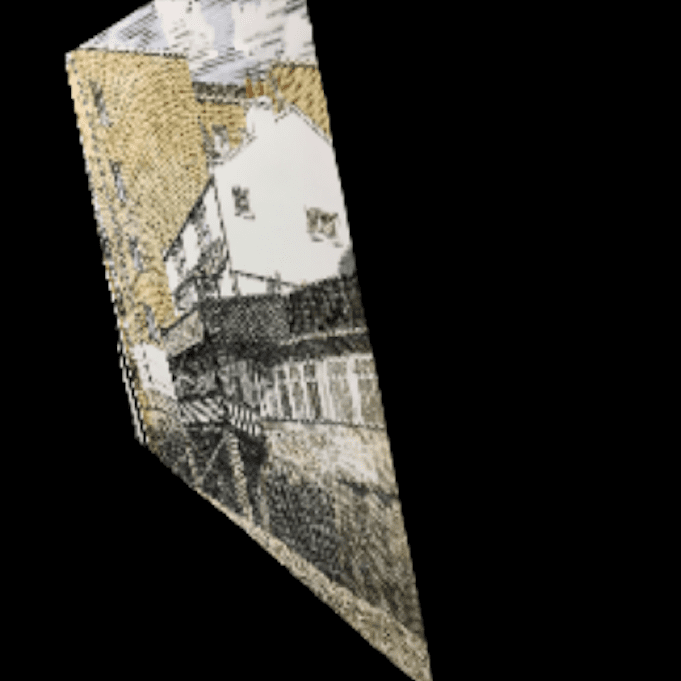}}\hfill
	\subfigure[]
	{\includegraphics[width=0.139\linewidth]{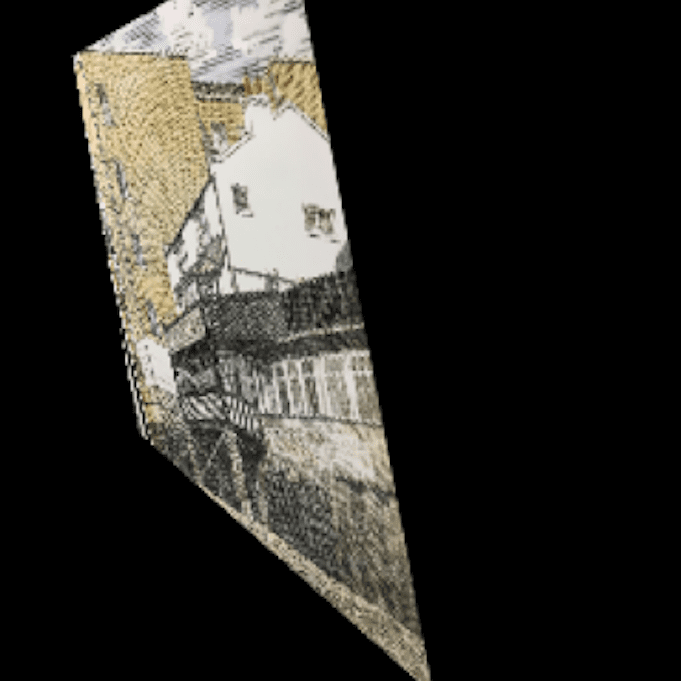}}\hfill\\
	\vspace{-21.5pt}
	\subfigure[]
	{\includegraphics[width=0.139\linewidth]{supplefig/hpatch/src_20.png}}\hfill
	\subfigure[]
	{\includegraphics[width=0.139\linewidth]{supplefig/hpatch/tgt_20.png}}\hfill
	\subfigure[]
	{\includegraphics[width=0.139\linewidth]{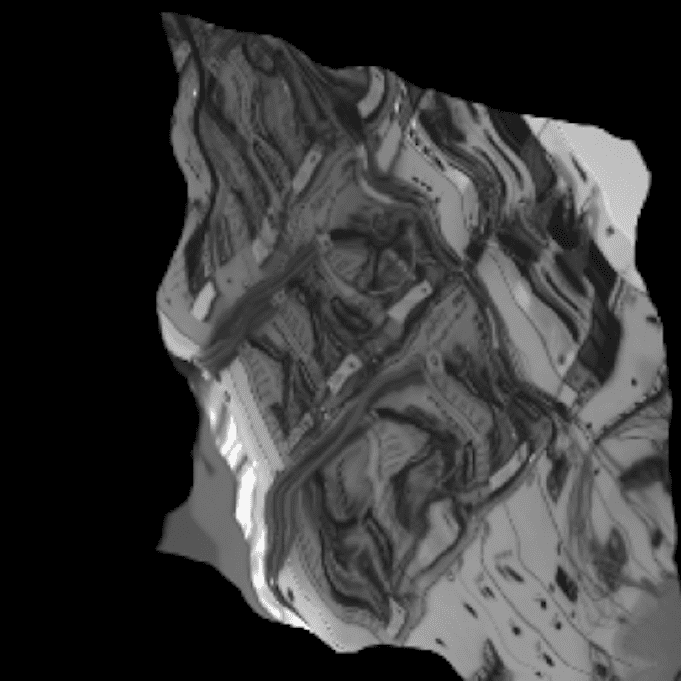}}\hfill
	\subfigure[]
	{\includegraphics[width=0.139\linewidth]{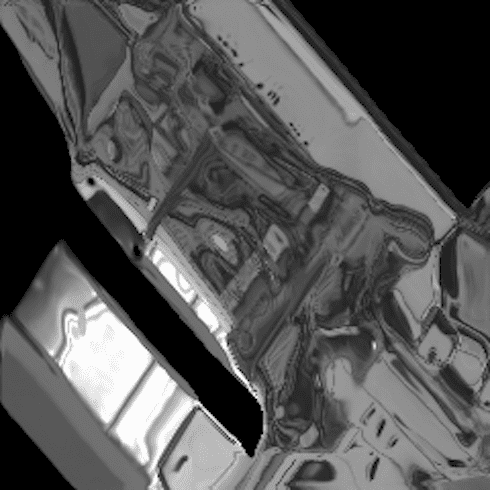}}\hfill
	\subfigure[]
	{\includegraphics[width=0.139\linewidth]{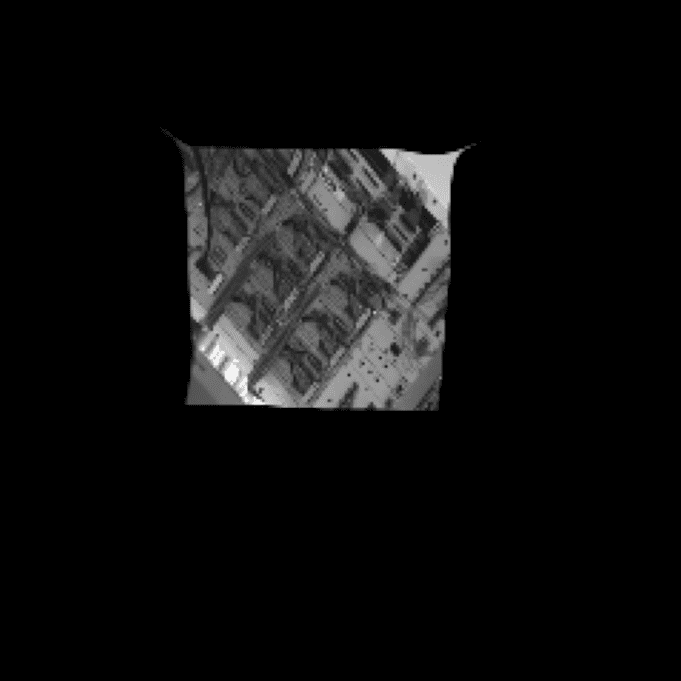}}\hfill
	\subfigure[]
	{\includegraphics[width=0.139\linewidth]{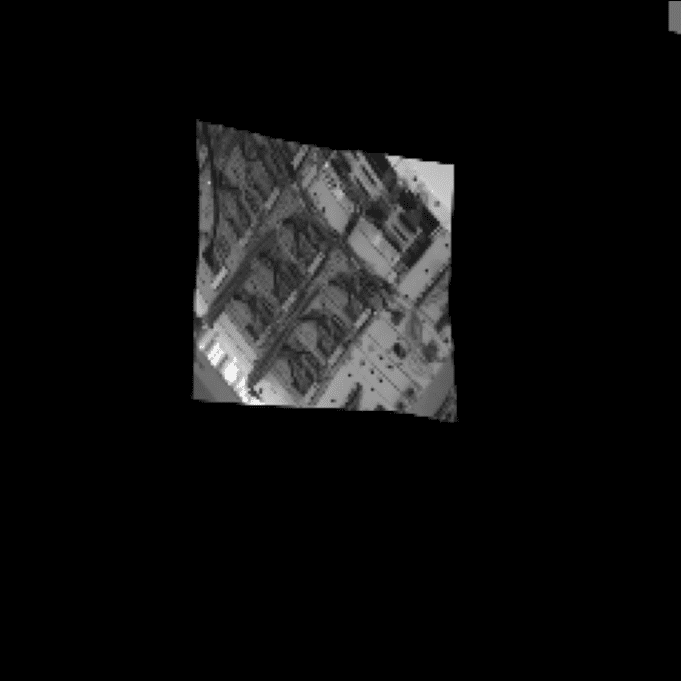}}\hfill
	\subfigure[]
	{\includegraphics[width=0.139\linewidth]{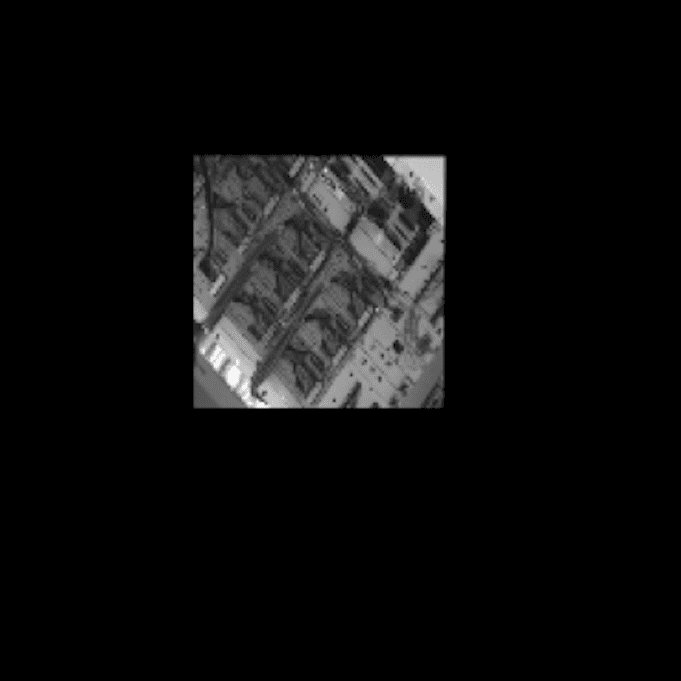}}\hfill\\
	\vspace{-21.5pt}	
	\subfigure[]
	{\includegraphics[width=0.139\linewidth]{supplefig/hpatch/src_24.png}}\hfill
	\subfigure[]
	{\includegraphics[width=0.139\linewidth]{supplefig/hpatch/tgt_24.png}}\hfill
	\subfigure[]
	{\includegraphics[width=0.139\linewidth]{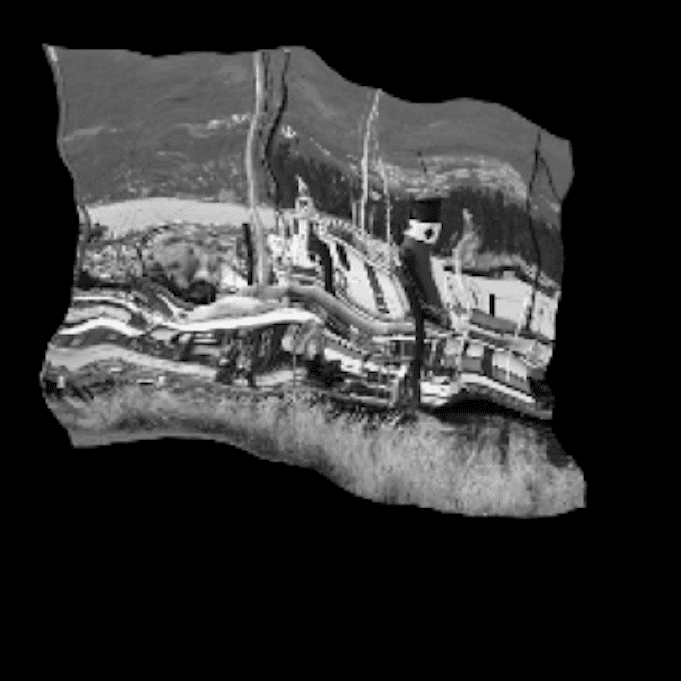}}\hfill
	\subfigure[]
	{\includegraphics[width=0.139\linewidth]{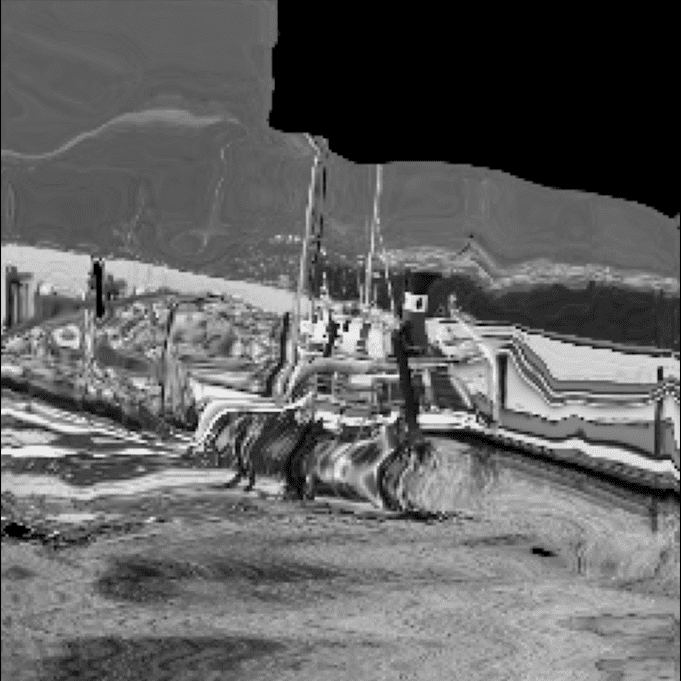}}\hfill
	\subfigure[]
	{\includegraphics[width=0.139\linewidth]{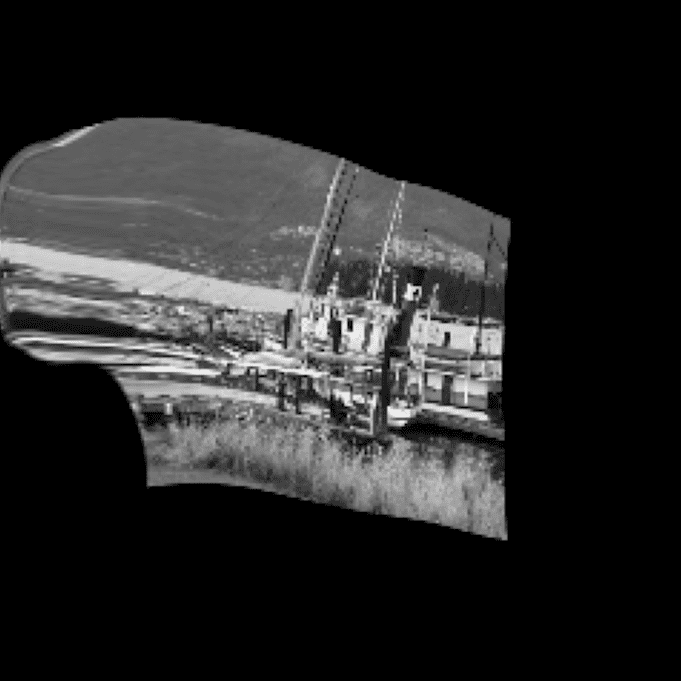}}\hfill
	\subfigure[]
	{\includegraphics[width=0.139\linewidth]{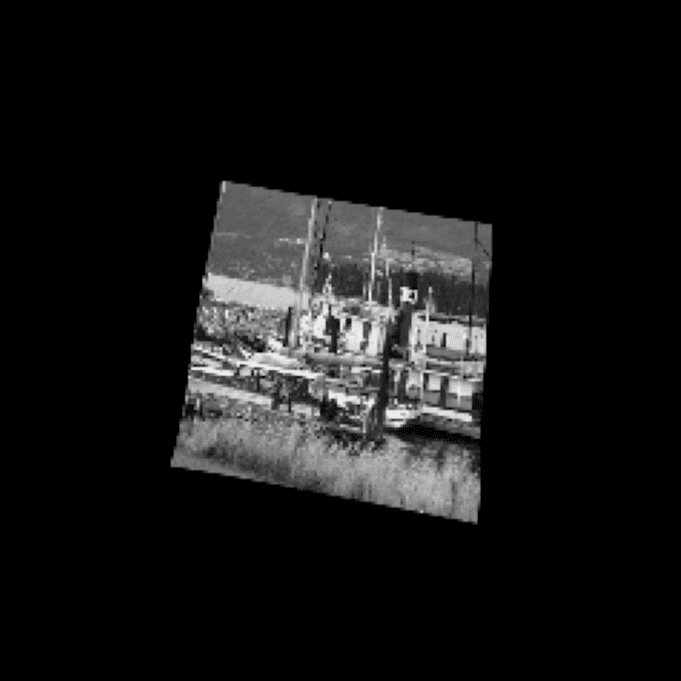}}\hfill
	\subfigure[]
	{\includegraphics[width=0.139\linewidth]{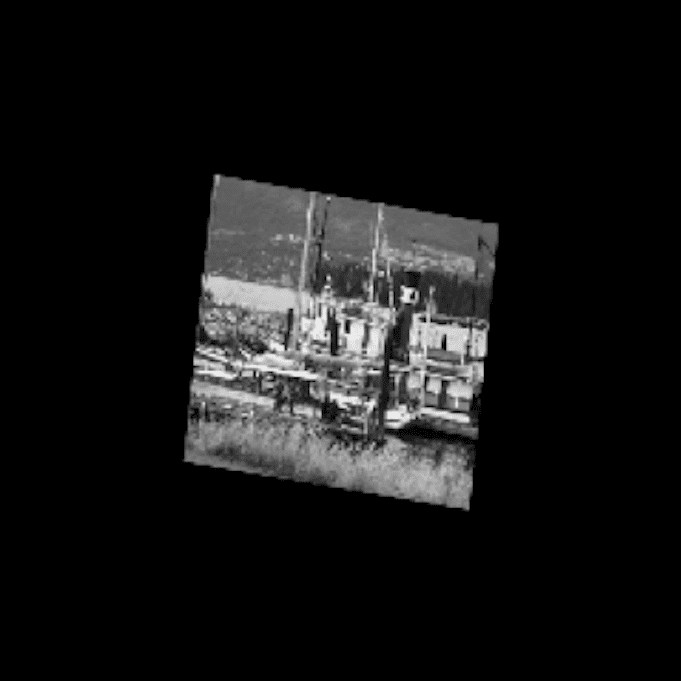}}\hfill\\
	\vspace{-21.5pt}
	\subfigure[]
	{\includegraphics[width=0.139\linewidth]{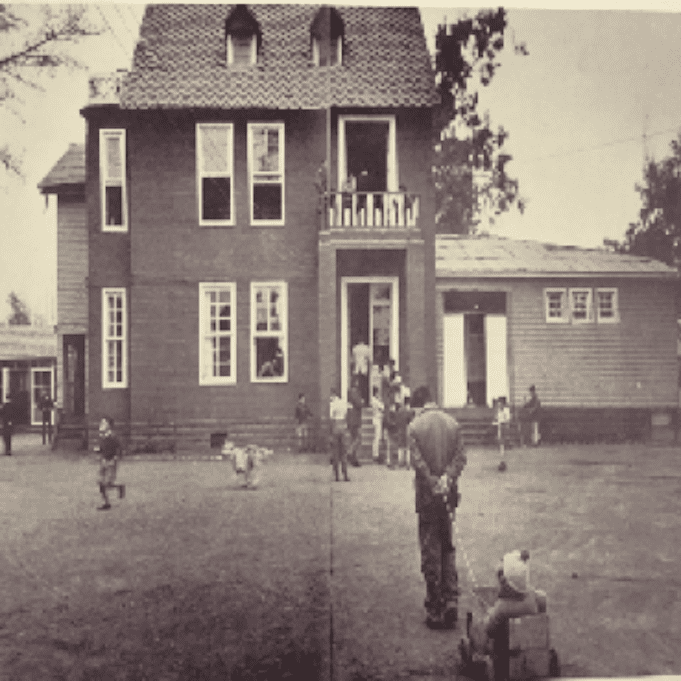}}\hfill
	\subfigure[]
	{\includegraphics[width=0.139\linewidth]{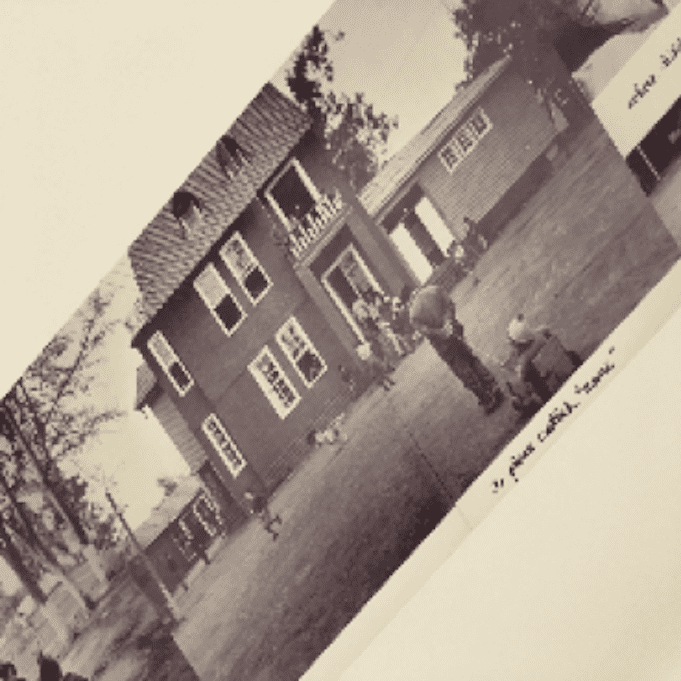}}\hfill
	\subfigure[]
	{\includegraphics[width=0.139\linewidth]{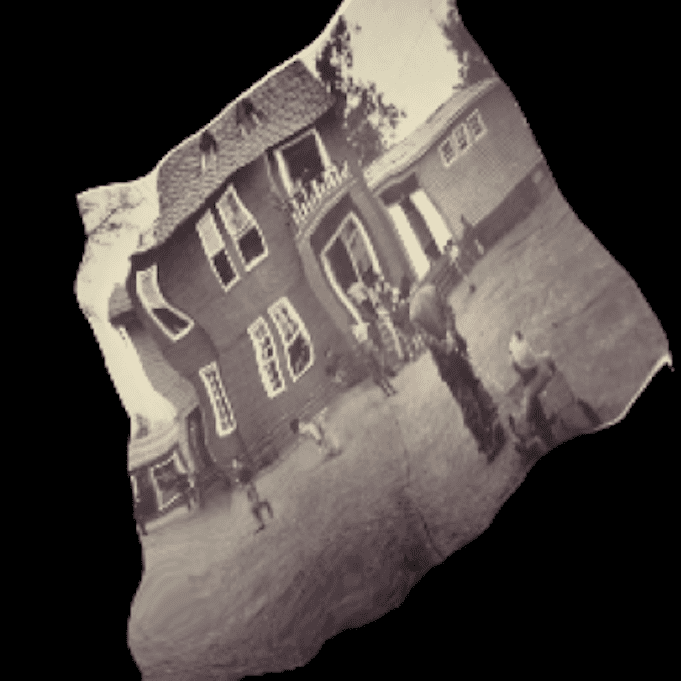}}\hfill
	\subfigure[]
	{\includegraphics[width=0.139\linewidth]{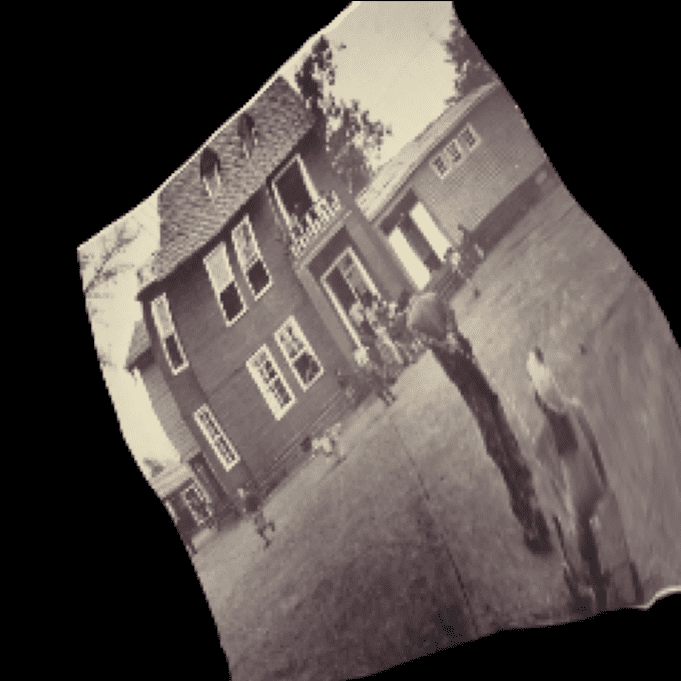}}\hfill
	\subfigure[]
	{\includegraphics[width=0.139\linewidth]{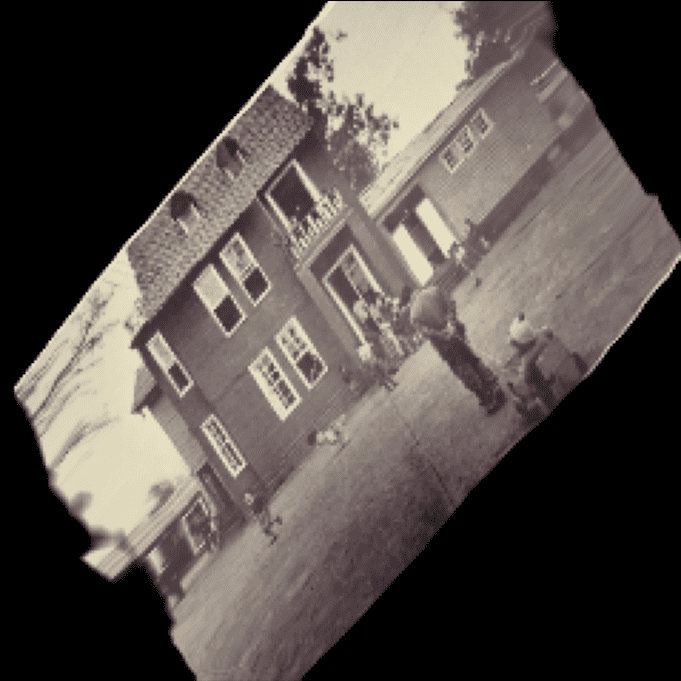}}\hfill
	\subfigure[]
	{\includegraphics[width=0.139\linewidth]{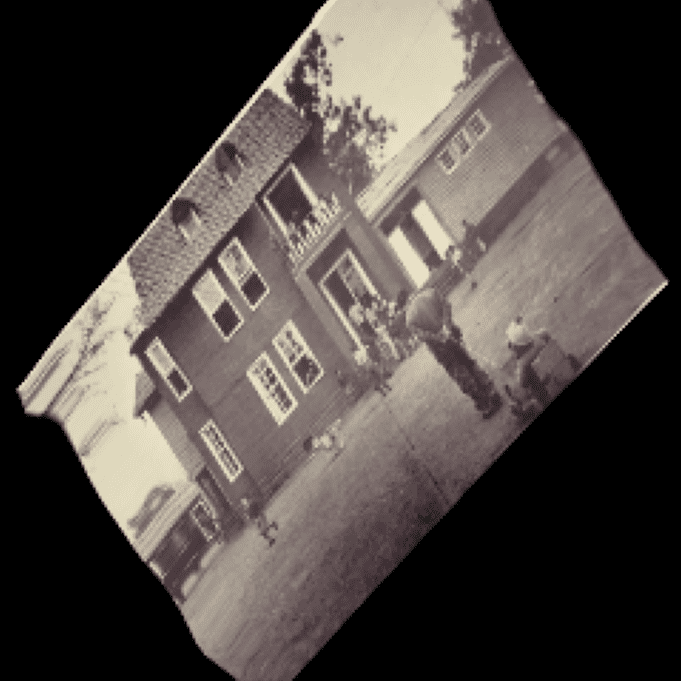}}\hfill
	\subfigure[]
	{\includegraphics[width=0.139\linewidth]{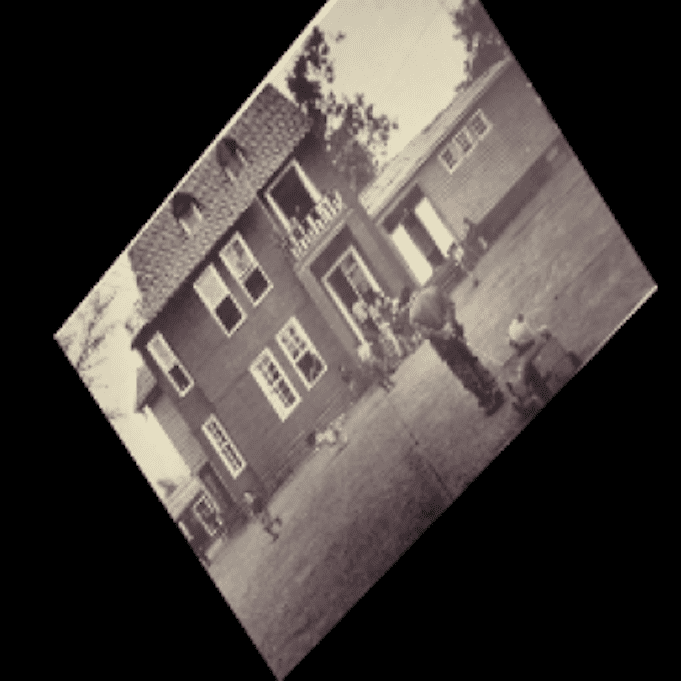}}\hfill\\
	\vspace{-21.5pt}
	\subfigure[]
	{\includegraphics[width=0.139\linewidth]{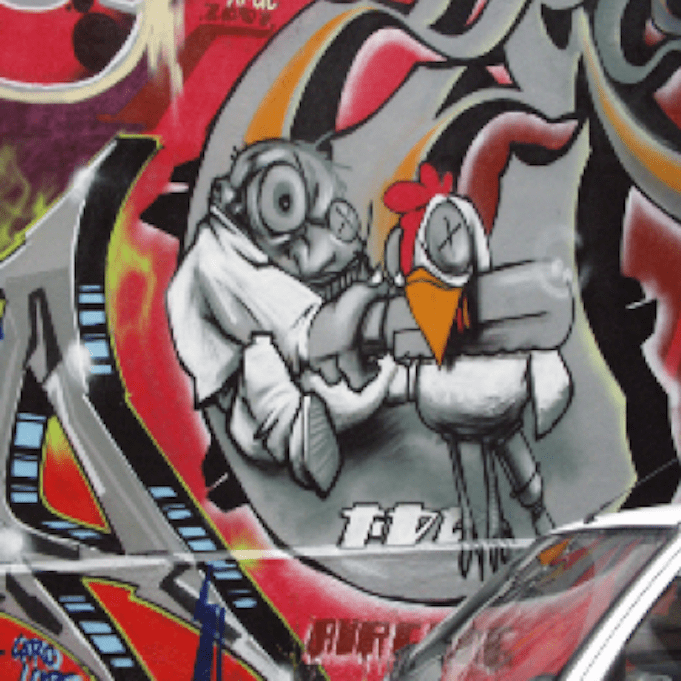}}\hfill
	\subfigure[]
	{\includegraphics[width=0.139\linewidth]{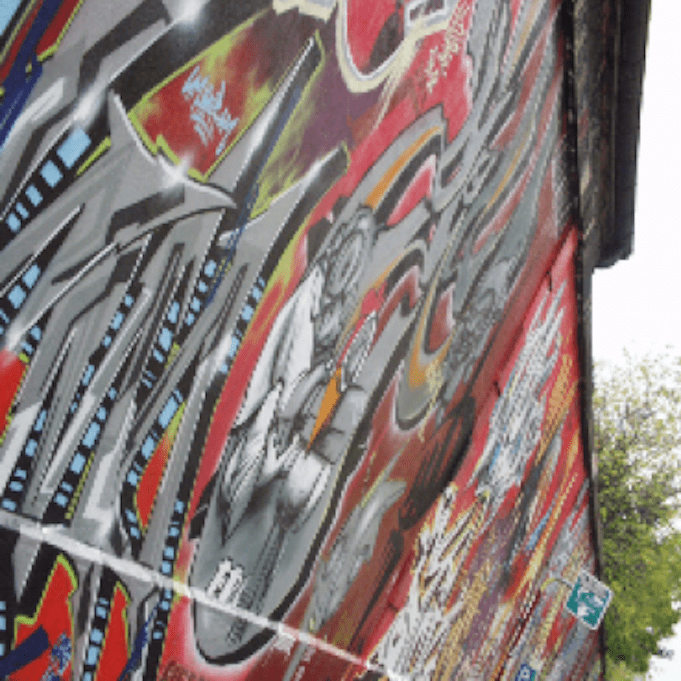}}\hfill
	\subfigure[]
	{\includegraphics[width=0.139\linewidth]{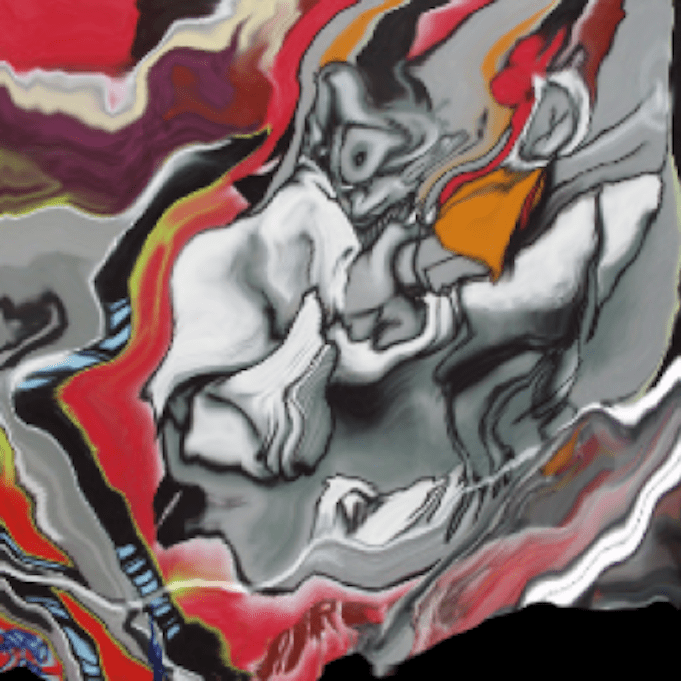}}\hfill
	\subfigure[]
	{\includegraphics[width=0.139\linewidth]{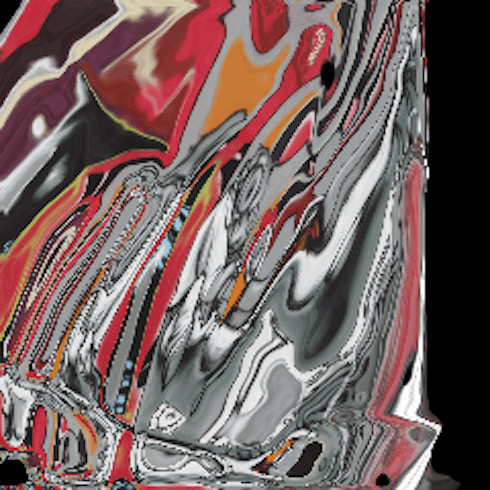}}\hfill
	\subfigure[]
	{\includegraphics[width=0.139\linewidth]{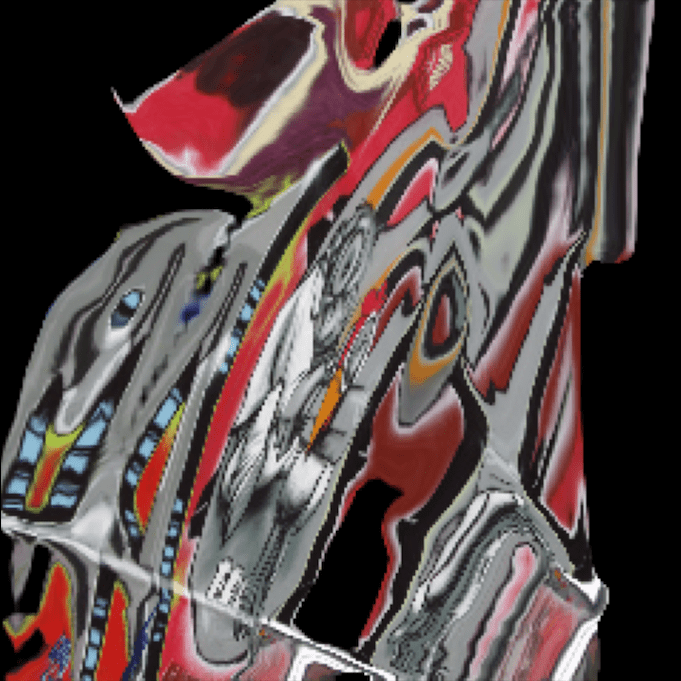}}\hfill
	\subfigure[]
	{\includegraphics[width=0.139\linewidth]{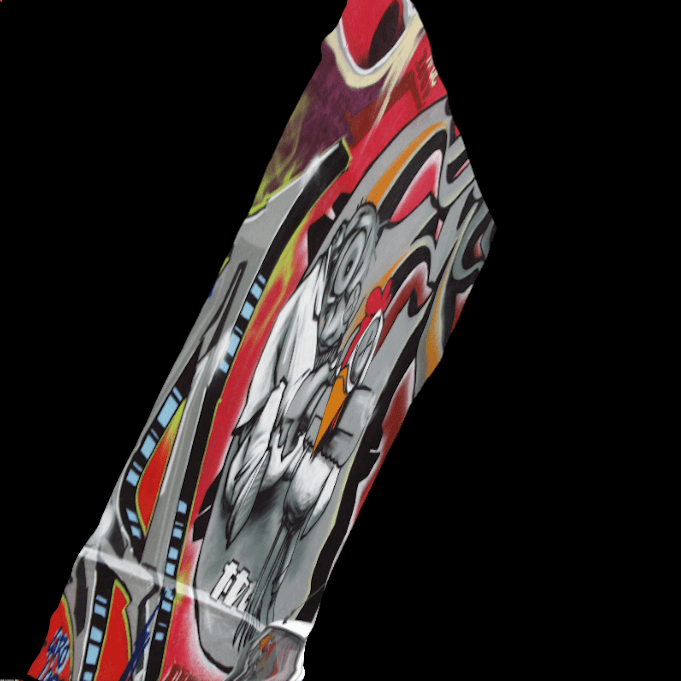}}\hfill
	\subfigure[]
	{\includegraphics[width=0.139\linewidth]{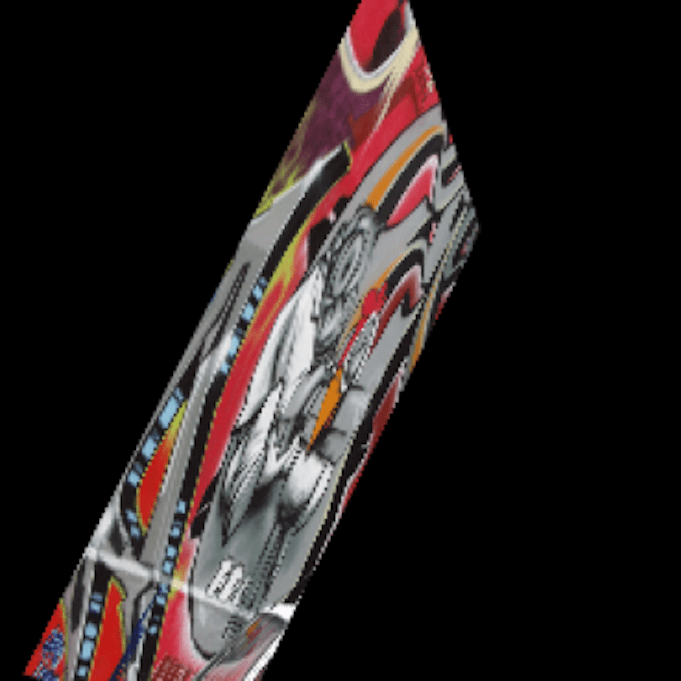}}\hfill\\
	\vspace{-21.5pt}
	\subfigure[(a)]
	{\includegraphics[width=0.139\linewidth]{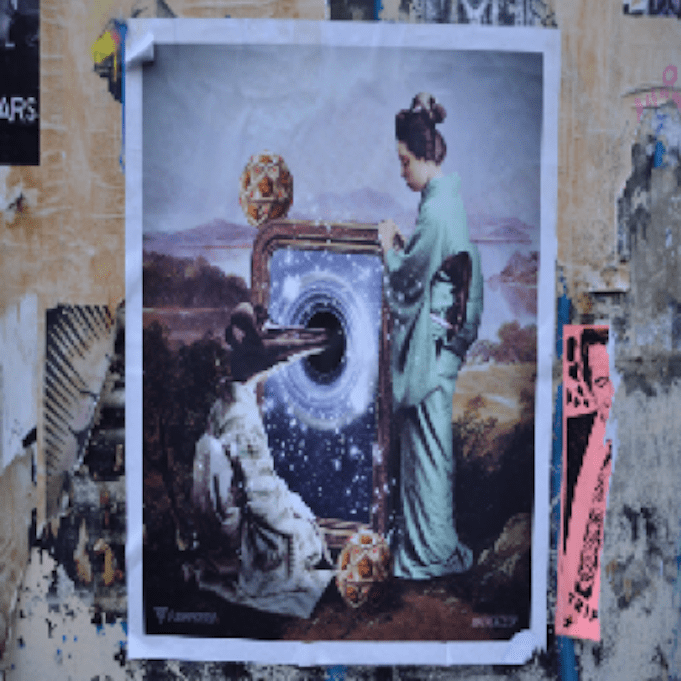}}\hfill
	\subfigure[(b)]
	{\includegraphics[width=0.139\linewidth]{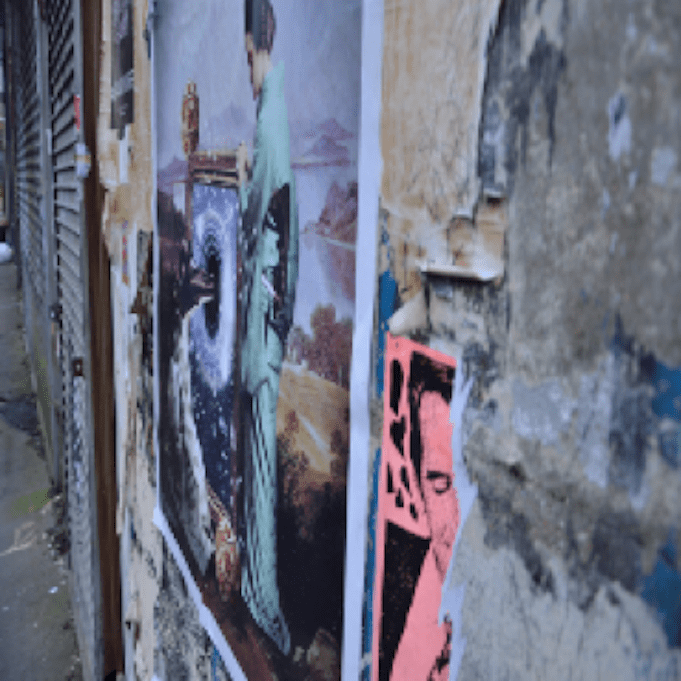}}\hfill
	\subfigure[(c)]
	{\includegraphics[width=0.139\linewidth]{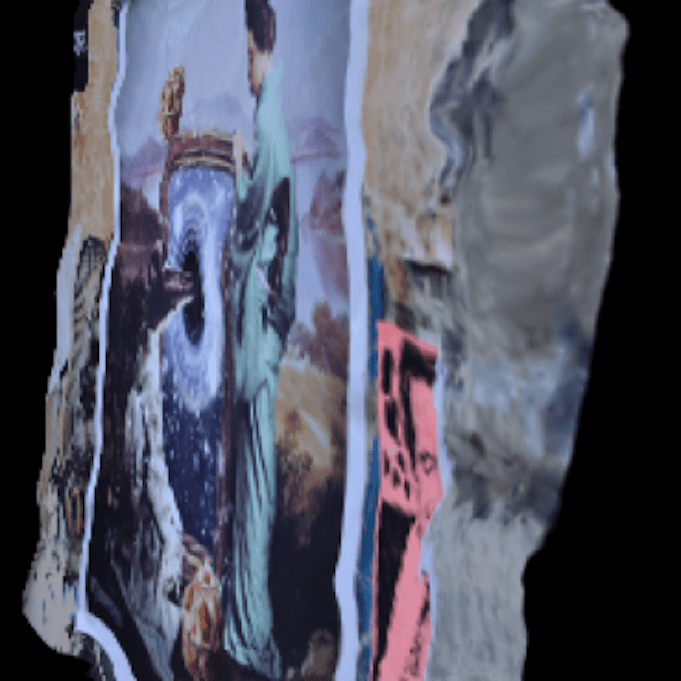}}\hfill
	\subfigure[(d)]
	{\includegraphics[width=0.139\linewidth]{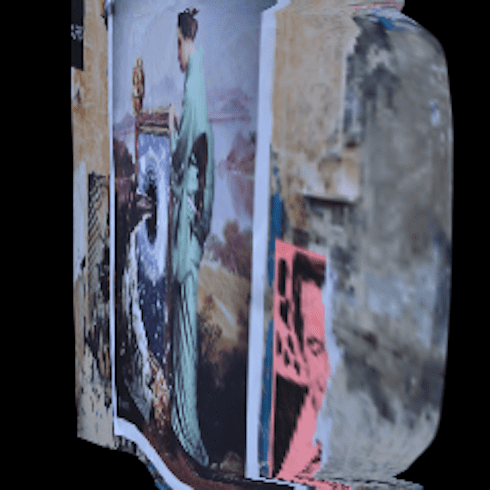}}\hfill
	\subfigure[(e)]
	{\includegraphics[width=0.139\linewidth]{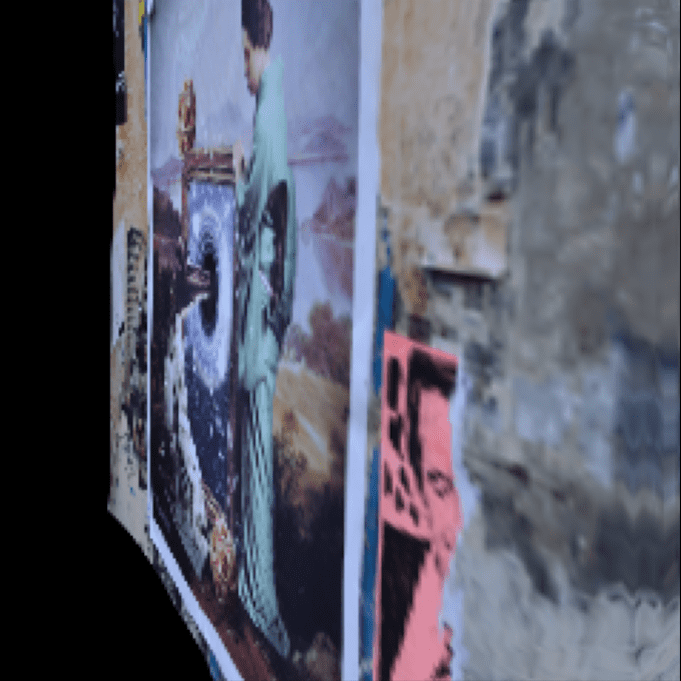}}\hfill
	\subfigure[(f)]
	{\includegraphics[width=0.139\linewidth]{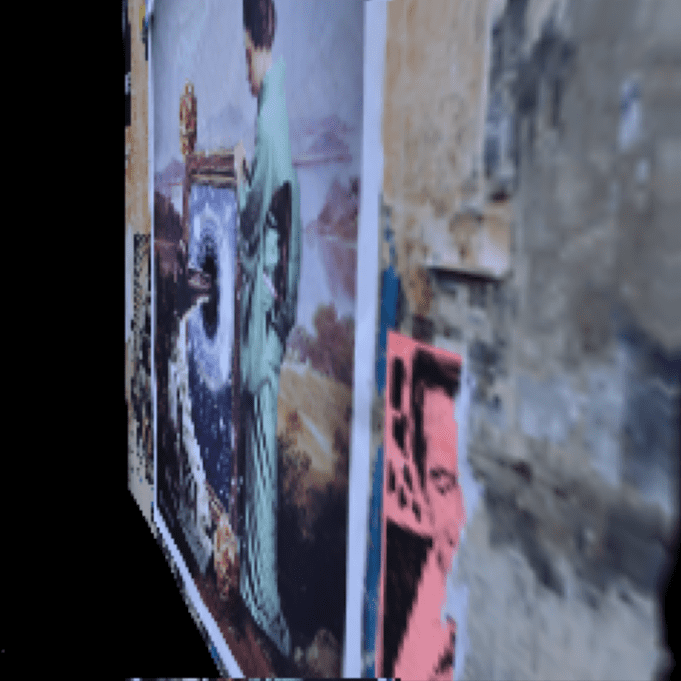}}\hfill
	\subfigure[(g)]
	{\includegraphics[width=0.139\linewidth]{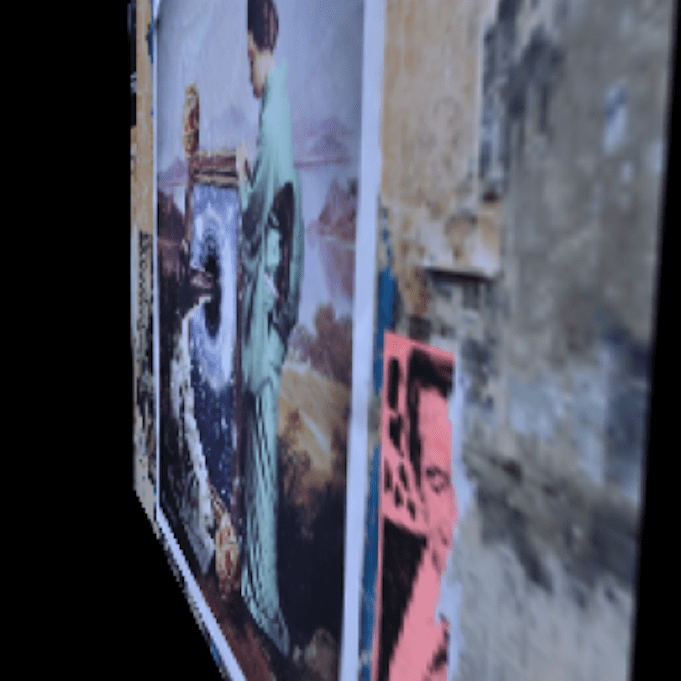}}\hfill\\
	\caption{\textbf{Qualitative results on the Hpatches benchmark~\cite{balntas2017hpatches}}. (a) source and (b) target images, warped source images using correspondences of 
		(c) GLU-Net~\cite{truong2020glu}, 
		(d) DMP, 
		(e) DMP$\dagger$, 
		(f) RANSAC-DMP, and (g) Ground-truth. Here, we provide only the samples with extremely large geometric variations to compare the outputs produced by each variants and GLU-Net. Note that DMP, starting from untrained network, achieves competitive results against GLU-Net trained on a large-scale dataset. Thanks to RANSAC~\cite{fischler1981random}, DMP starts the optimization with good initialization, which results RANSAC-DMP producing highly accurate flow fields.}\label{img:4}\vspace{-10pt}
\end{figure*}

\newpage

\begin{figure*}[h]
	\centering
	\renewcommand{\thesubfigure}{}
	\subfigure[]
	{\includegraphics[width=0.162\linewidth]{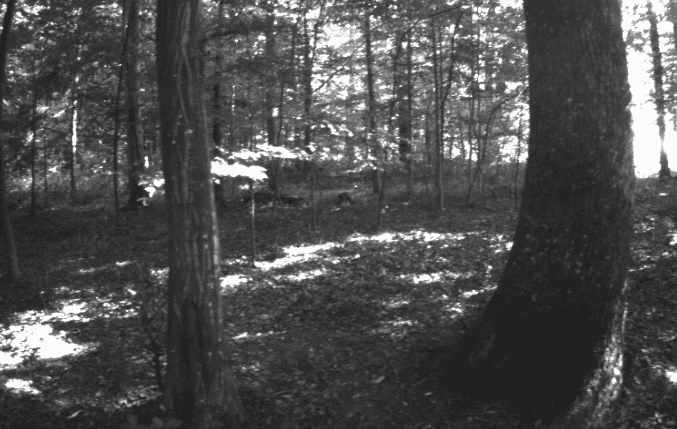}}\hfill
	\subfigure[]
	{\includegraphics[width=0.162\linewidth]{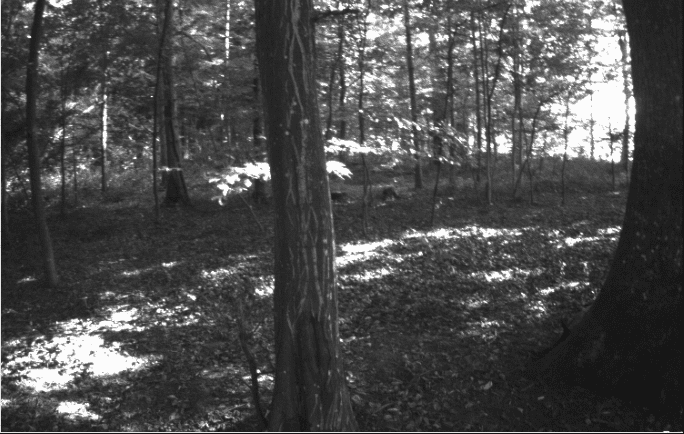}}\hfill
	\subfigure[]
	{\includegraphics[width=0.162\linewidth]{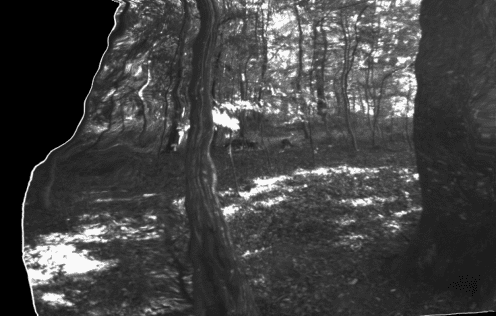}}\hfill
	\subfigure[]
	{\includegraphics[width=0.162\linewidth]{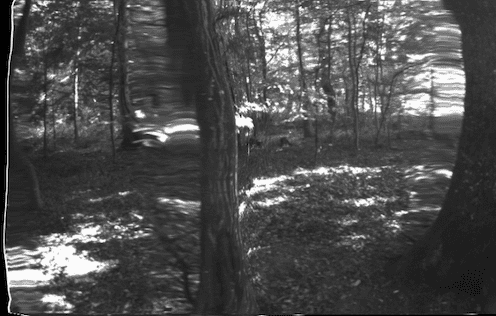}}\hfill
	\subfigure[]
	{\includegraphics[width=0.162\linewidth]{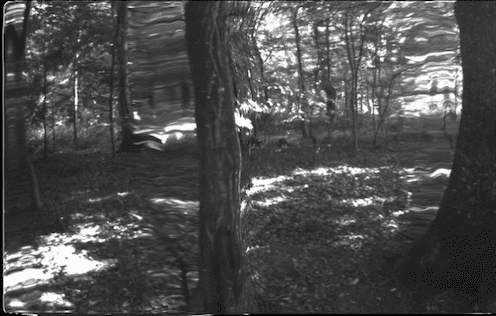}}\hfill
	\subfigure[]
	{\includegraphics[width=0.162\linewidth]{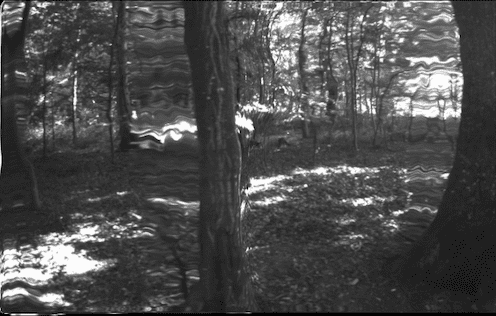}}\hfill\\
	\vspace{-21.5pt}
	\subfigure[]
	{\includegraphics[width=0.162\linewidth]{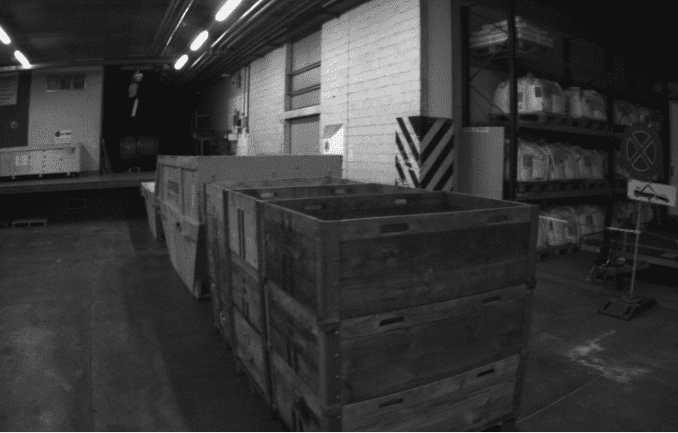}}\hfill
	\subfigure[]
	{\includegraphics[width=0.162\linewidth]{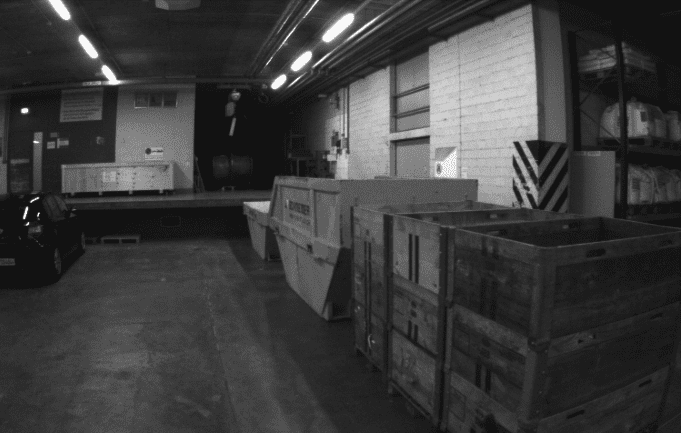}}\hfill
	\subfigure[]
	{\includegraphics[width=0.162\linewidth]{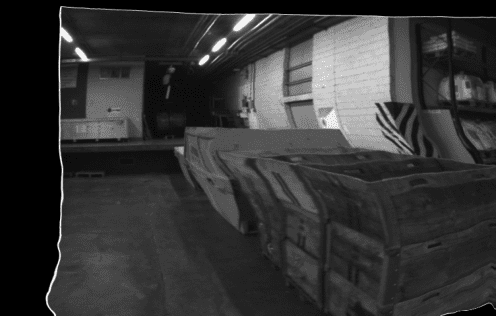}}\hfill
	\subfigure[]
	{\includegraphics[width=0.162\linewidth]{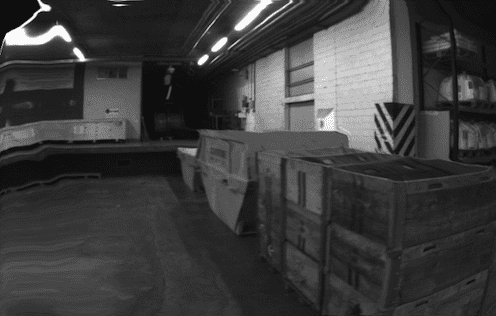}}\hfill
	\subfigure[]
	{\includegraphics[width=0.162\linewidth]{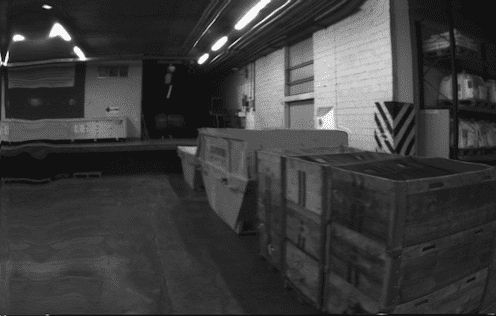}}\hfill
	\subfigure[]
	{\includegraphics[width=0.162\linewidth]{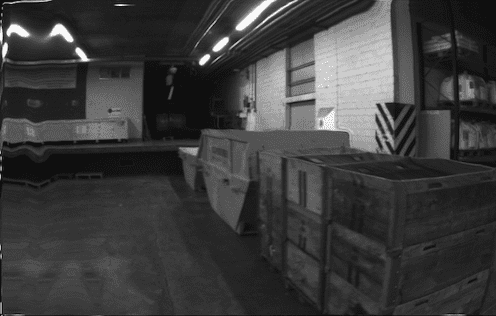}}\hfill\\
	\vspace{-21.5pt}
	\subfigure[]
	{\includegraphics[width=0.162\linewidth]{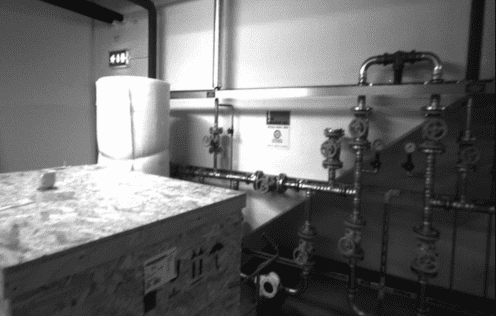}}\hfill
	\subfigure[]
	{\includegraphics[width=0.162\linewidth]{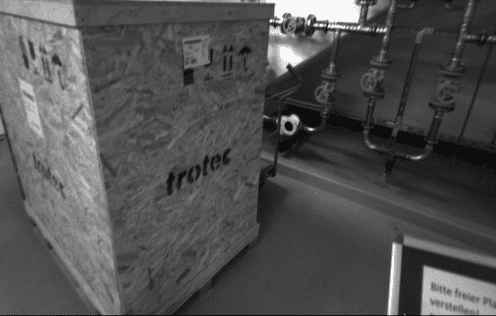}}\hfill
	\subfigure[]
	{\includegraphics[width=0.162\linewidth]{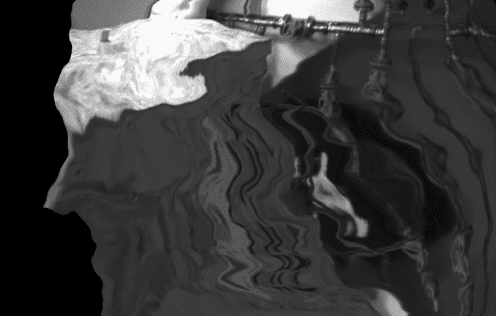}}\hfill
	\subfigure[]
	{\includegraphics[width=0.162\linewidth]{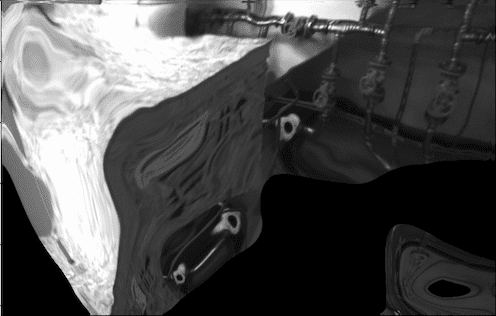}}\hfill
	\subfigure[]
	{\includegraphics[width=0.162\linewidth]{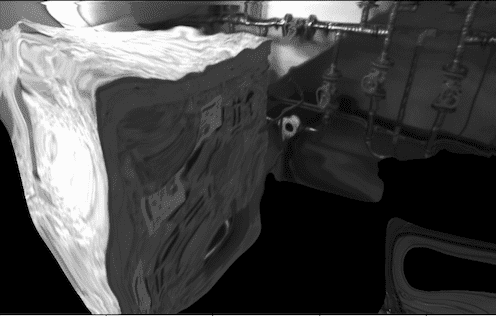}}\hfill
	\subfigure[]
	{\includegraphics[width=0.162\linewidth]{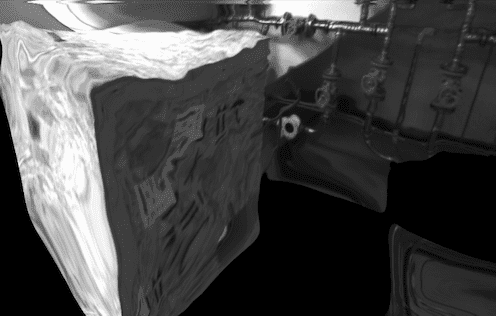}}\hfill\\
	\vspace{-21.5pt}
	\subfigure[]
	{\includegraphics[width=0.162\linewidth]{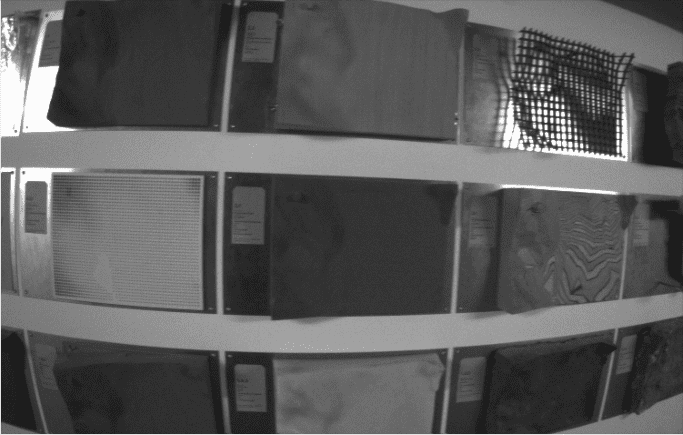}}\hfill
	\subfigure[]
	{\includegraphics[width=0.162\linewidth]{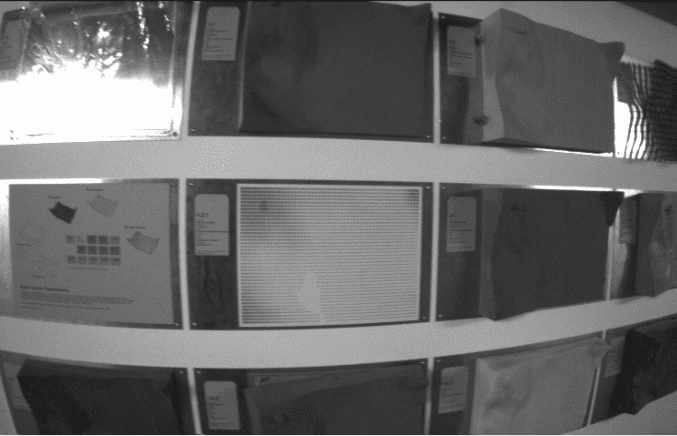}}\hfill
	\subfigure[]
	{\includegraphics[width=0.162\linewidth]{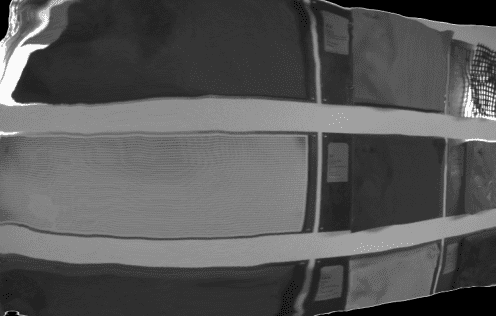}}\hfill
	\subfigure[]
	{\includegraphics[width=0.162\linewidth]{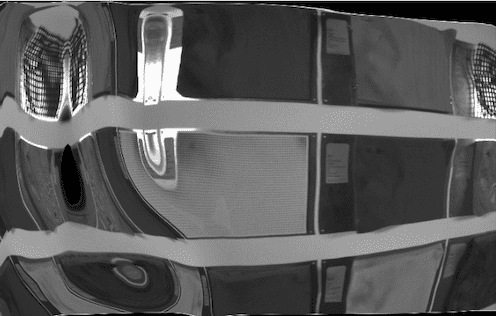}}\hfill
	\subfigure[]
	{\includegraphics[width=0.162\linewidth]{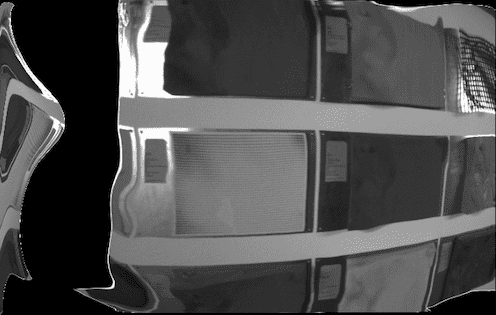}}\hfill
	\subfigure[]
	{\includegraphics[width=0.162\linewidth]{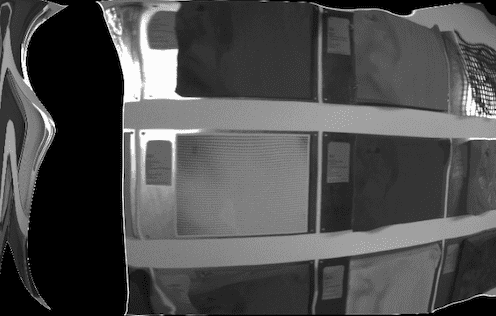}}\hfill\\
	\vspace{-21.5pt}
	\subfigure[]
	{\includegraphics[width=0.162\linewidth]{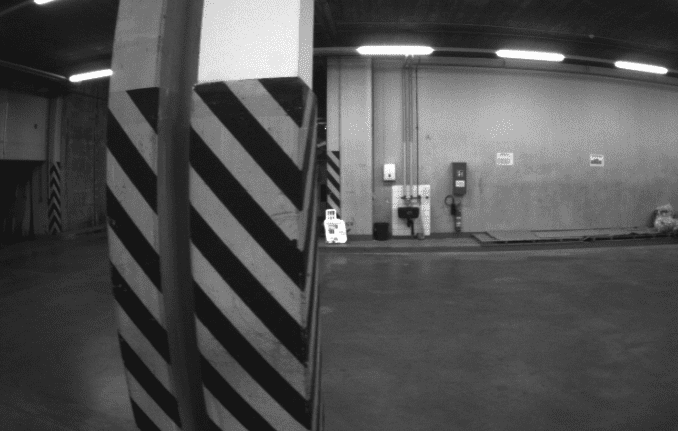}}\hfill
	\subfigure[]
	{\includegraphics[width=0.162\linewidth]{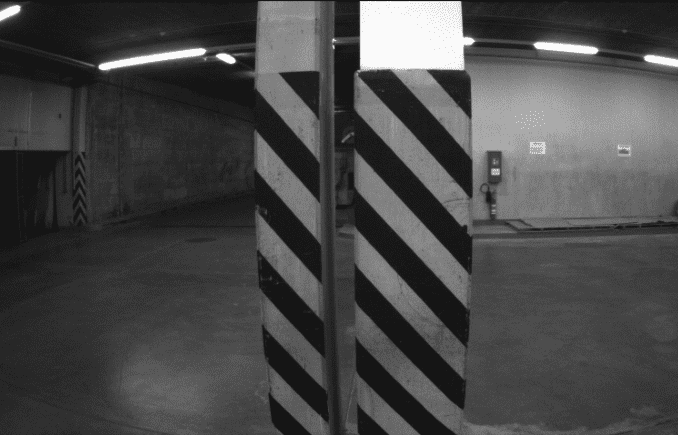}}\hfill
	\subfigure[]
	{\includegraphics[width=0.162\linewidth]{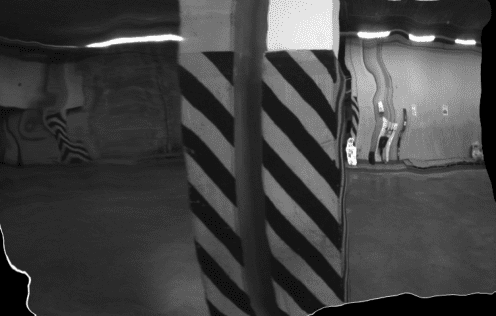}}\hfill
	\subfigure[]
	{\includegraphics[width=0.162\linewidth]{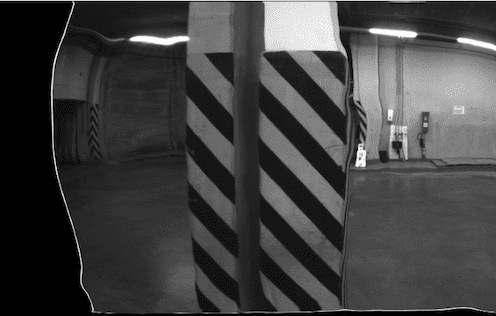}}\hfill
	\subfigure[]
	{\includegraphics[width=0.162\linewidth]{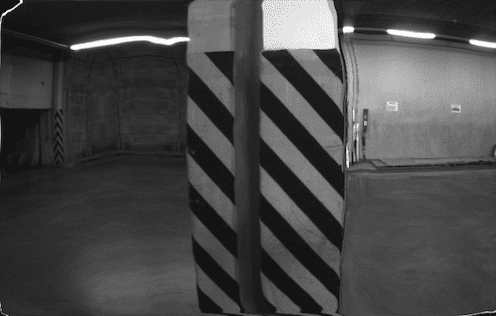}}\hfill
	\subfigure[]
	{\includegraphics[width=0.162\linewidth]{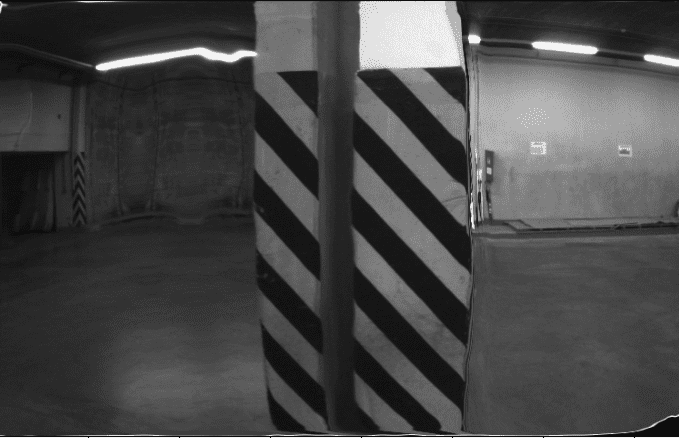}}\hfill\\
	\vspace{-21.5pt}
	\subfigure[]
	{\includegraphics[width=0.162\linewidth]{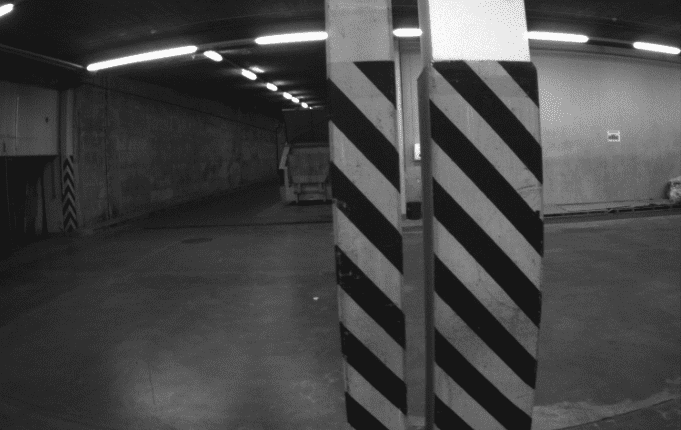}}\hfill
	\subfigure[]
	{\includegraphics[width=0.162\linewidth]{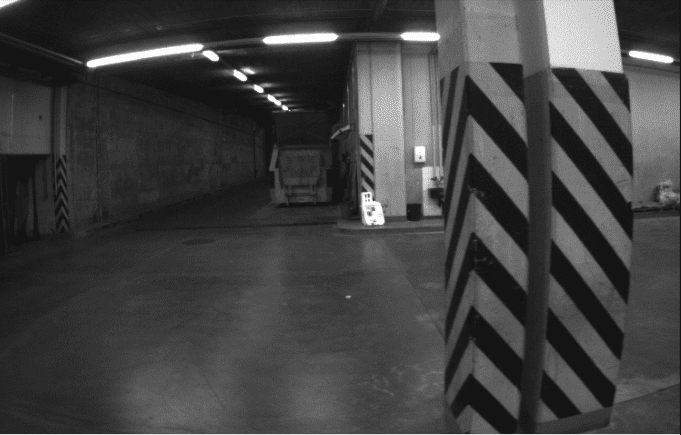}}\hfill
	\subfigure[]
	{\includegraphics[width=0.162\linewidth]{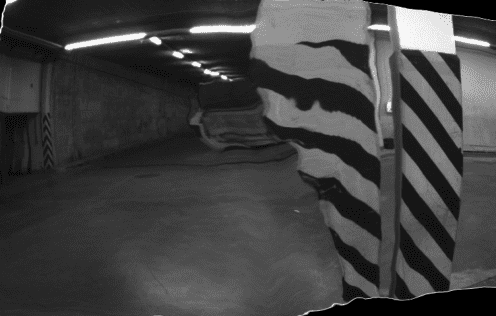}}\hfill
	\subfigure[]
	{\includegraphics[width=0.162\linewidth]{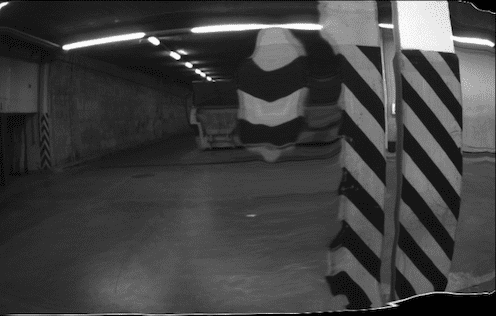}}\hfill
	\subfigure[]
	{\includegraphics[width=0.162\linewidth]{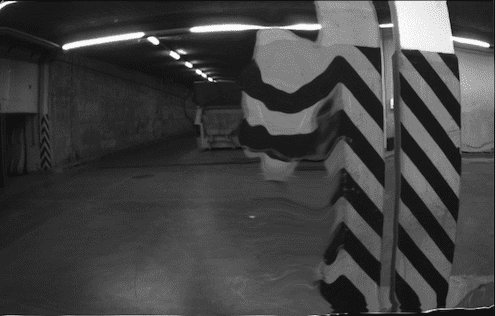}}\hfill
	\subfigure[]
	{\includegraphics[width=0.162\linewidth]{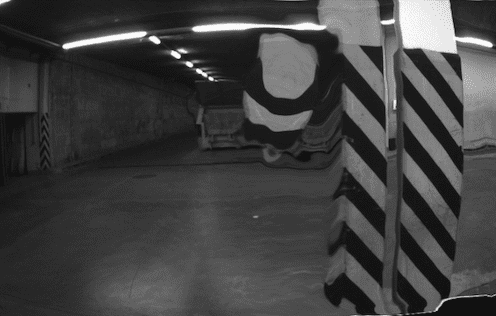}}\hfill\\
	\vspace{-21.5pt}
	\subfigure[(a)]
	{\includegraphics[width=0.162\linewidth]{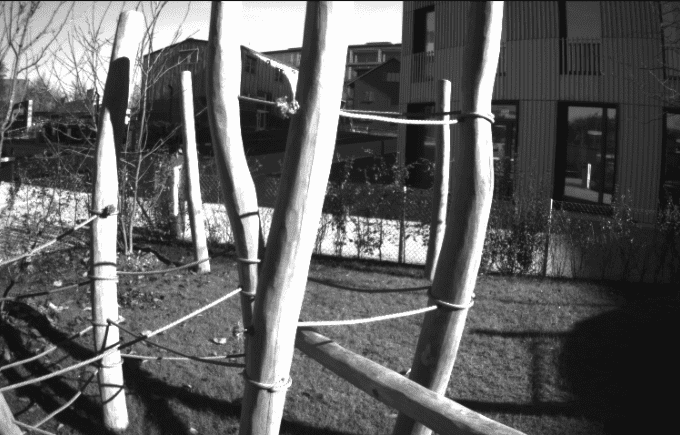}}\hfill
	\subfigure[(b)]
	{\includegraphics[width=0.162\linewidth]{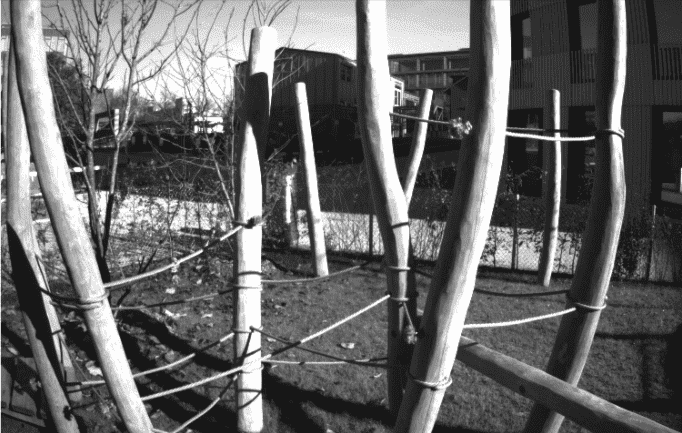}}\hfill
	\subfigure[(c)]
	{\includegraphics[width=0.162\linewidth]{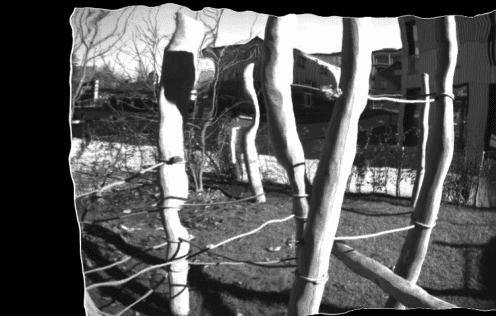}}\hfill
	\subfigure[(d)]
	{\includegraphics[width=0.162\linewidth]{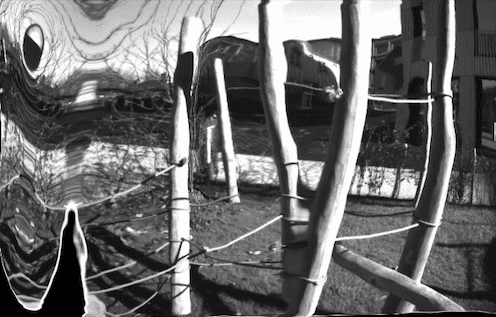}}\hfill
	\subfigure[(e)]
	{\includegraphics[width=0.162\linewidth]{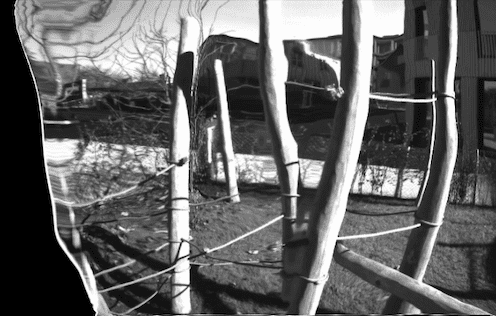}}\hfill
	\subfigure[(f)]
	{\includegraphics[width=0.162\linewidth]{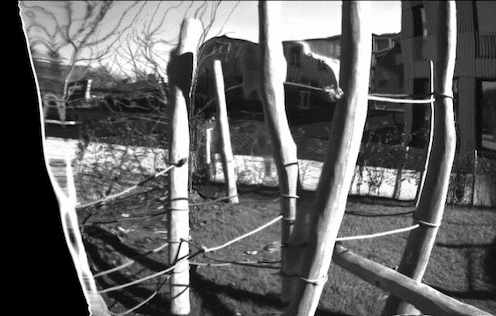}}\hfill\\
	\caption{Qualitative results on the ETH3D benchmark~\cite{schops2017multi}: (a) source and (b) target images, warped source images using correspondences of (c) GLU-Net~\cite{truong2020glu}, (d) DMP, (e) A-DMP, and (f) DMP$\dagger$ . Note that our loss function allows error correction, allowing more optimal estimation of flow fields.  }\label{img:5}\vspace{-10pt}
\end{figure*}

\newpage

\begin{figure*}[h]
	\centering
	\renewcommand{\thesubfigure}{}
	\subfigure[]
	{\includegraphics[width=0.162\linewidth]{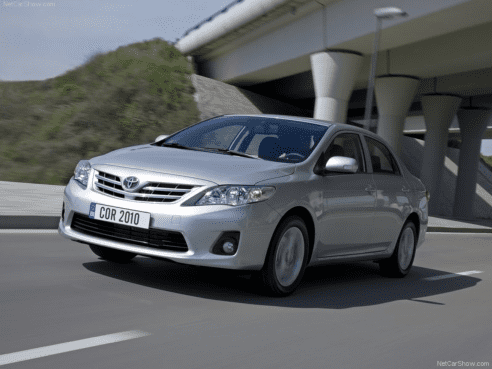}}\hfill
	\subfigure[]
	{\includegraphics[width=0.162\linewidth]{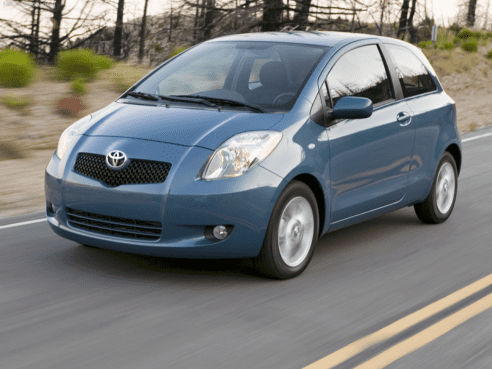}}\hfill
	\subfigure[]
	{\includegraphics[width=0.162\linewidth]{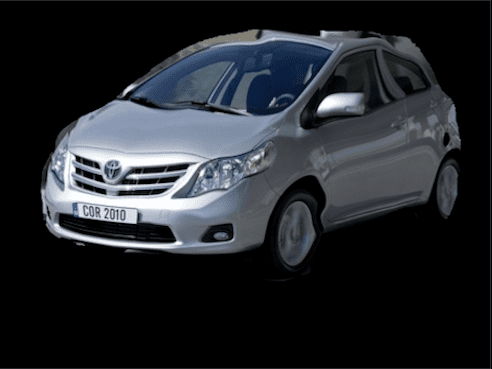}}\hfill
	\subfigure[]
	{\includegraphics[width=0.162\linewidth]{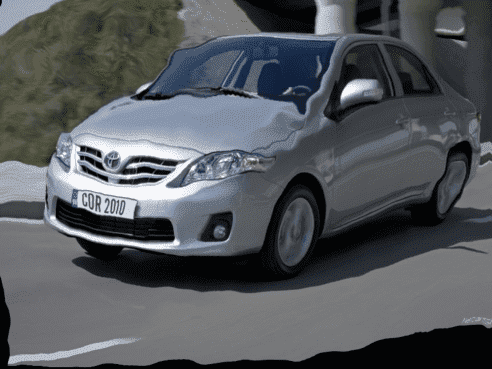}}\hfill
	\subfigure[]
	{\includegraphics[width=0.162\linewidth]{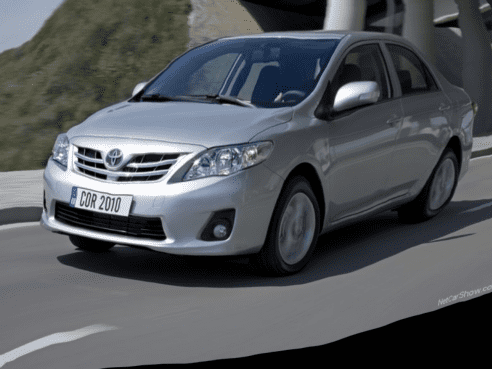}}\hfill
	\subfigure[]
	{\includegraphics[width=0.162\linewidth]{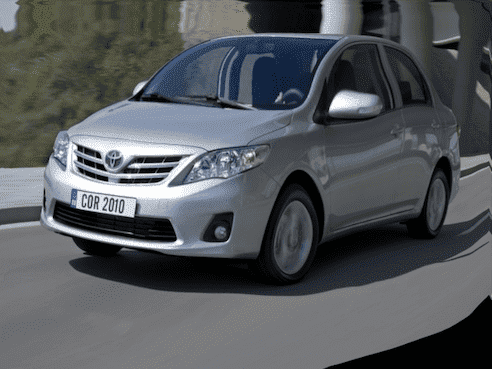}}\hfill\\
	\vspace{-21.5pt}
	\subfigure[]
	{\includegraphics[width=0.162\linewidth]{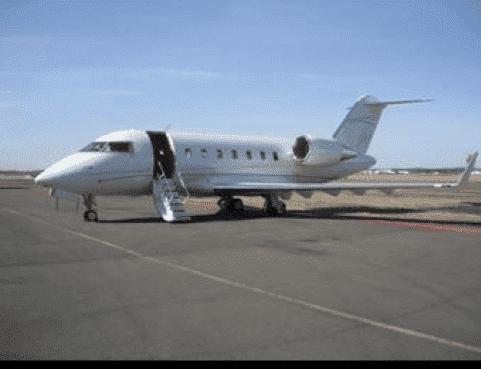}}\hfill
	\subfigure[]
	{\includegraphics[width=0.162\linewidth]{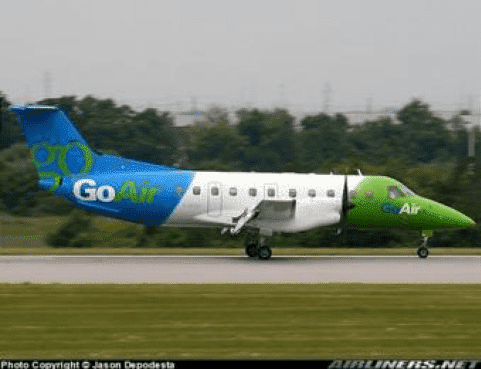}}\hfill
	\subfigure[]
	{\includegraphics[width=0.162\linewidth]{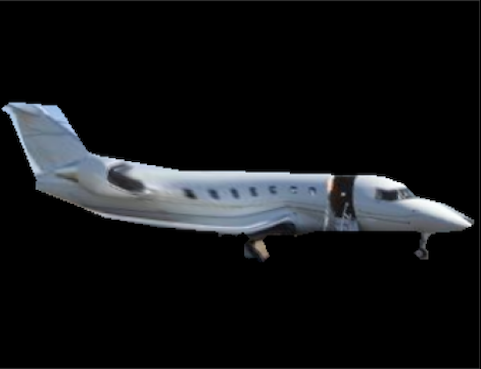}}\hfill
	\subfigure[]
	{\includegraphics[width=0.162\linewidth]{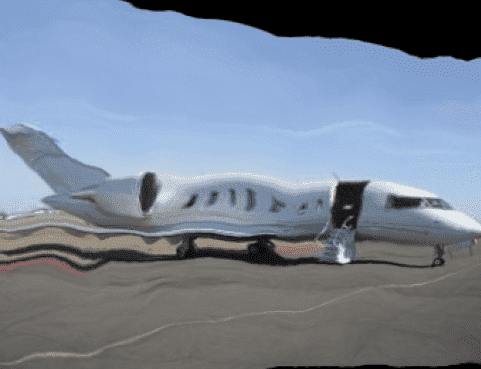}}\hfill
	\subfigure[]
	{\includegraphics[width=0.162\linewidth]{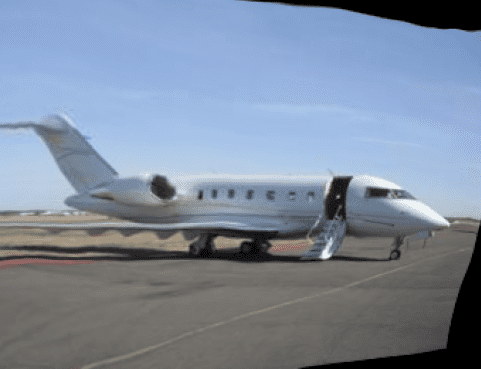}}\hfill
	\subfigure[]
	{\includegraphics[width=0.162\linewidth]{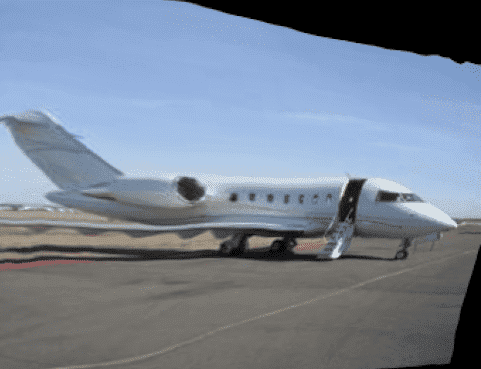}}\hfill\\
	\vspace{-21.5pt}
	\subfigure[]
	{\includegraphics[width=0.162\linewidth]{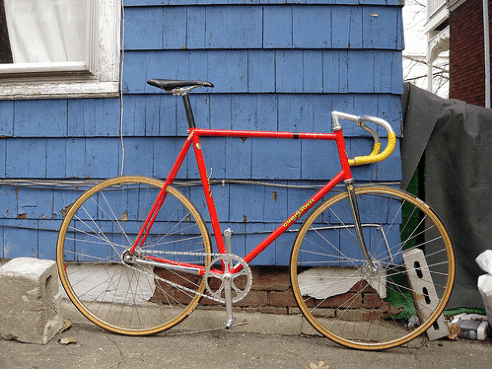}}\hfill
	\subfigure[]
	{\includegraphics[width=0.162\linewidth]{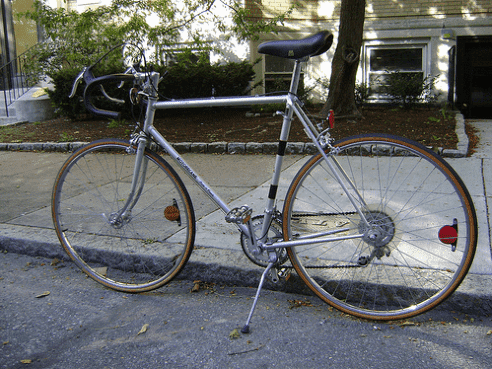}}\hfill
	\subfigure[]
	{\includegraphics[width=0.162\linewidth]{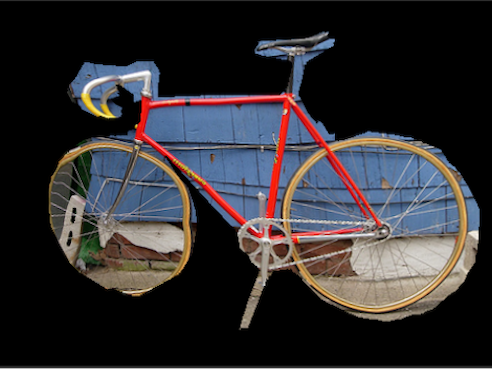}}\hfill
	\subfigure[]
	{\includegraphics[width=0.162\linewidth]{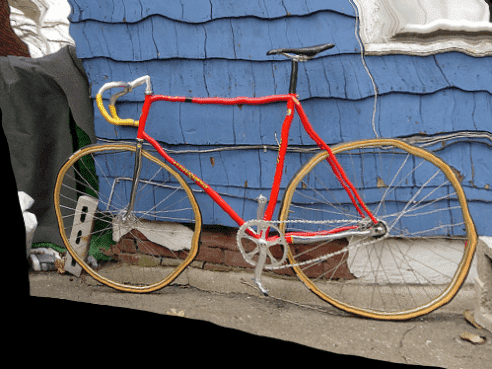}}\hfill
	\subfigure[]
	{\includegraphics[width=0.162\linewidth]{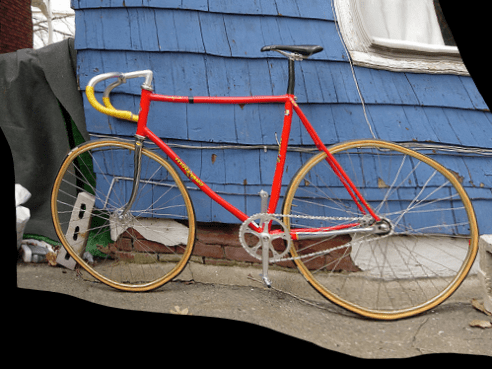}}\hfill
	\subfigure[]
	{\includegraphics[width=0.162\linewidth]{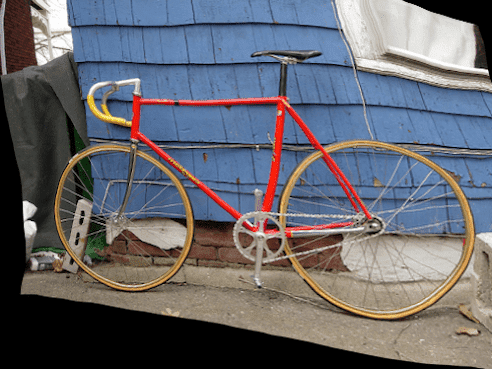}}\hfill\\
	\vspace{-21.5pt}
	\subfigure[]
	{\includegraphics[width=0.162\linewidth]{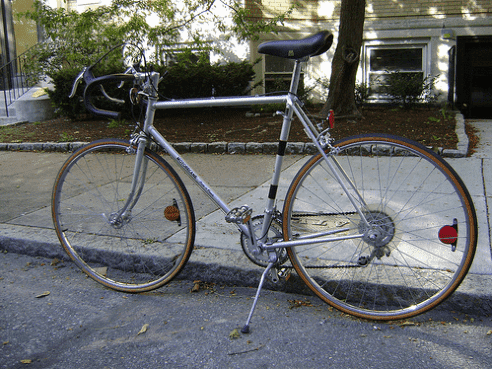}}\hfill
	\subfigure[]
	{\includegraphics[width=0.162\linewidth]{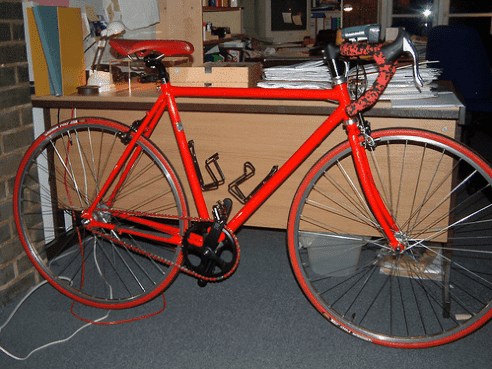}}\hfill
	\subfigure[]
	{\includegraphics[width=0.162\linewidth]{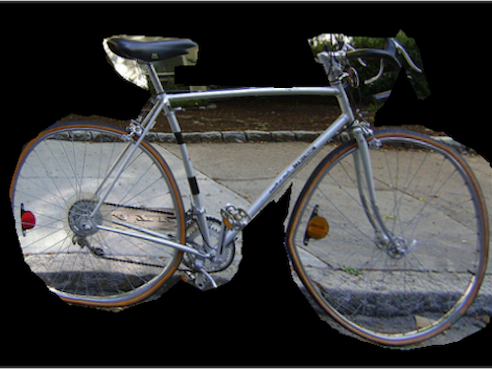}}\hfill
	\subfigure[]
	{\includegraphics[width=0.162\linewidth]{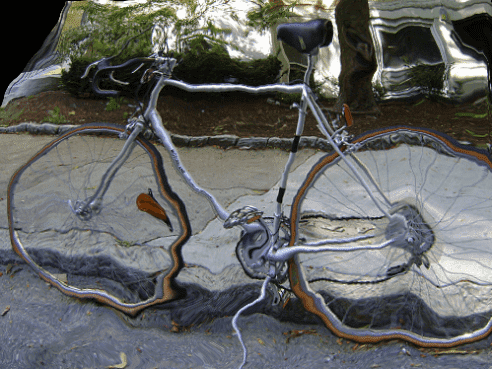}}\hfill
	\subfigure[]
	{\includegraphics[width=0.162\linewidth]{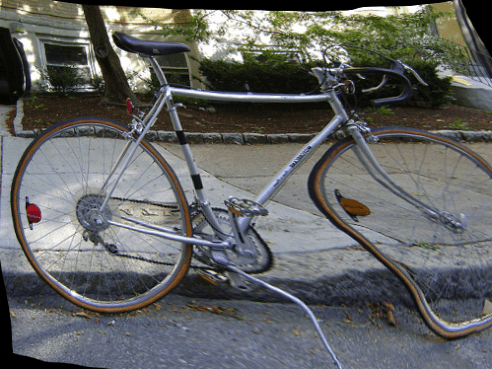}}\hfill
	\subfigure[]
	{\includegraphics[width=0.162\linewidth]{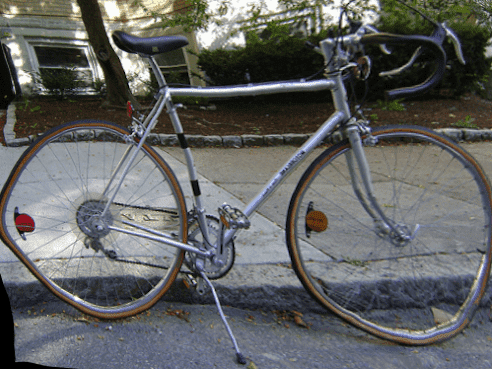}}\hfill\\
	\vspace{-21.5pt}
	\subfigure[]
	{\includegraphics[width=0.162\linewidth]{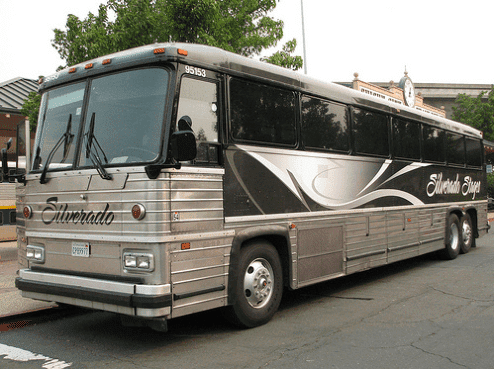}}\hfill
	\subfigure[]
	{\includegraphics[width=0.162\linewidth]{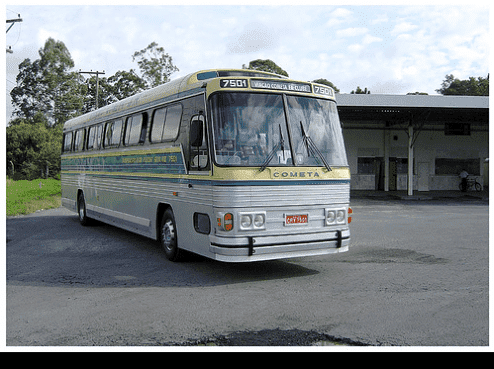}}\hfill
	\subfigure[]
	{\includegraphics[width=0.162\linewidth]{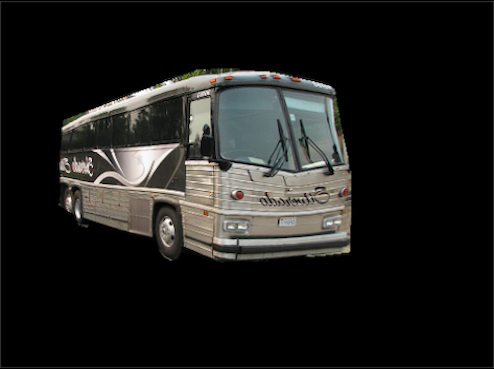}}\hfill
	\subfigure[]
	{\includegraphics[width=0.162\linewidth]{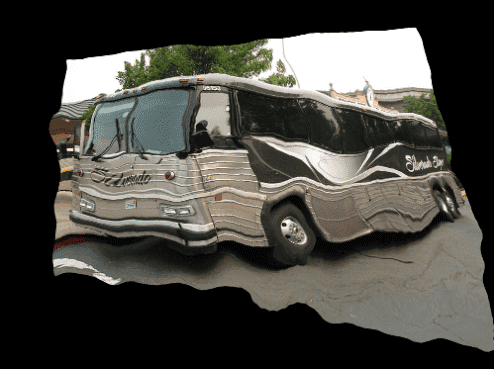}}\hfill
	\subfigure[]
	{\includegraphics[width=0.162\linewidth]{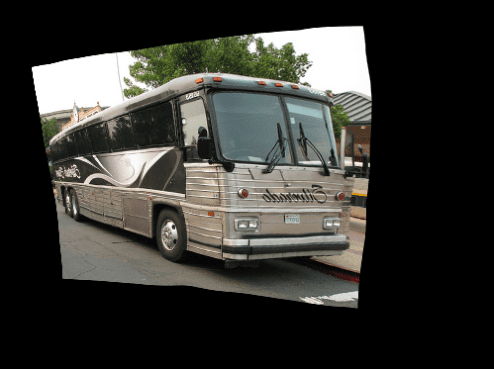}}\hfill
	\subfigure[]
	{\includegraphics[width=0.162\linewidth]{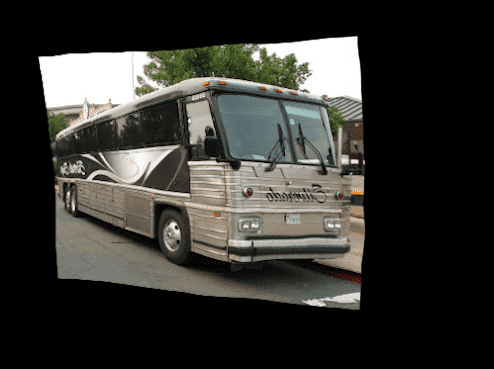}}\hfill\\
	\vspace{-21.5pt}
	\subfigure[]
	{\includegraphics[width=0.162\linewidth]{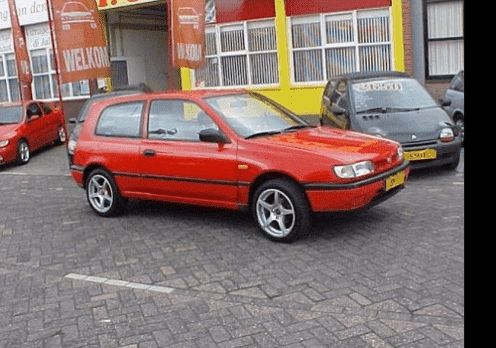}}\hfill
	\subfigure[]
	{\includegraphics[width=0.162\linewidth]{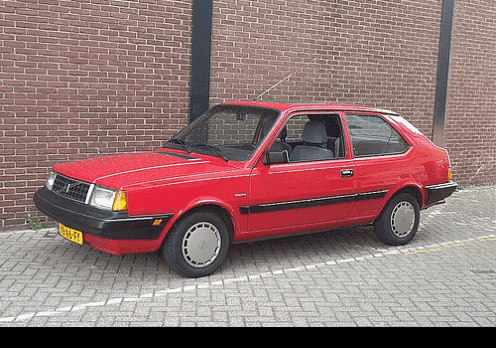}}\hfill
	\subfigure[]
	{\includegraphics[width=0.162\linewidth]{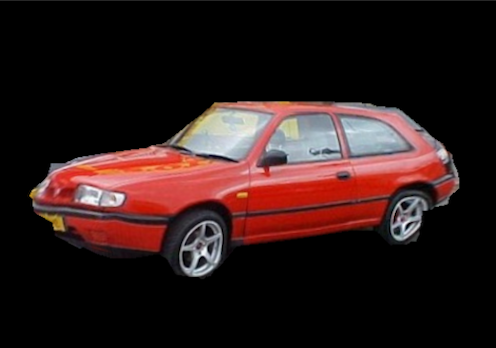}}\hfill
	\subfigure[]
	{\includegraphics[width=0.162\linewidth]{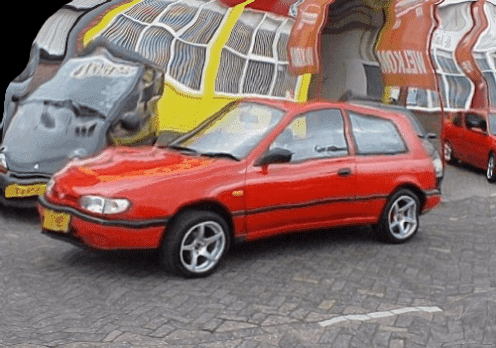}}\hfill
	\subfigure[]
	{\includegraphics[width=0.162\linewidth]{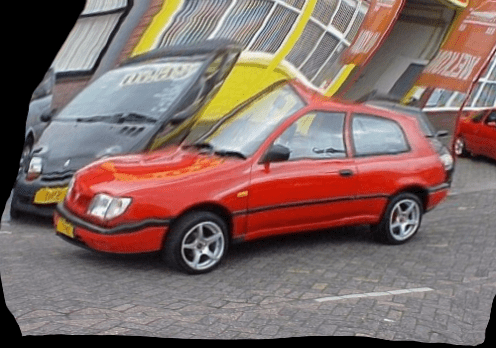}}\hfill
	\subfigure[]
	{\includegraphics[width=0.162\linewidth]{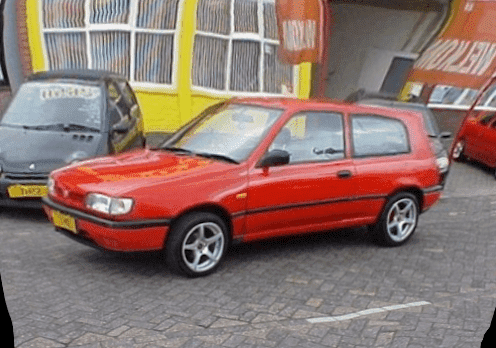}}\hfill\\
	\vspace{-21.5pt}
	\subfigure[]
	{\includegraphics[width=0.162\linewidth]{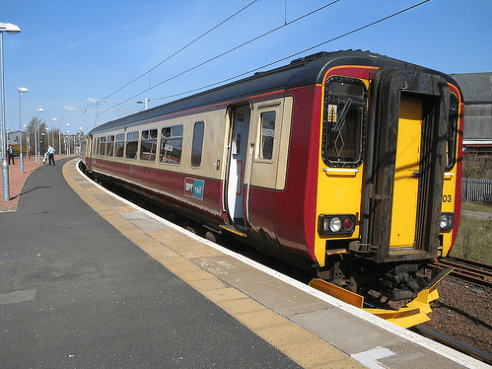}}\hfill
	\subfigure[]
	{\includegraphics[width=0.162\linewidth]{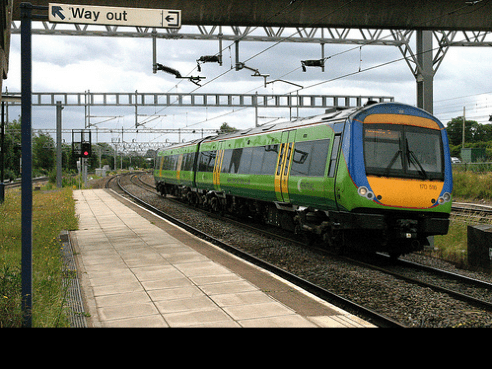}}\hfill
	\subfigure[]
	{\includegraphics[width=0.162\linewidth]{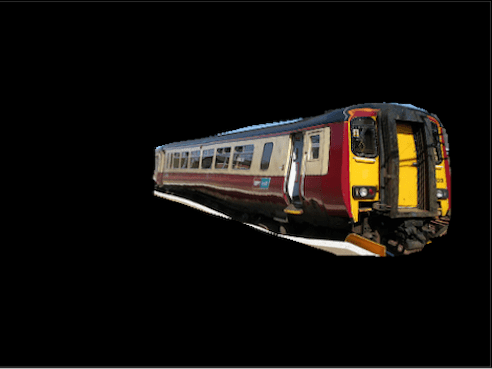}}\hfill
	\subfigure[]
	{\includegraphics[width=0.162\linewidth]{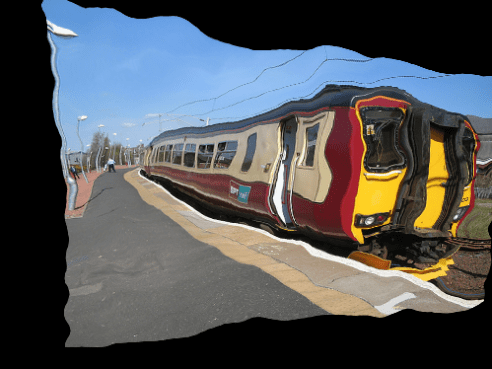}}\hfill
	\subfigure[]
	{\includegraphics[width=0.162\linewidth]{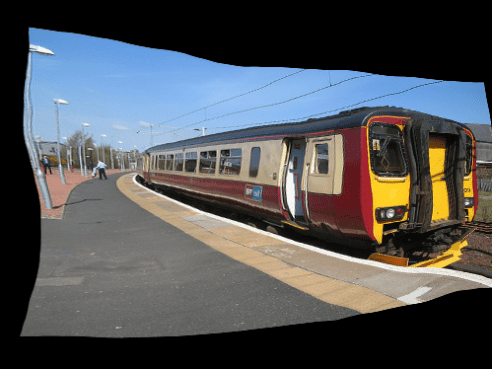}}\hfill
	\subfigure[]
	{\includegraphics[width=0.162\linewidth]{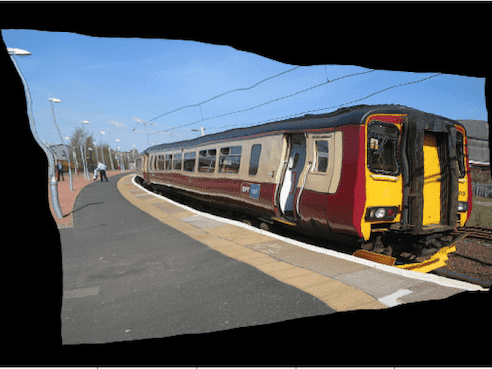}}\hfill\\
	\vspace{-21.5pt}
	\subfigure[]
	{\includegraphics[width=0.162\linewidth]{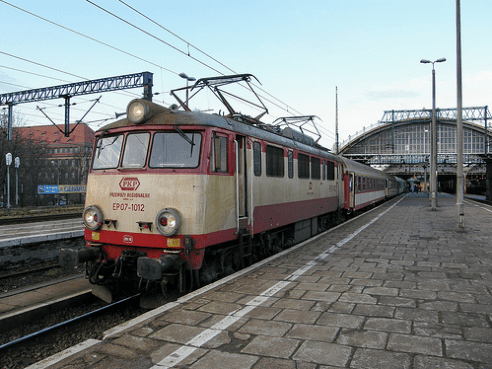}}\hfill
	\subfigure[]
	{\includegraphics[width=0.162\linewidth]{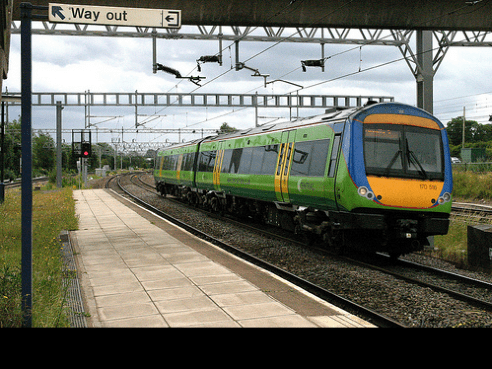}}\hfill
	\subfigure[]
	{\includegraphics[width=0.162\linewidth]{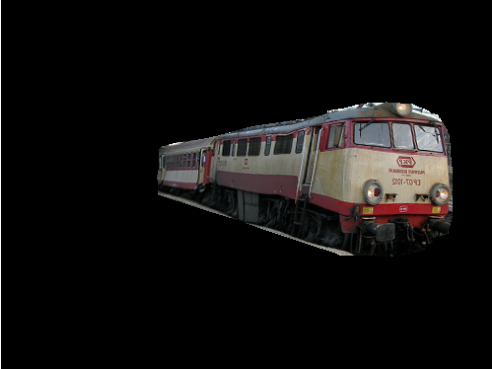}}\hfill
	\subfigure[]
	{\includegraphics[width=0.162\linewidth]{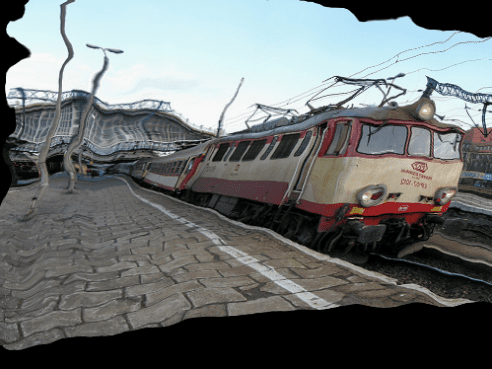}}\hfill
	\subfigure[]
	{\includegraphics[width=0.162\linewidth]{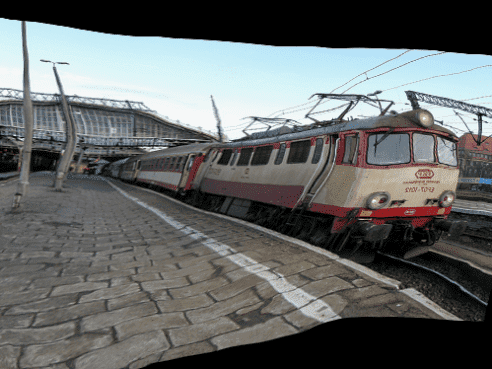}}\hfill
	\subfigure[]
	{\includegraphics[width=0.162\linewidth]{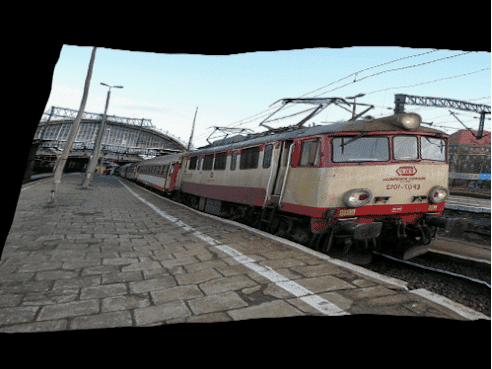}}\hfill\\
	\vspace{-21.5pt}
	\subfigure[(a)]
	{\includegraphics[width=0.162\linewidth]{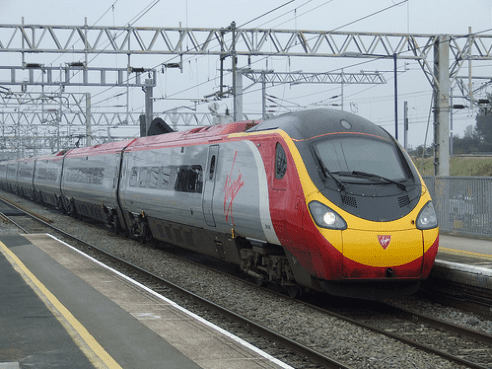}}\hfill
	\subfigure[(b)]
	{\includegraphics[width=0.162\linewidth]{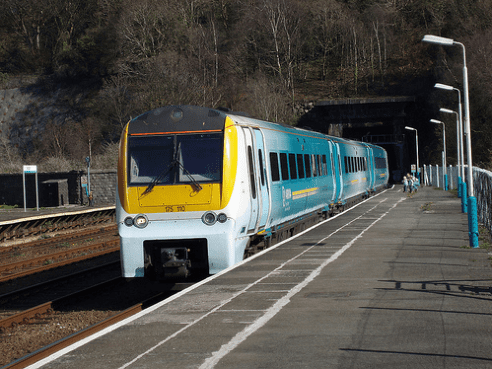}}\hfill
	\subfigure[(c)]
	{\includegraphics[width=0.162\linewidth]{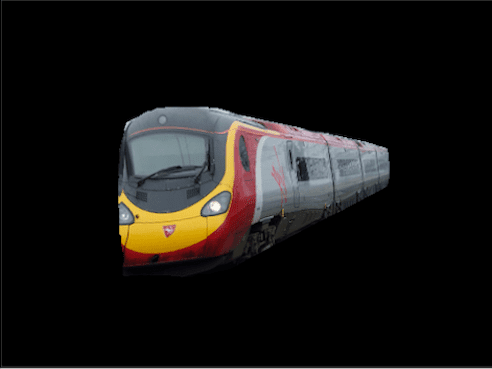}}\hfill
	\subfigure[(d)]
	{\includegraphics[width=0.162\linewidth]{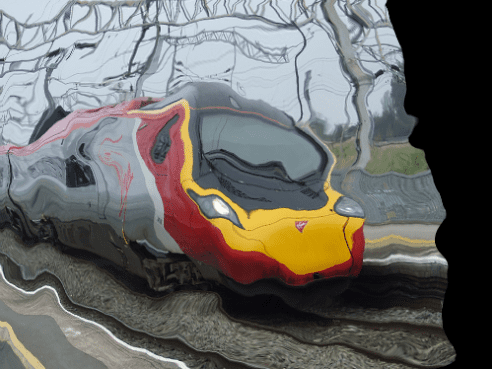}}\hfill
	\subfigure[(e)]
	{\includegraphics[width=0.162\linewidth]{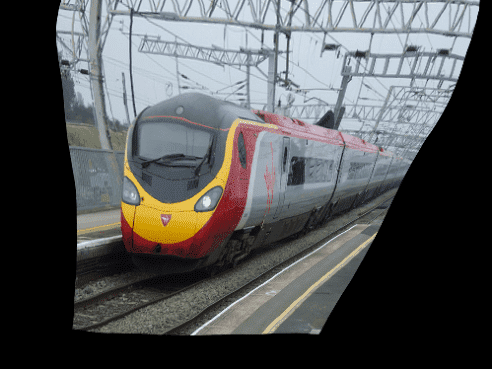}}\hfill
	\subfigure[(f)]
	{\includegraphics[width=0.162\linewidth]{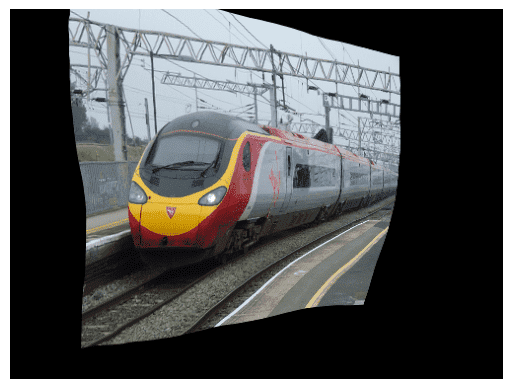}}\hfill\\
	\caption{\textbf{Qualitative results on the  TSS~\cite{taniai2016joint} benchmarks}. (a) source image, (b) target image, (c) ground-truth,  (d) GLU-Net~\cite{truong2020glu}, (e) DMP, and (f) DMP$\dagger$-ResN. It is clearly visible that warped source images produced by our models resemble the target images. Note that more accurate flow fields are estimated when ResNet is used for the feature backbone network.}\label{img:6}\vspace{-10pt}
\end{figure*}

\newpage

\begin{figure*}[t]
	\centering
	\renewcommand{\thesubfigure}{}
	\subfigure[]
	{\includegraphics[width=0.162\linewidth]{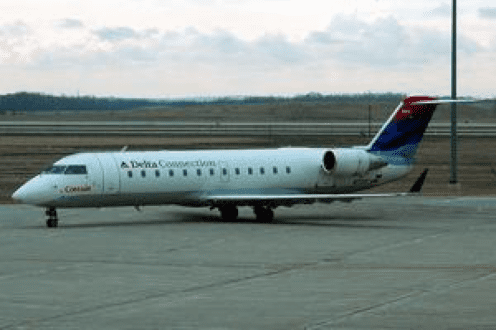}}
	\subfigure[]
	{\includegraphics[width=0.162\linewidth]{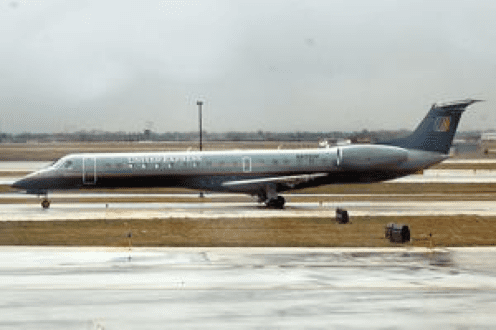}}
	\subfigure[]
	{\includegraphics[width=0.162\linewidth]{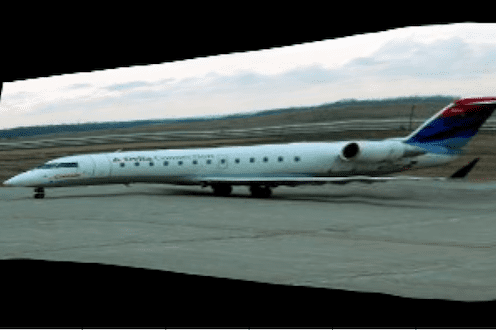}}
	\subfigure[]
	{\includegraphics[width=0.162\linewidth]{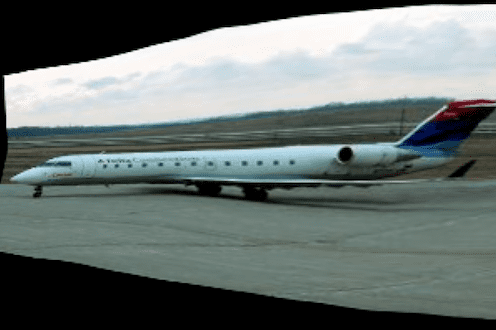}}
	\subfigure[]
	{\includegraphics[width=0.162\linewidth]{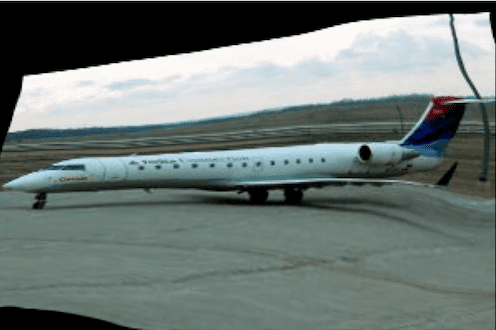}}
	\subfigure[]
	{\includegraphics[width=0.162\linewidth]{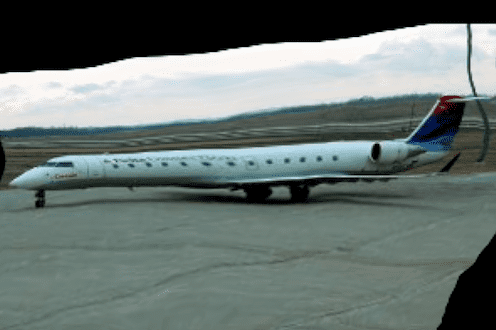}}\\
	\vspace{-21.5pt}
	\subfigure[]
	{\includegraphics[width=0.162\linewidth]{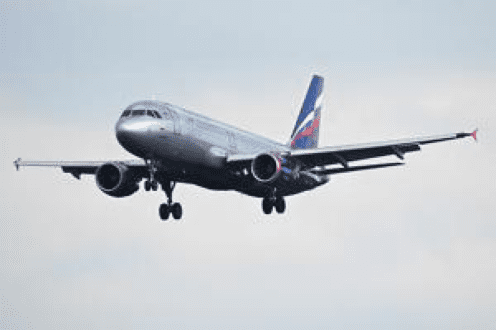}}
	\subfigure[]
	{\includegraphics[width=0.162\linewidth]{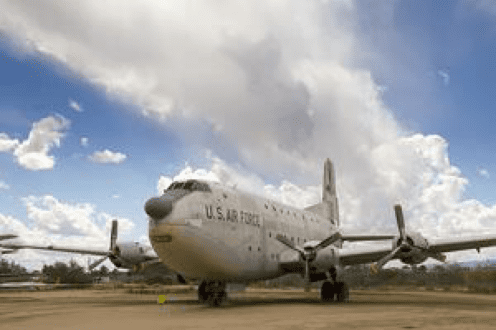}}
	\subfigure[]
	{\includegraphics[width=0.162\linewidth]{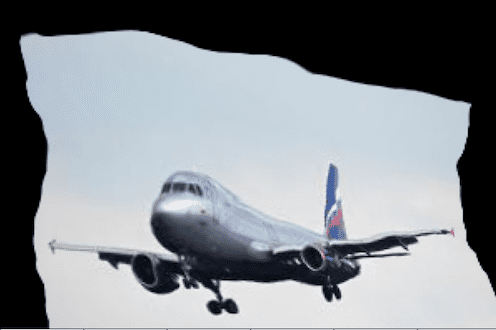}}
	\subfigure[]
	{\includegraphics[width=0.162\linewidth]{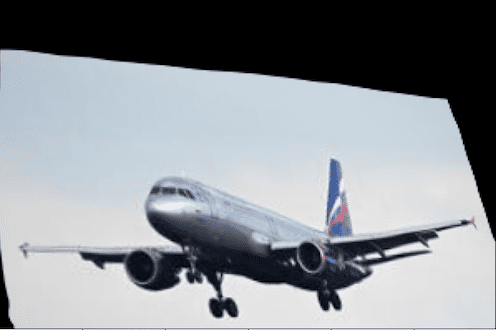}}
	\subfigure[]
	{\includegraphics[width=0.162\linewidth]{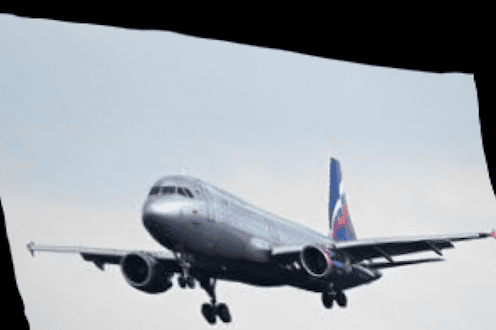}}
	\subfigure[]
	{\includegraphics[width=0.162\linewidth]{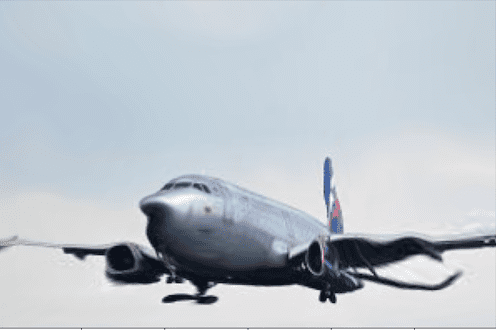}}\\
	\vspace{-21.5pt}
	\subfigure[]
	{\includegraphics[width=0.162\linewidth]{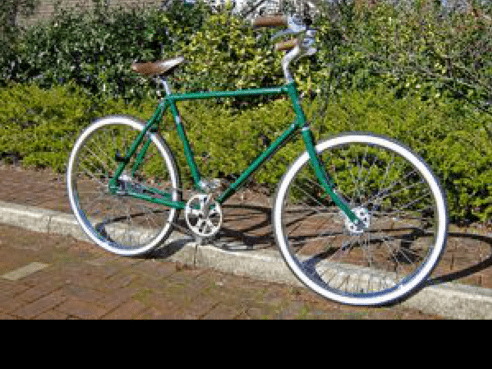}}
	\subfigure[]
	{\includegraphics[width=0.162\linewidth]{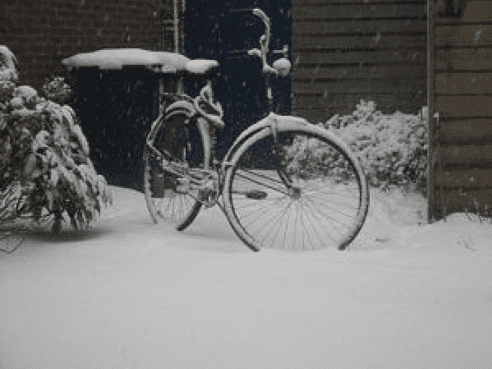}}
	\subfigure[]
	{\includegraphics[width=0.162\linewidth]{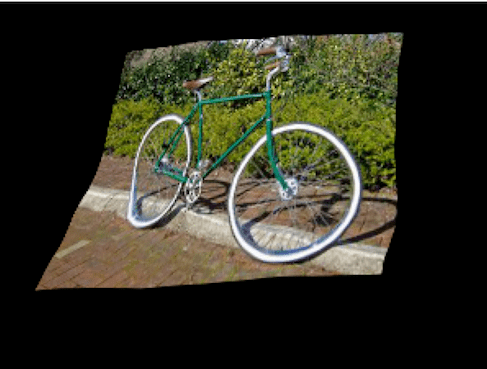}}
	\subfigure[]
	{\includegraphics[width=0.162\linewidth]{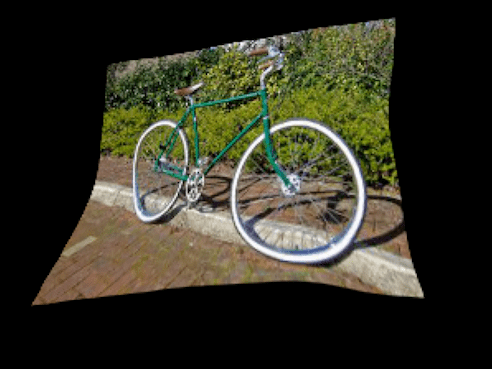}}
	\subfigure[]
	{\includegraphics[width=0.162\linewidth]{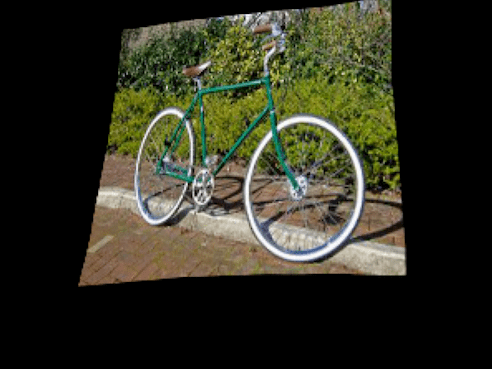}}
	\subfigure[]
	{\includegraphics[width=0.162\linewidth]{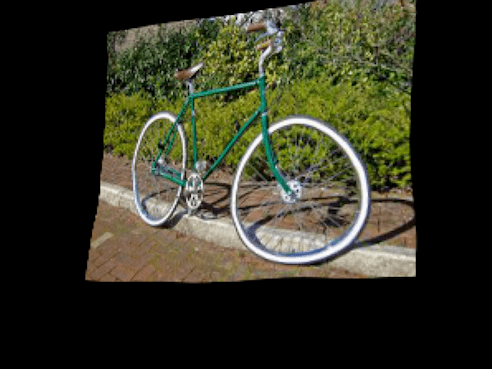}}\\
	\vspace{-21.5pt}
	\subfigure[]
	{\includegraphics[width=0.162\linewidth]{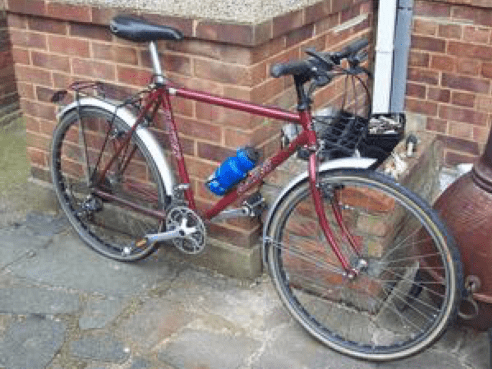}}
	\subfigure[]
	{\includegraphics[width=0.162\linewidth]{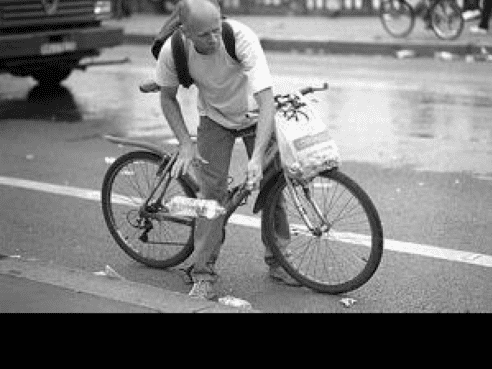}}
	\subfigure[]
	{\includegraphics[width=0.162\linewidth]{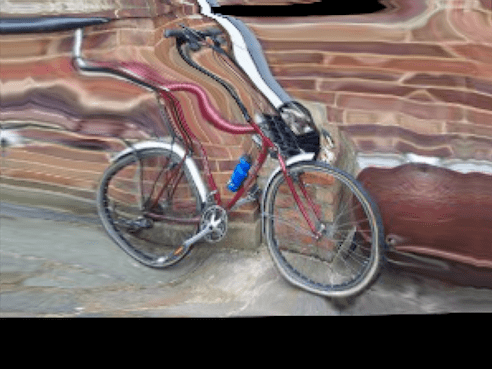}}
	\subfigure[]
	{\includegraphics[width=0.162\linewidth]{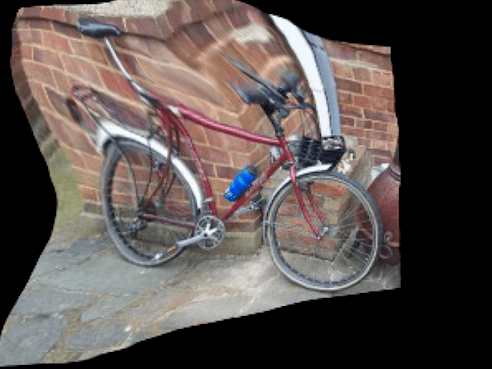}}
	\subfigure[]
	{\includegraphics[width=0.162\linewidth]{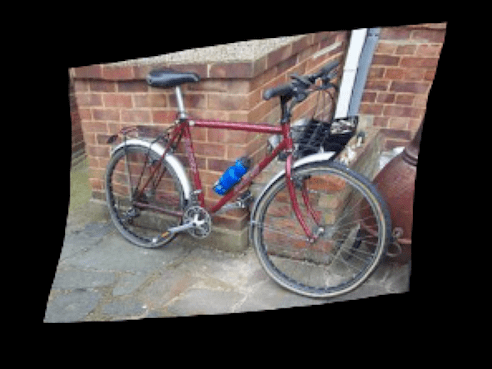}}
	\subfigure[]
	{\includegraphics[width=0.162\linewidth]{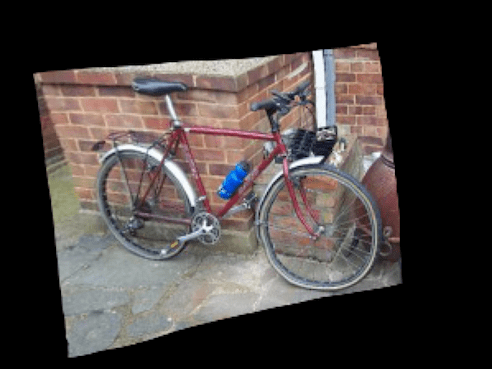}}\\
	\vspace{-21.5pt}
	\subfigure[]
	{\includegraphics[width=0.162\linewidth]{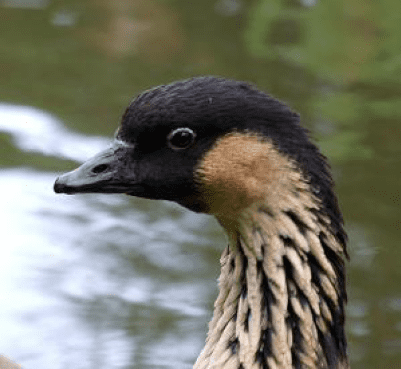}}
	\subfigure[]
	{\includegraphics[width=0.162\linewidth]{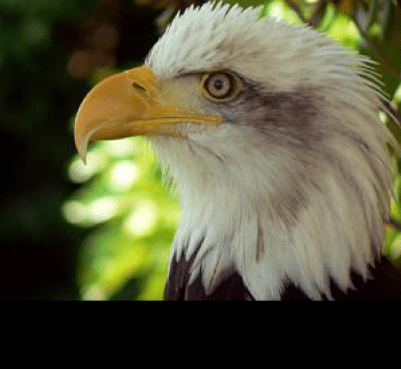}}
	\subfigure[]
	{\includegraphics[width=0.162\linewidth]{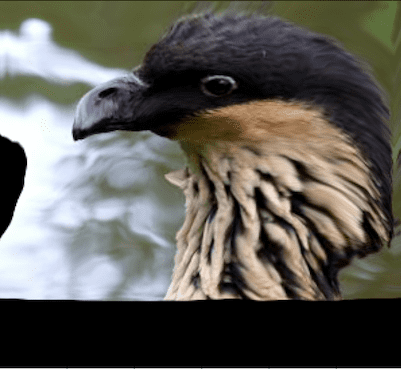}}
	\subfigure[]
	{\includegraphics[width=0.162\linewidth]{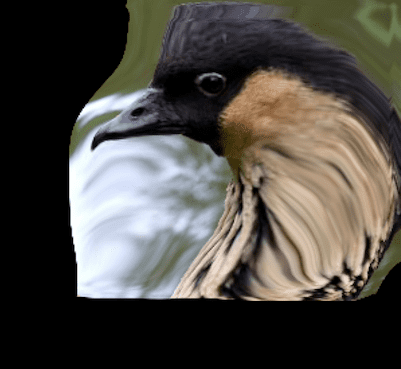}}
	\subfigure[]
	{\includegraphics[width=0.162\linewidth]{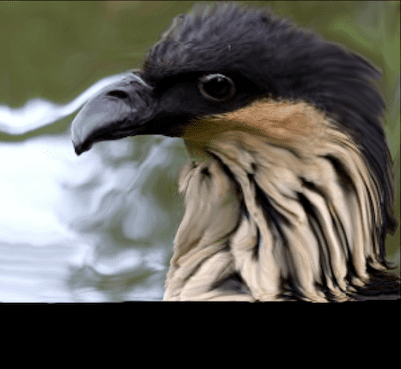}}
	\subfigure[]
	{\includegraphics[width=0.162\linewidth]{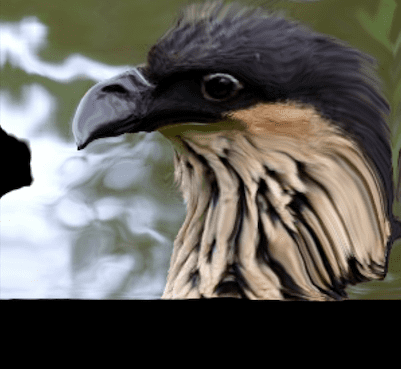}}\\
	\vspace{-21.5pt}
	\subfigure[]
	{\includegraphics[width=0.162\linewidth]{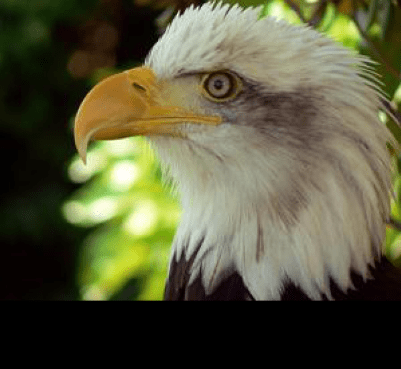}}
	\subfigure[]
	{\includegraphics[width=0.162\linewidth]{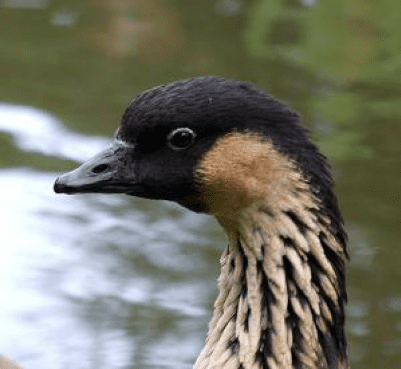}}
	\subfigure[]
	{\includegraphics[width=0.162\linewidth]{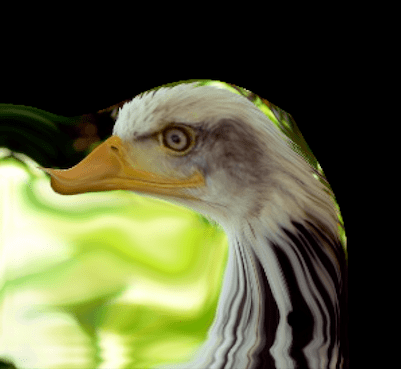}}
	\subfigure[]
	{\includegraphics[width=0.162\linewidth]{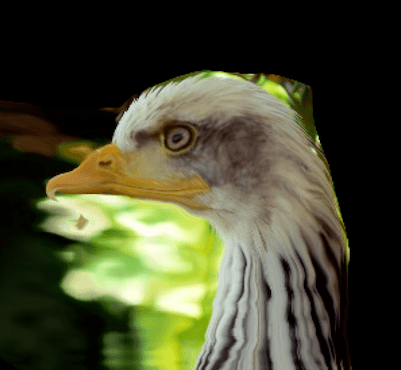}}
	\subfigure[]
	{\includegraphics[width=0.162\linewidth]{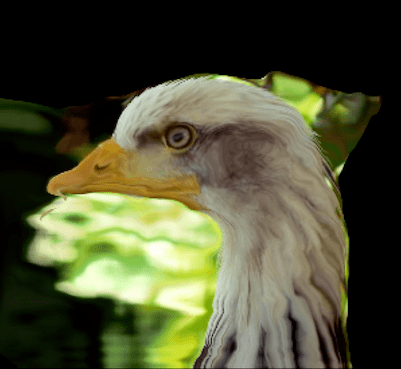}}
	\subfigure[]
	{\includegraphics[width=0.162\linewidth]{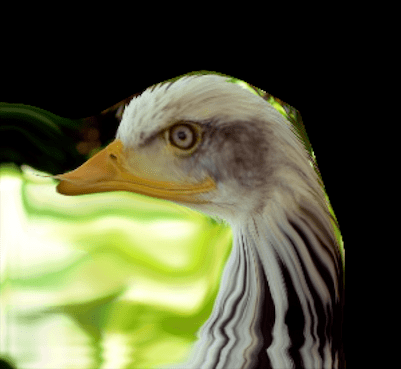}}\\
	\vspace{-21.5pt}
	\subfigure[]
	{\includegraphics[width=0.162\linewidth]{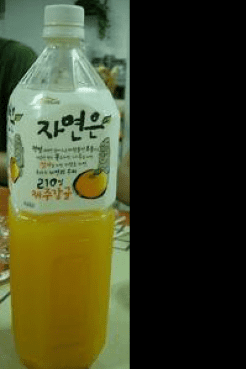}}
	\subfigure[]
	{\includegraphics[width=0.162\linewidth]{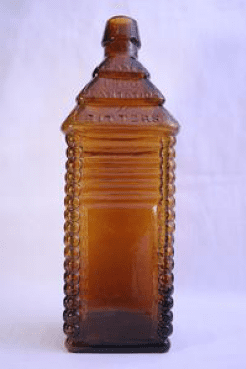}}
	\subfigure[]
	{\includegraphics[width=0.162\linewidth]{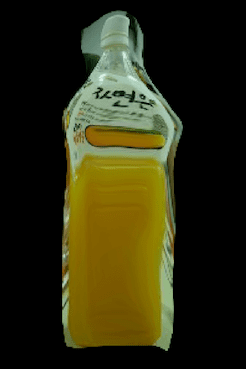}}
	\subfigure[]
	{\includegraphics[width=0.162\linewidth]{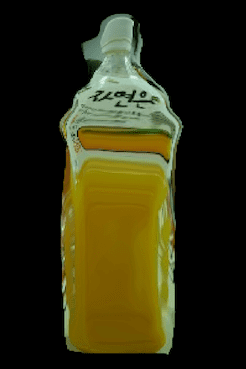}}
	\subfigure[]
	{\includegraphics[width=0.162\linewidth]{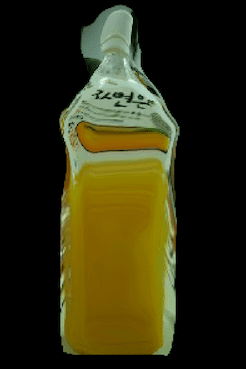}}
	\subfigure[]
	{\includegraphics[width=0.162\linewidth]{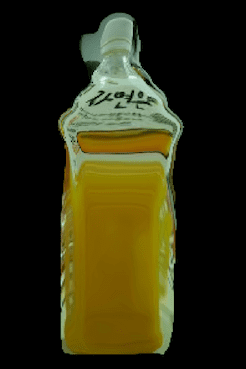}}\\
	\vspace{-21.5pt}
	\subfigure[]
	{\includegraphics[width=0.162\linewidth]{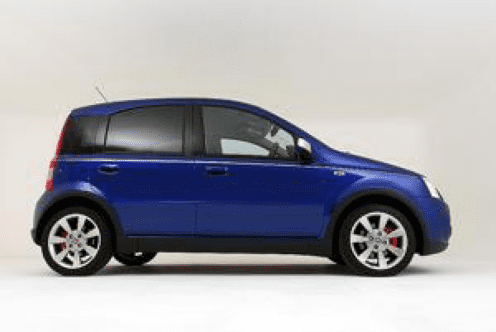}}
	\subfigure[]
	{\includegraphics[width=0.162\linewidth]{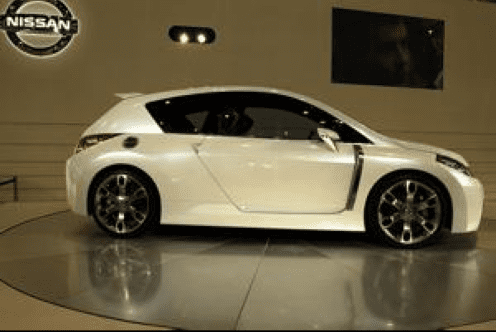}}
	\subfigure[]
	{\includegraphics[width=0.162\linewidth]{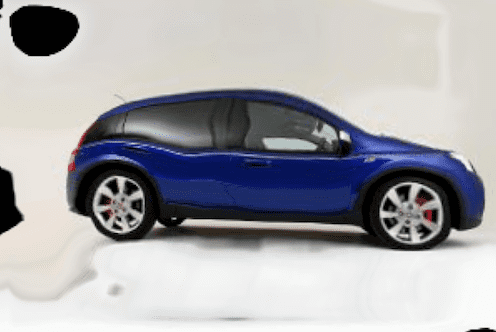}}
	\subfigure[]
	{\includegraphics[width=0.162\linewidth]{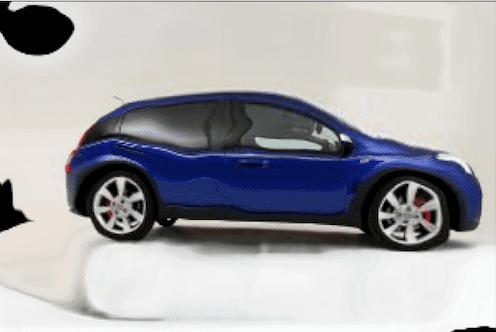}}
	\subfigure[]
	{\includegraphics[width=0.162\linewidth]{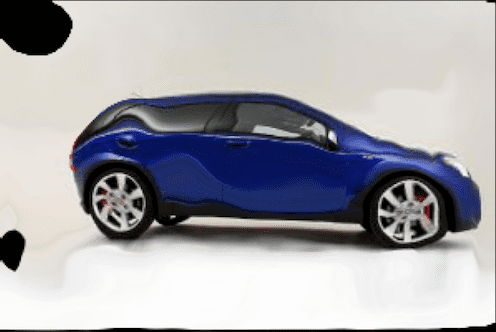}}
	\subfigure[]
	{\includegraphics[width=0.162\linewidth]{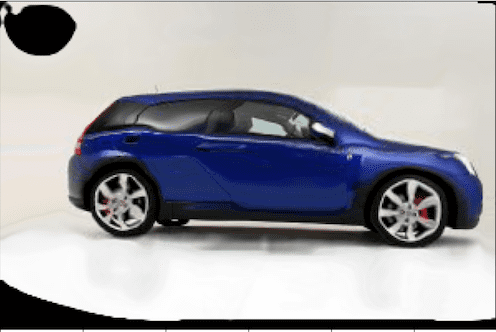}}\\
	\vspace{-21.5pt}
	\subfigure[(a)]
	{\includegraphics[width=0.162\linewidth]{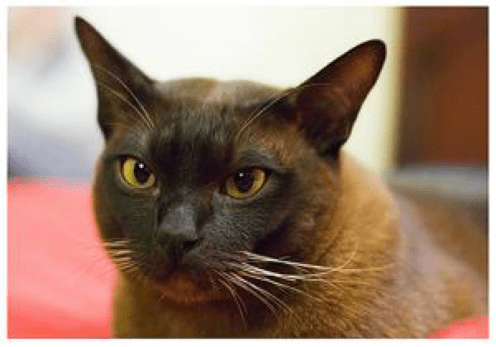}}
	\subfigure[(b)]
	{\includegraphics[width=0.162\linewidth]{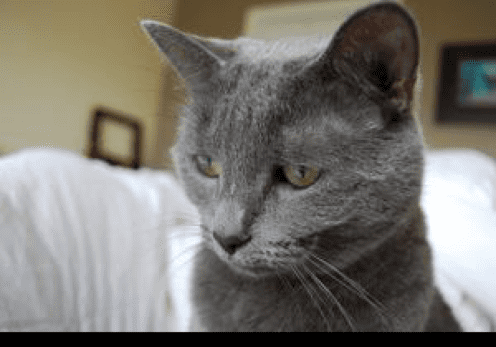}}
	\subfigure[(c)]
	{\includegraphics[width=0.162\linewidth]{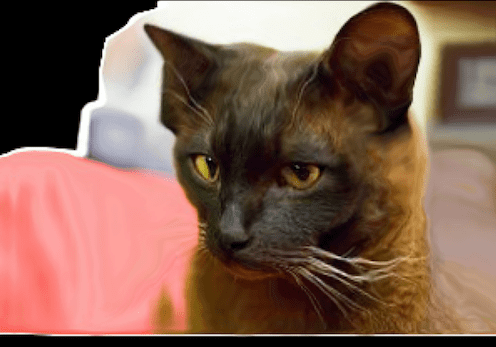}}
	\subfigure[(d)]
	{\includegraphics[width=0.162\linewidth]{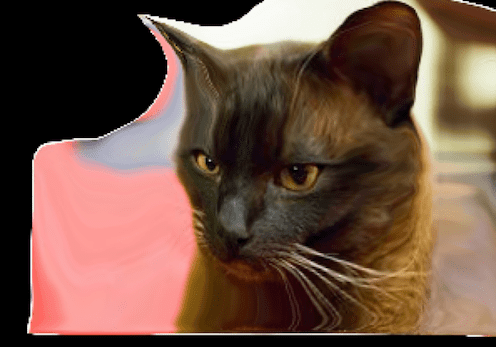}}
	\subfigure[(e)]
	{\includegraphics[width=0.162\linewidth]{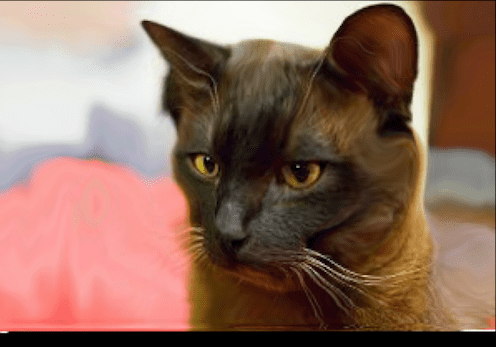}}
	\subfigure[(f)]
	{\includegraphics[width=0.162\linewidth]{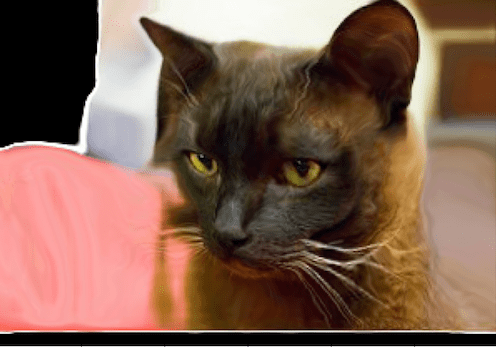}}\\
	\caption{\textbf{Qualitative results on the  PF-PASCAL~\cite{ham2016proposal} benchmarks}. (a) source image, (b) target image, (c) DMP,  (d) A-DMP, (e) DMP$\dagger$, and (f) DMP$\dagger$-ResN.}\label{img:7}\vspace{-10pt}
\end{figure*}

\end{document}